\documentclass[11pt]{article}

\usepackage[final]{acl}

\usepackage{times}
\usepackage{latexsym}

\usepackage[T1]{fontenc}

\usepackage[utf8]{inputenc}
\usepackage{hyperref}
\usepackage{url}
\usepackage{xspace}
\usepackage{enumitem}
\usepackage{amssymb}
\usepackage{float}  
\usepackage{tcolorbox}
\usepackage{tcolorbox}
\definecolor{chart}{HTML}{1f77b4}

\newtcolorbox{example}[1][]{
  colback=chart!5!white,
  colframe=chart,
  floatplacement=floating,
  title=\centering \textsf{\small #1}
}

\newtcbox{\hlprimarytab}{on line, box align=base, colback=BlueGreen!20,colframe=blue,size=fbox,arc=3pt, before upper=\strut, top=-2.5pt, bottom=-4.5pt, left=-2pt, right=-2pt, boxrule=0pt}
\newtcbox{\hlsecondarytab}{on line, box align=base, colback=WildStrawberry!10,colframe=orange,size=fbox,arc=3pt, before upper=\strut, top=-2.5pt, bottom=-4.5pt, left=-2pt, right=-2pt, boxrule=0pt}
\newtcbox{\hlwhite}{on line, box align=base, colback=WildStrawberry!8,colframe=white,size=fbox,arc=2pt, before upper=\strut, top=-3pt, bottom=-4.5pt, left=-2pt, right=-2pt, boxrule=0pt}
\newtcbox{\hlyellow}{on line, box align=base, colback=BlueGreen!10,colframe=white,size=fbox,arc=2pt, before upper=\strut, top=-3pt, bottom=-4.5pt, left=-2pt, right=-2pt, boxrule=0pt}

\newcommand{\dar}[1]{%
  \hlsecondarytab{{\scriptsize$\downarrow$~#1}}%
}

\newcommand{\uar}[1]{%
  \hlprimarytab{{\scriptsize$\uparrow$~#1}}%
}

\newcommand{\uarbig}{\scriptsize\hlprimarytab{$\uparrow\uparrow$}}

\newcommand{\orange}[1]{{\hlsecondarytab{#1}}}
\newcommand{\blue}[1]{{\hlprimarytab{#1}}}

\usepackage{booktabs}
\usepackage{wrapfig}
\usepackage{graphicx}
\usepackage[export]{adjustbox}
\usepackage{longtable}
\usepackage{multirow}
\usepackage{multicol}
\usepackage{tabularx}
\usepackage{booktabs}
\usepackage{amsmath,amsthm}
\newtheorem{theorem}{Theorem}

\usepackage{multirow}
\usepackage{hyperref}
\usepackage{url}
\usepackage{enumitem}
\usepackage{graphicx}
\usepackage{subcaption}
\usepackage[table]{xcolor} 
\usepackage{xspace}
\usepackage{algorithm}
\usepackage{algorithmicx}
\usepackage{algpseudocode}
 \usepackage{xcolor}
 \usepackage[dvipsnames]{xcolor}
\usepackage{framed}
\usepackage{xcolor} 
\usepackage{microtype}

\usepackage{inconsolata}

\usepackage{graphicx}

\title{Spectral Characterization and Mitigation of \\ Sequential Knowledge Editing Collapse}

\author{
  \textbf{Chi Zhang}\textsuperscript{1},
  \textbf{Mengqi Zhang}\textsuperscript{1}\thanks{Corresponding author.},
  \textbf{Xiaotian Ye}\textsuperscript{2}, 
  \textbf{Runxi Cheng}\textsuperscript{3}, \\
  \textbf{Zisheng Zhou}\textsuperscript{1}, 
  \textbf{Ying Zhou}\textsuperscript{1}, 
  \textbf{Pengjie Ren}\textsuperscript{1}, 
  \textbf{Zhumin Chen}\textsuperscript{1} \\
  \textsuperscript{1}Shandong University \\
  \textsuperscript{2}School of Computer Science, Beijing University of Posts and Telecommunications \\
  \textsuperscript{3}Tsinghua Shenzhen International Graduate School, Tsinghua University \\
  \small{
    \href{mailto:mengqi.zhang@sdu.edu.cn}{zhangchi2001@sdu.edu.cn, mengqi.zhang@sdu.edu.cn}
    }
}
\newcommand{\themodel}{{REVIVE}\xspace}
\begin{document}
\maketitle
\begin{abstract}
Sequential knowledge editing in large language models often causes catastrophic collapse of the model's general abilities, especially for parameter-modifying methods. Existing approaches mitigate this issue through heuristic constraints on parameter updates, yet the mechanisms underlying such degradation remain insufficiently understood. In this work, we present a spectral analysis of sequential knowledge editing and show that a model’s general abilities are closely associated with dominant singular directions of pretrained weight matrices. These directions are highly sensitive to perturbations and are progressively disrupted by repeated edits, closely tracking the collapse in both editing efficacy and general performance. Building on this insight, we propose \themodel, a plug-and-play framework that stabilizes sequential editing by explicitly preserving the dominant singular subspace. \themodel represents parameter updates in the spectral basis of the original weights and filters components that would interfere with the protected region. Extensive experiments across multiple models and benchmarks show that \themodel consistently improves editing efficacy while substantially preserving general abilities under long-horizon sequential editing, including extreme settings with up to 20,000 edits. 
\end{abstract}

\section{Introduction}
Large language models (LLMs) often generate outdated or incorrect information due to limitations in pretraining data or evolving real-world knowledge~\citep{de2021editing}. {Knowledge editing} \citep{memit,zhang2026disentangling,zhanguncovering,zhang2025kele} provides a lightweight way to correct specific facts without retraining models from scratch. In practice, however, edits are rarely isolated: models are updated repeatedly over time, leading to the setting of \textbf{sequential knowledge editing}~\citep{alphaedit}, which requires both accurate edits and the preservation of general abilities~\citep{rect}. These challenges are particularly pronounced for \emph{parameter-modifying} methods that directly update model weights, which are the focus of this work.



\begin{figure}[t]
    \centering
    \begin{subfigure}{0.48\textwidth}
        \centering
        \includegraphics[width=\linewidth]{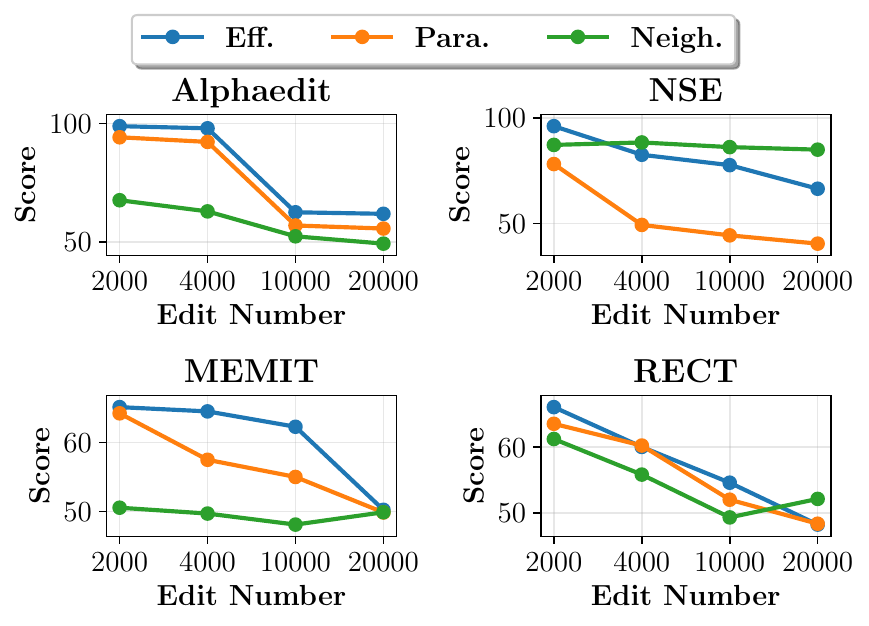}
    \end{subfigure}
    \caption{Performance of representative sequential knowledge editing methods on \textsc{CounterFact} using LLaMA3. }
    \label{fig:motivation}
    \vspace{-1.5em}
\end{figure}

Despite the development of several sequential knowledge editing methods, including \textsc{RECT}~\citep{rect}, \textsc{PRUNE}~\citep{prune}, and \textsc{AlphaEdit}~\citep{alphaedit}, long-horizon degradation remains largely unresolved. As illustrated in Figure~\ref{fig:motivation}, although these approaches appear to be effective in short or small-scale setting, they still suffer pronounced drops in both editing efficacy and general abilities as the number of edits increases. A key limitation is that most existing methods primarily mitigate interference across updates through heuristic constraints or regularization, such as bounding update magnitudes, reducing interference across edits, or minimizing disruption to unrelated knowledge, without explicitly modeling how parameter updates interact with the intrinsic structure of pretrained weights. In particular, they do not clarify which parts of the pretrained weights are most relevant to general abilities, or how repeated updates accumulate and gradually degrade them.  As a result, these methods provide limited mechanistic insight into why performance progressively degrades as edits accumulate. 



In this work, we conduct a systematic analysis of sequential knowledge editing from a spectral perspective (\S\ref{sec:analysis}). Through a detailed study of the singular value decomposition (SVD) of weight matrices, we obtain three key findings: (i) general abilities are concentrated in a low-rank spectral structure spanned by dominant singular directions (\S\ref{anal:where}); 
(ii) this structure is unusually fragile, such that even small perturbations aligned with dominant directions can disproportionately degrade general performance (\S\ref{anal:how}); and 
(iii) repeated edits progressively distort these dominant directions in practice (\S\ref{sec:memitlongsequenceediting}). Together, these findings provide strong empirical evidence that \textbf{interference with the dominant singular subspace is closely tied to failure in sequential editing, suggesting a primary mechanism underlying this collapse.}


Despite this insight, preserving dominant spectral structure during editing is nontrivial: editing objectives are local and offer little direct control over global spectral properties, and parameter-modifying editors differ substantially across methods. To address these challenges, we develop a \textbf{\underline{R}}obust s\textbf{\underline{E}}quential editing \textbf{\underline{V}}ia dom\textbf{\underline{I}}nant subspace preser\textbf{\underline{V}}ation fram\textbf{\underline{E}}work (\themodel) (\S\ref{method}), a plug-and-play framework that stabilizes sequential knowledge editing by explicitly protecting dominant spectral structure. The core idea of \themodel is to analyze parameter updates in the singular vector basis of the original weight matrix, which enables a fine-grained characterization of how edits interact with intrinsic functional directions. Based on this representation, \themodel identifies the dominant subspace using a spectral energy criterion and constructs constrained updates by filtering out components that interfere with this critical region. This design allows factual knowledge to be incorporated through lower-energy directions while preserving the dominant spectral structure associated with general abilities. 

Our contributions can be summarized as follows:
\begin{itemize}[leftmargin=*]

\item We present a systematic spectral analysis of sequential knowledge editing, revealing that interference with the dominant singular subspace of pretrained weight matrices is a key mechanism underlying model collapse.

\item Based on this analysis, we propose \themodel, a simple and plug-and-play framework that stabilizes sequential knowledge editing by explicitly preserving the dominant singular subspace during parameter updates.

\item We conduct extensive experiments on multiple models and benchmarks, demonstrating that \themodel consistently outperforms state-of-the-art methods in both editing efficacy and the preservation of general abilities under sequential editing.
\end{itemize}

\section{Why Sequential Editing Collapses: A Spectral Perspective}
\label{sec:analysis}
To understand why sequential editing leads to model collapse, this section presents a spectral analysis of parameter matrices. We begin by viewing each weight matrix as a composition of independent input-output mappings derived from its Singular Value Decomposition (SVD). Based on this perspective, we empirically investigate three key questions: (i) \emph{where a model’s general abilities are concentrated within these spectral components}, (ii) \emph{how sensitive these components are to perturbations}, and (iii) \emph{how sequential editing progressively distorts these components in practice}.

Together, these analyses lead to a unified insight: \textbf{sequential editing fails when cumulative update progressively corrupt the dominant singular directions of weight matrices, which are strongly associated with general abilities.} As mainstream editing methods primarily target feed-forward network (FFN) layers for modification \citep{memit,rome}, we ground our investigation in the FFN matrices of LLaMA3-8B. 

\subsection{Spectral View of Weight Matrices}
\label{anal:spect}
From a spectral perspective, a parameter matrix $\mathbf{W} \in \mathbb{R}^{m \times n}$ can be decomposed into a set of {independent input-output mappings} using Singular Value Decomposition (SVD):
\begin{equation} \label{eq:svd-decomposition}
\mathbf{W} = \mathbf{U} \Sigma \mathbf{V}^\top = \sum_{i=1}^{r} \sigma_i \mathbf{u}_i \mathbf{v}_i^\top,
\end{equation}
where $\mathbf{U}\in\mathbb{R}^{m \times m}$ and $\mathbf{V}\in\mathbb{R}^{n \times n}$ contain orthogonal left and right singular vectors, respectively, and $\Sigma \in\mathbb{R}^{m \times n}$ contains the singular values $\sigma_1 \geq \sigma_2 \geq \dots \geq \sigma_r \geq 0$. 

Each rank-one component $\sigma_i \mathbf{u}_i \mathbf{v}_i^\top$ acts as a distinct {input-output mapping}: an input vector $\mathbf{x}\in\mathbb{R}^n$ is projected onto $\mathbf{v}_i$, scaled by $\sigma_i$, and expanded along $\mathbf{u}_i$ to produce the output. The orthogonality of the singular vectors ensures these mappings operate independently. Pretraining learns this highly structured functional decomposition, making it a critical component of the model's general abilities \citep{llmsvd,hao2025token}. Crucially, this spectral view provides a unified lens for analyzing the failure of sequential editing. In the remainder of this section, we use this perspective to systematically investigate the three questions outlined above.



\subsection{Where Are General Abilities Concentrated?} 
\label{anal:where}
To identify where general abilities are concentrated, we evaluate model performance on GLUE tasks (MRPC, CoLA, RTE, NLI) \citep{GLUE} using weight matrices reconstructed from subsets of their singular components. 

Specifically, we define the {singular value energy} of an index set $\mathcal{I}$ as $E_\mathcal{I}$ $=$ $\sum\nolimits_{i \in \mathcal{I}} \sigma_i$, and the total energy as $E_{\text{total}}$ $=$ $\sum_{i=1}^r \sigma_i$. We then reconstruct the weight matrix using the smallest set of singular components that captures $n\%$ of the total energy, $\tilde{\mathbf{W}}^{n} \;=\; \sum\nolimits_{i \in \mathcal{I}} \sigma_i \mathbf{u}_i \mathbf{v}_i^\top$, and evaluate model performance using the reconstructed matrices.

\noindent\textbf{Finding: General abilities are highly concentrated in the dominant singular subspace.} As shown in Figure~\ref{fig:retainability}, reconstructing weight matrices with only the top $5\%$ of singular components (by energy) recovers approximately $62.6\%$ of the model’s original performance. This result indicates that a substantial portion of a model’s general abilities is concentrated in a small, low-rank subspace spanned by the singular vectors associated with the largest singular values.

\subsection{How Sensitive Are These Components to Perturbations?} 
\label{anal:how}
To evaluate the robustness of different singular components, we partition the spectrum into ten non-overlapping groups by cumulative energy ($0$--$10\%$, …, $90$--$100\%$). For each group $\mathcal{G}$, we inject a {structured perturbation} that selectively acts on the corresponding input singular directions. Concretely, we first generate
\begin{equation}
\mathbf{\Delta} = \sum_{j \in \mathcal{G}} \sum_{i=1}^r \alpha_{i,j}\, \mathbf{u}_i \mathbf{v}_j^\top,
\alpha_{i,j} \sim \mathcal{N}(0,1),
\end{equation}
which randomly remaps the input directions $\{v_j\}_{j\in G}$ to output directions. The perturbation is then normalized and scaled to a fixed strength: $\varepsilon$: $\mathbf{\tilde\Delta} = \varepsilon \cdot \tfrac{{\mathbf{\Delta}}}{\|{\mathbf{\Delta}}\|_F}.$ This ensures identical Frobenius norms across groups, thereby enabling a fair comparison. We evaluate the impact of the perturbed matrix $\mathbf{W}' = \mathbf{W} + \mathbf{\tilde\Delta}$ on general performance. A symmetric analysis on output directions is provided in Appendix \ref{app:outputsidepertuabtion} and exhibits similar trends.

\begin{figure}[t!]
    \centering
    \begin{minipage}[b]{0.23\textwidth}
        \centering
        \includegraphics[width=\textwidth]{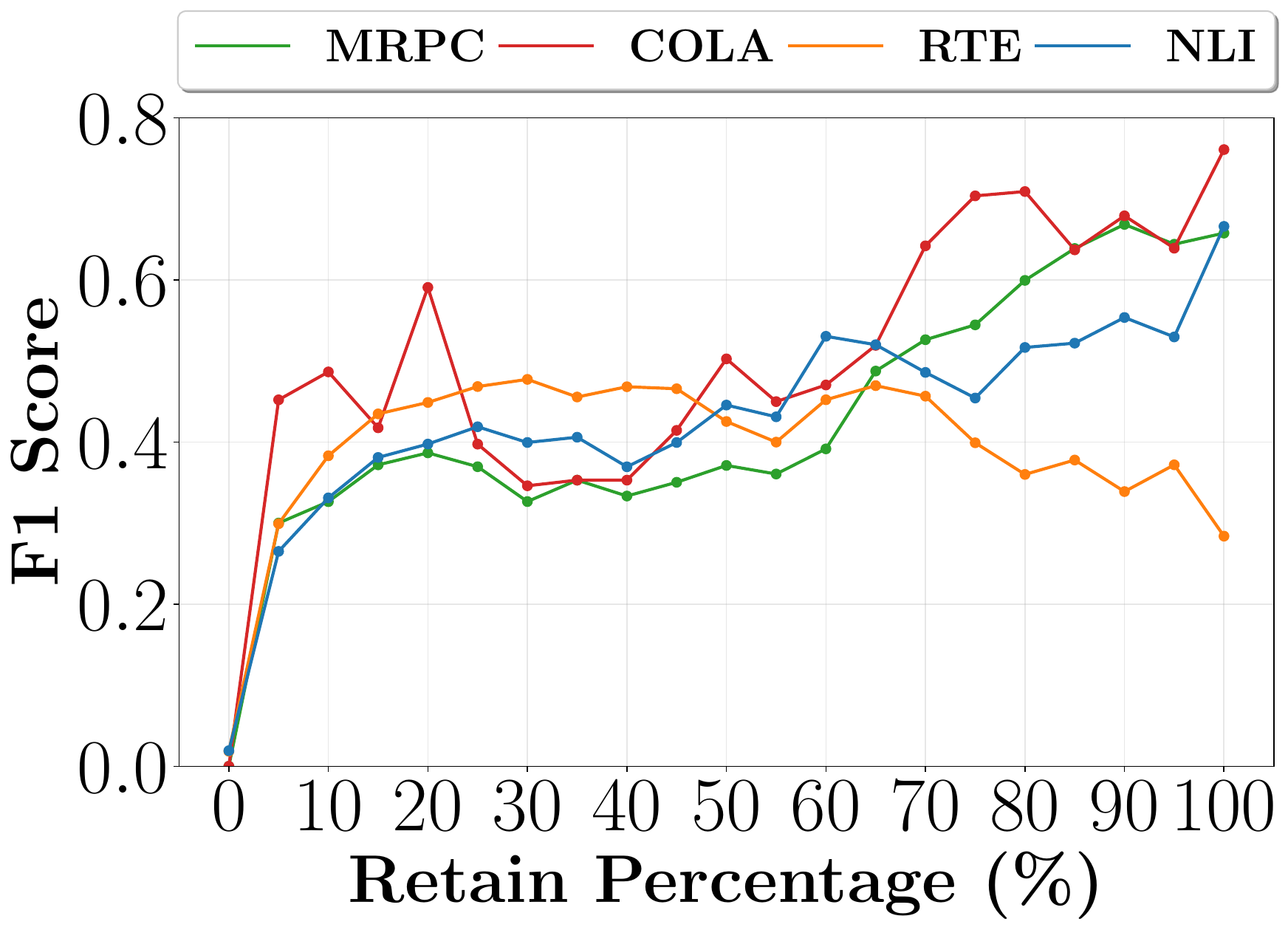}
        \caption{General ability recovery with increasing proportion of retained singular components.}
        \label{fig:retainability}
    \end{minipage}
    \hfill
    \begin{minipage}[b]{0.23\textwidth}
        \centering
        \includegraphics[width=\textwidth]{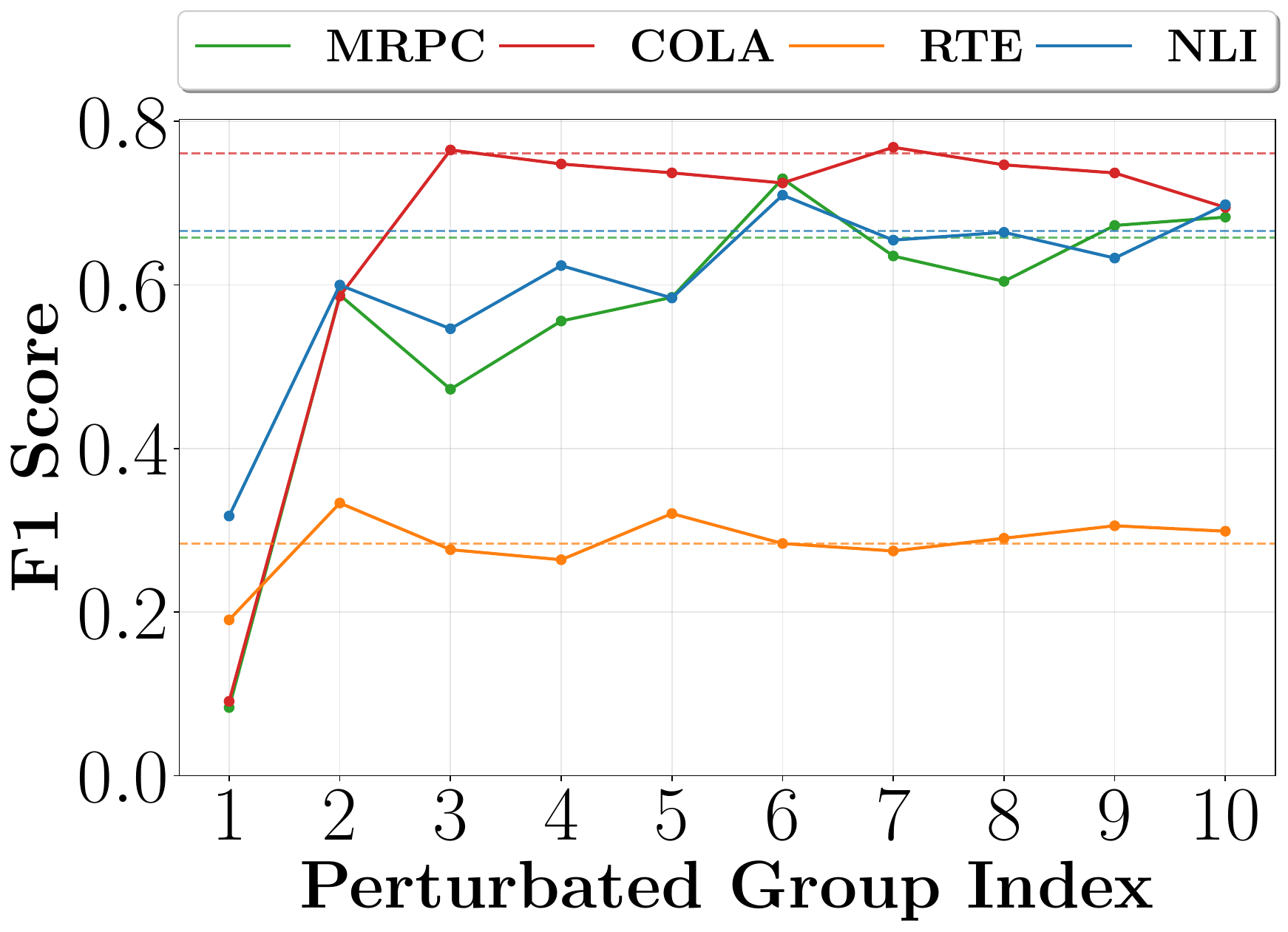}
        \caption{Sensitivity of model's general ability to perturbations across different spectral groups.}
        \label{fig:datasetB}
    \end{minipage}
\end{figure}

\begin{figure*}[thbp]
    \centering
    \captionsetup[subfigure]{font=footnotesize, skip=2pt}
    \small

    \begin{minipage}[c]{0.66\textwidth} 
        \begin{subfigure}[t]{0.23\textwidth}
            \includegraphics[width=\textwidth]{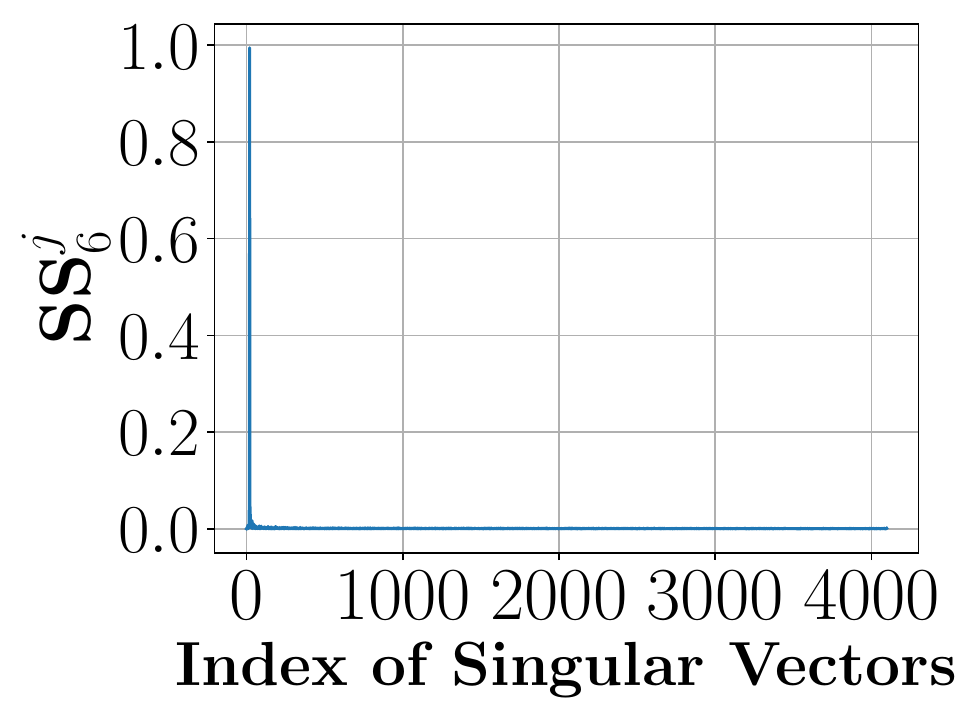}
        \end{subfigure}
        \begin{subfigure}[t]{0.23\textwidth}
            \includegraphics[width=\textwidth]{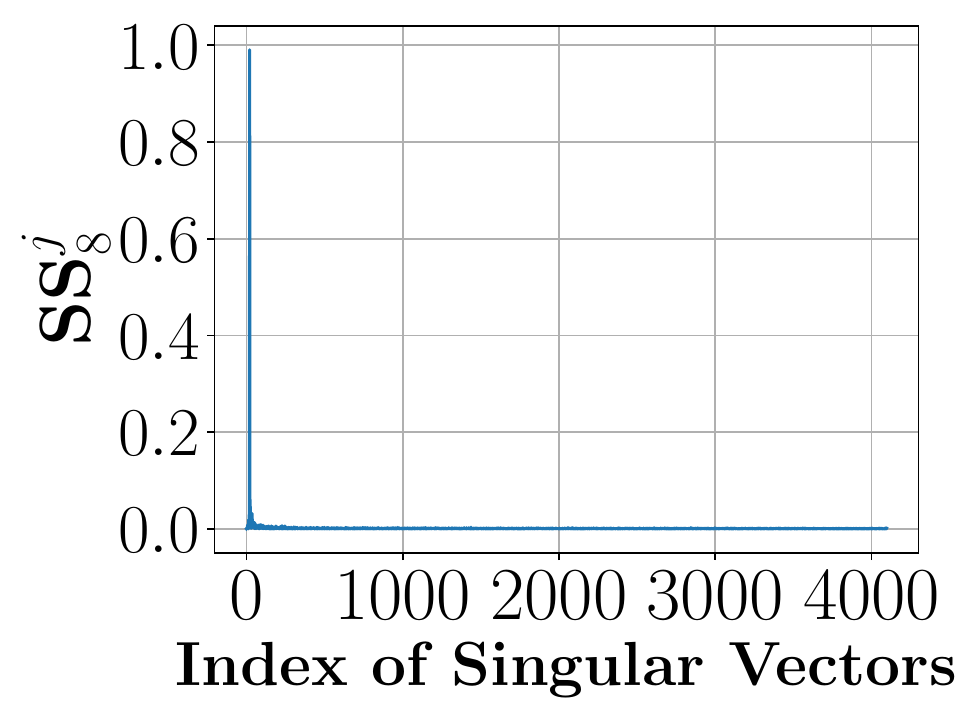}
        \end{subfigure}
        \begin{subfigure}[t]{0.23\textwidth}
            \includegraphics[width=\textwidth]{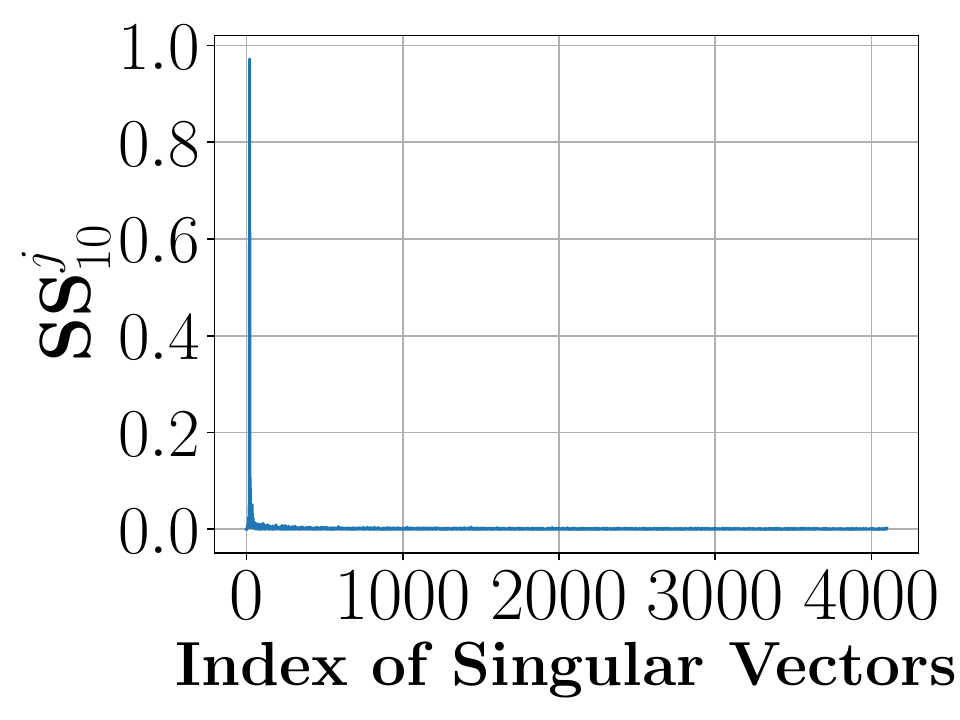}
        \end{subfigure}
        \begin{subfigure}[t]{0.23\textwidth}
            \includegraphics[width=\textwidth]{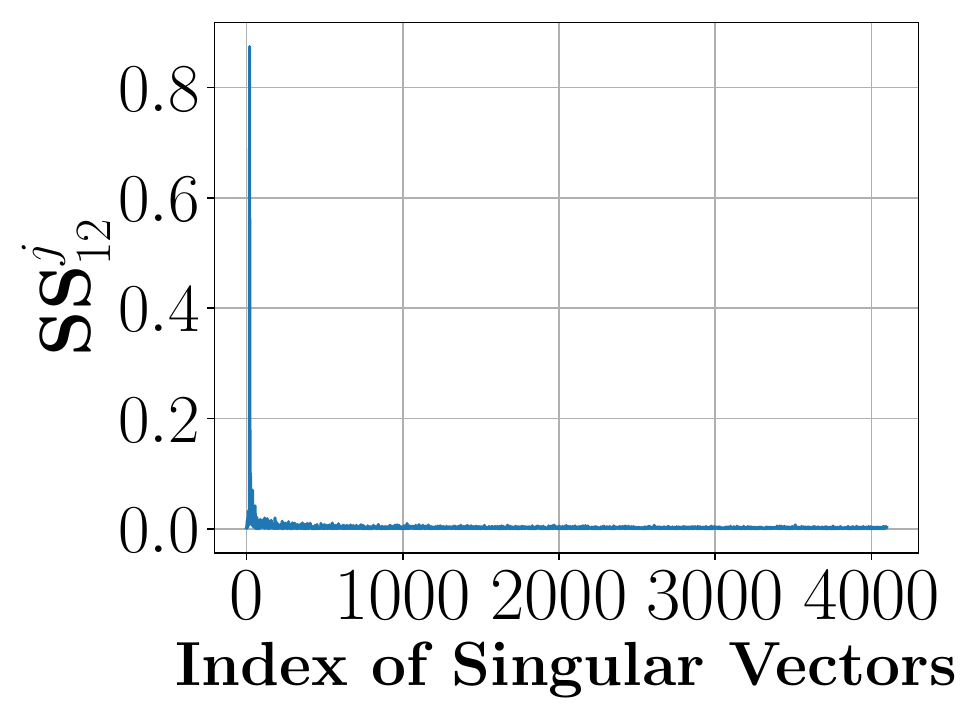}
        \end{subfigure}

        \par\vspace{0.3em}

        \begin{subfigure}[t]{0.23\textwidth}
            \includegraphics[width=\textwidth]{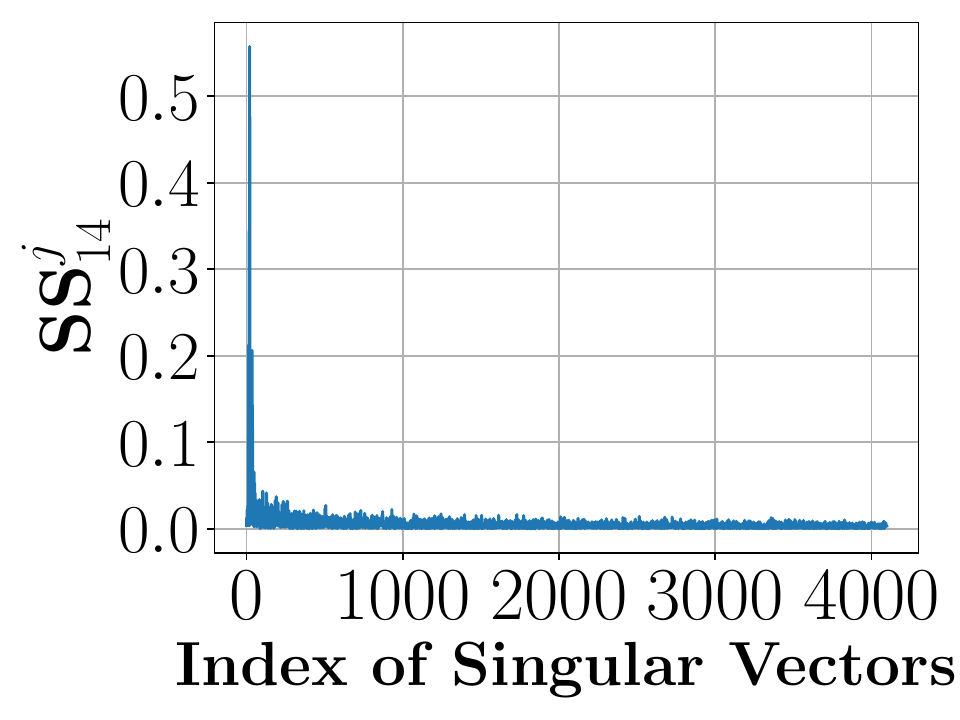}
        \end{subfigure}
        \begin{subfigure}[t]{0.23\textwidth}
            \includegraphics[width=\textwidth]{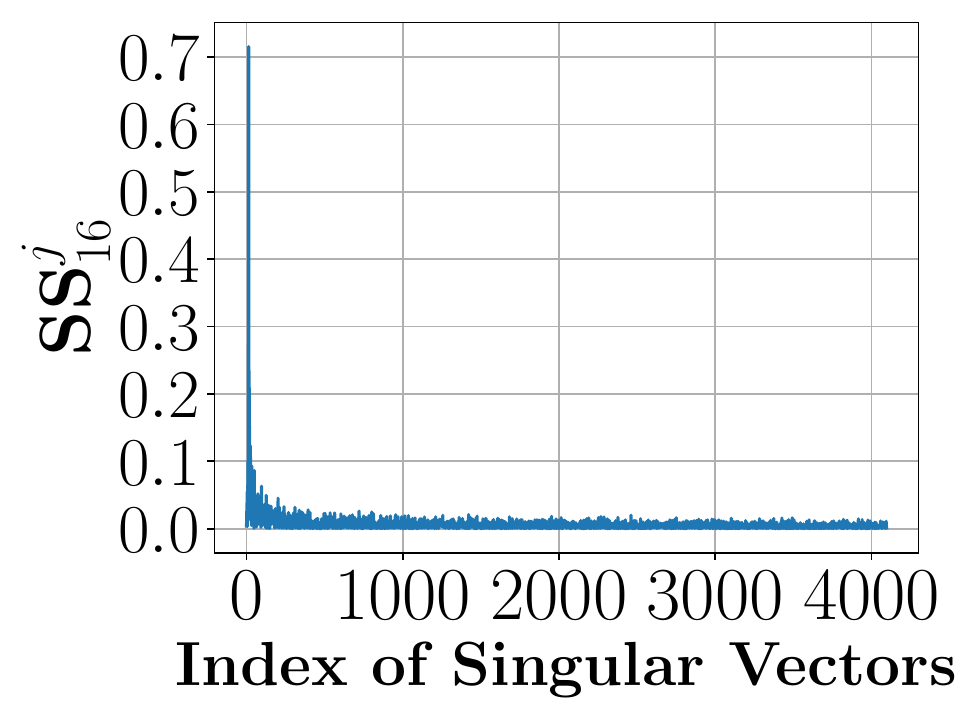}
        \end{subfigure}
        \begin{subfigure}[t]{0.23\textwidth}
            \includegraphics[width=\textwidth]{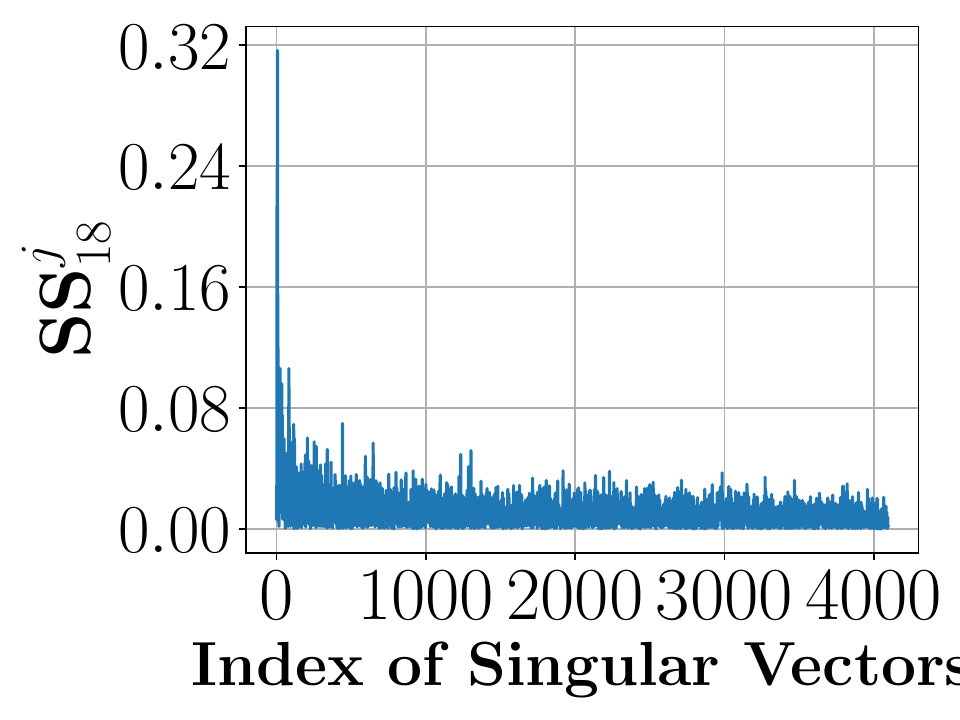}
        \end{subfigure}
        \begin{subfigure}[t]{0.23\textwidth}
            \includegraphics[width=\textwidth]{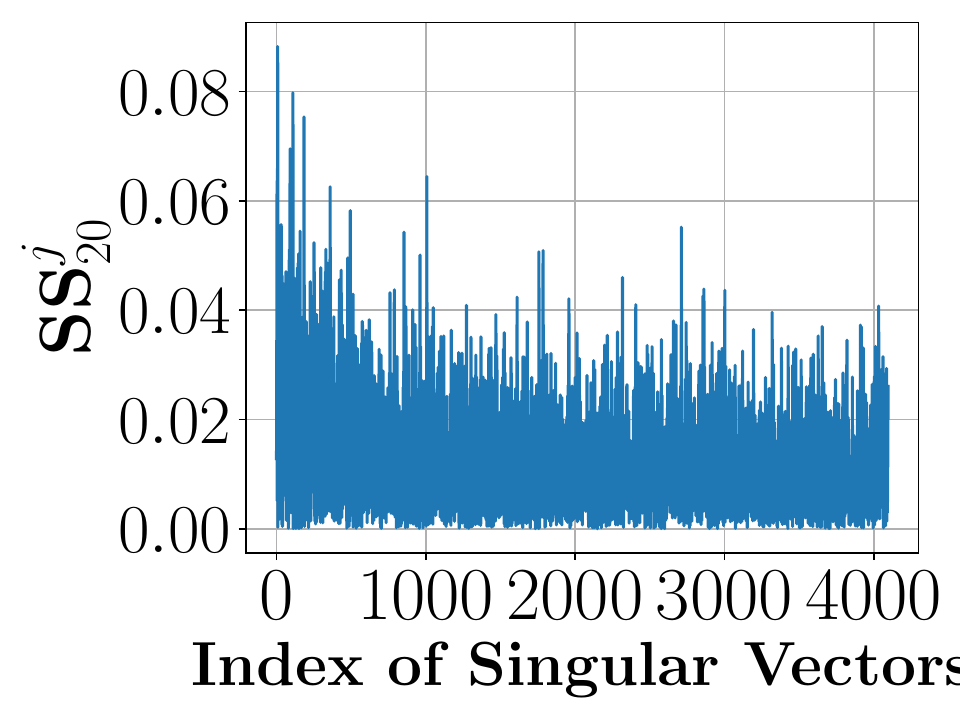}
        \end{subfigure}

        \caption{Singular vector similarity (SS) under sequential editing from rounds 6 to 20.}
        \label{fig:ss_combined}
    \end{minipage}
    \hfill
\begin{minipage}[c]{0.30\textwidth} 
        \includegraphics[
            height=\dimexpr0.65\textwidth*2/1\relax, 
            width=\linewidth,
            keepaspectratio
        ]{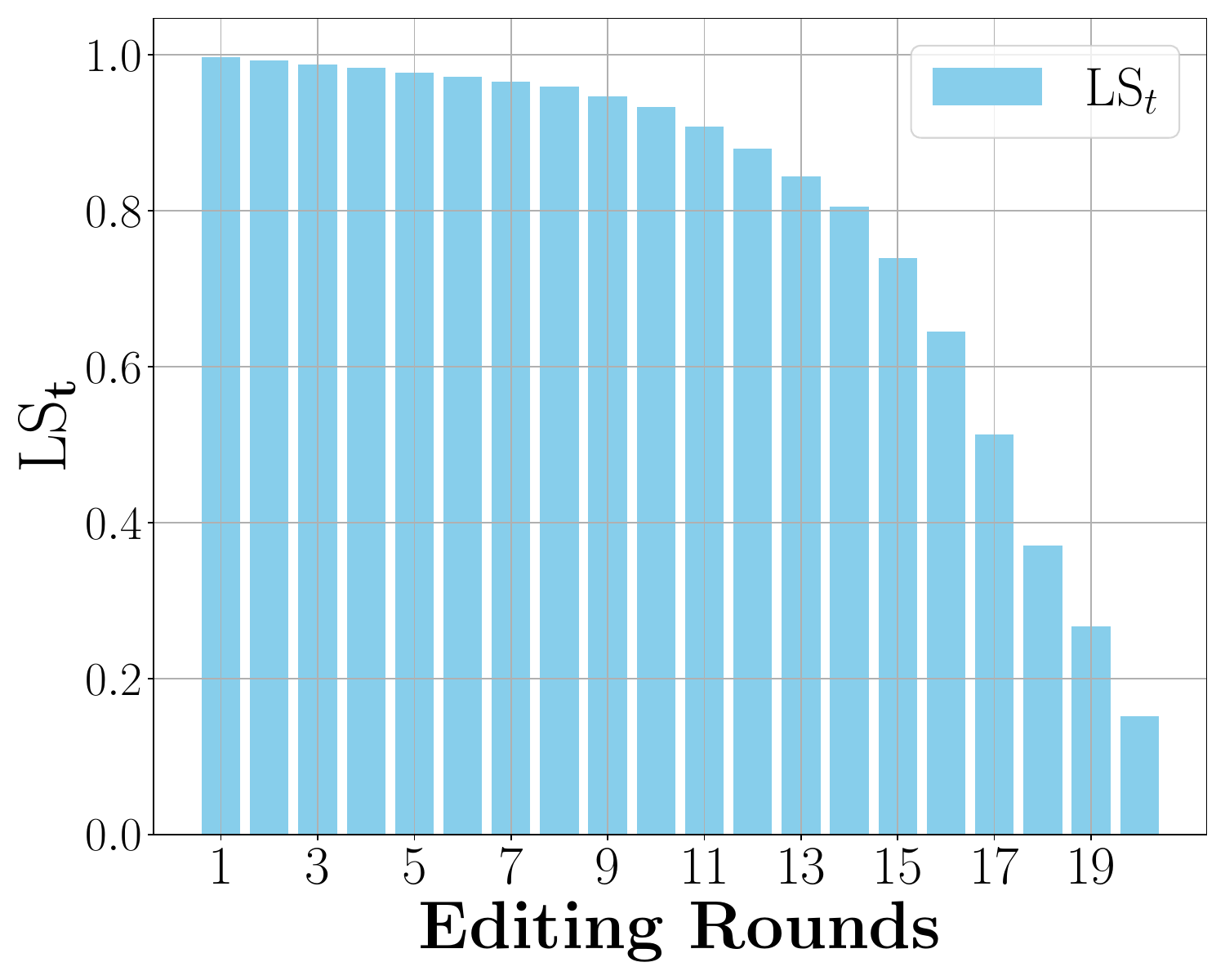}
            \vspace{-2.5em}
        \caption{$\text{LS}_t$ from rounds 1 to 20.}
        \label{fig:ls_vs_rounds}
    \end{minipage}
\end{figure*}

\noindent \textbf{Finding: The dominant singular subspace most strongly associated with general abilities is highly sensitive to perturbations.} As shown in Figure~\ref{fig:datasetB}, perturbations applied to high-energy singular components (e.g., $0$--$20\%$), which span the dominant singular subspace, lead to sharp and consistent degradation in performance. In contrast, perturbations to low-energy components ($70$--$100\%$) produce only minor or negligible effects. These results reveal a paradoxical property: \textbf{the spectral directions most associated with general abilities are also the most fragile under perturbations.} 


\subsection{How Does Sequential Editing Distort These Components in Practice?}
\label{sec:memitlongsequenceediting}

\begin{figure}[thbp]
    \centering
    \begin{minipage}[t]{0.23\textwidth}
        \centering
        \includegraphics[width=\textwidth]{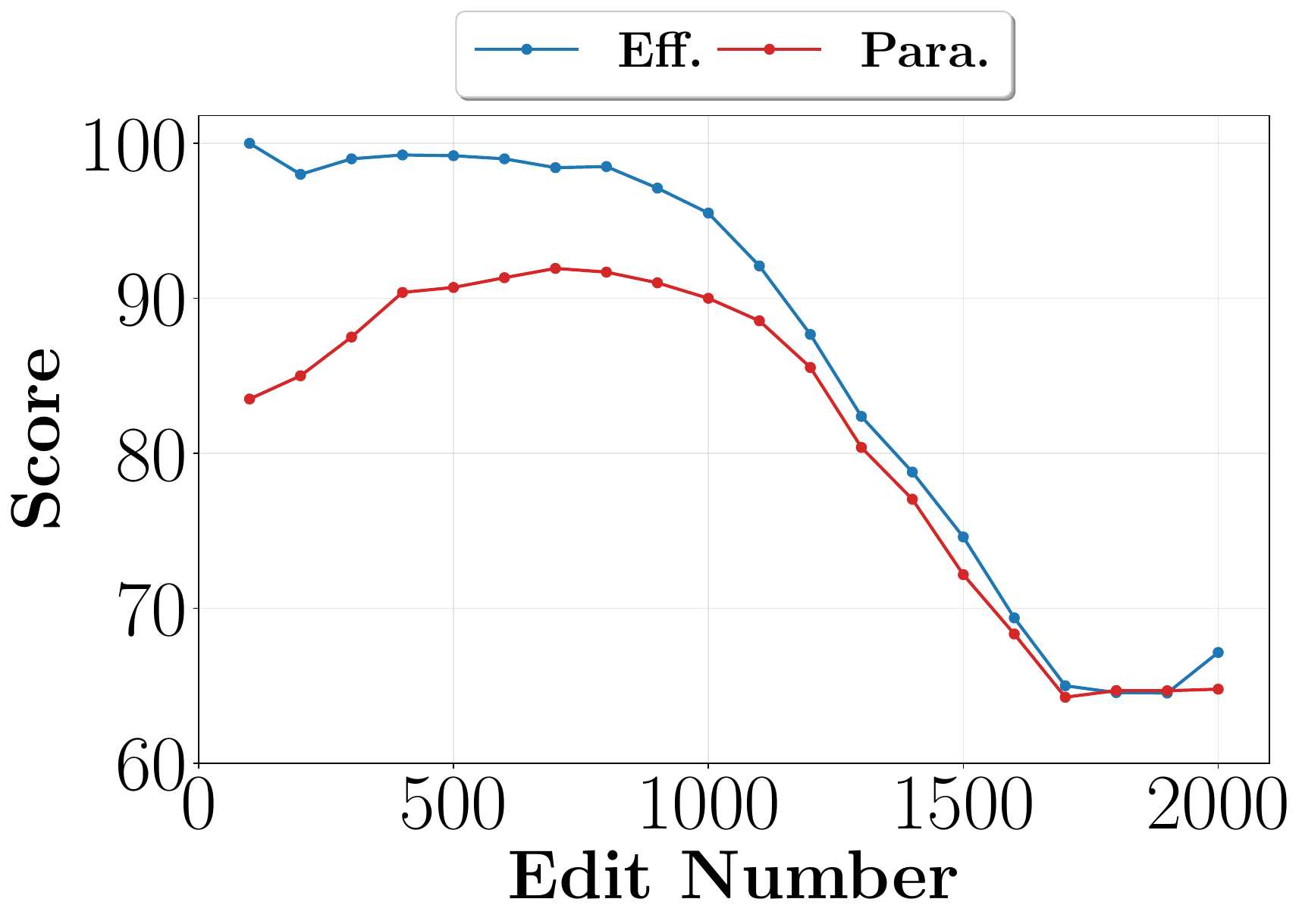}
        \caption{The Efficacy Score and Paraphrase Score vs. Edit Number.}
        \label{fig:eff-vs-run}
    \end{minipage}
    \hfill
    \begin{minipage}[t]{0.23\textwidth}
        \centering
        \includegraphics[width=\textwidth]{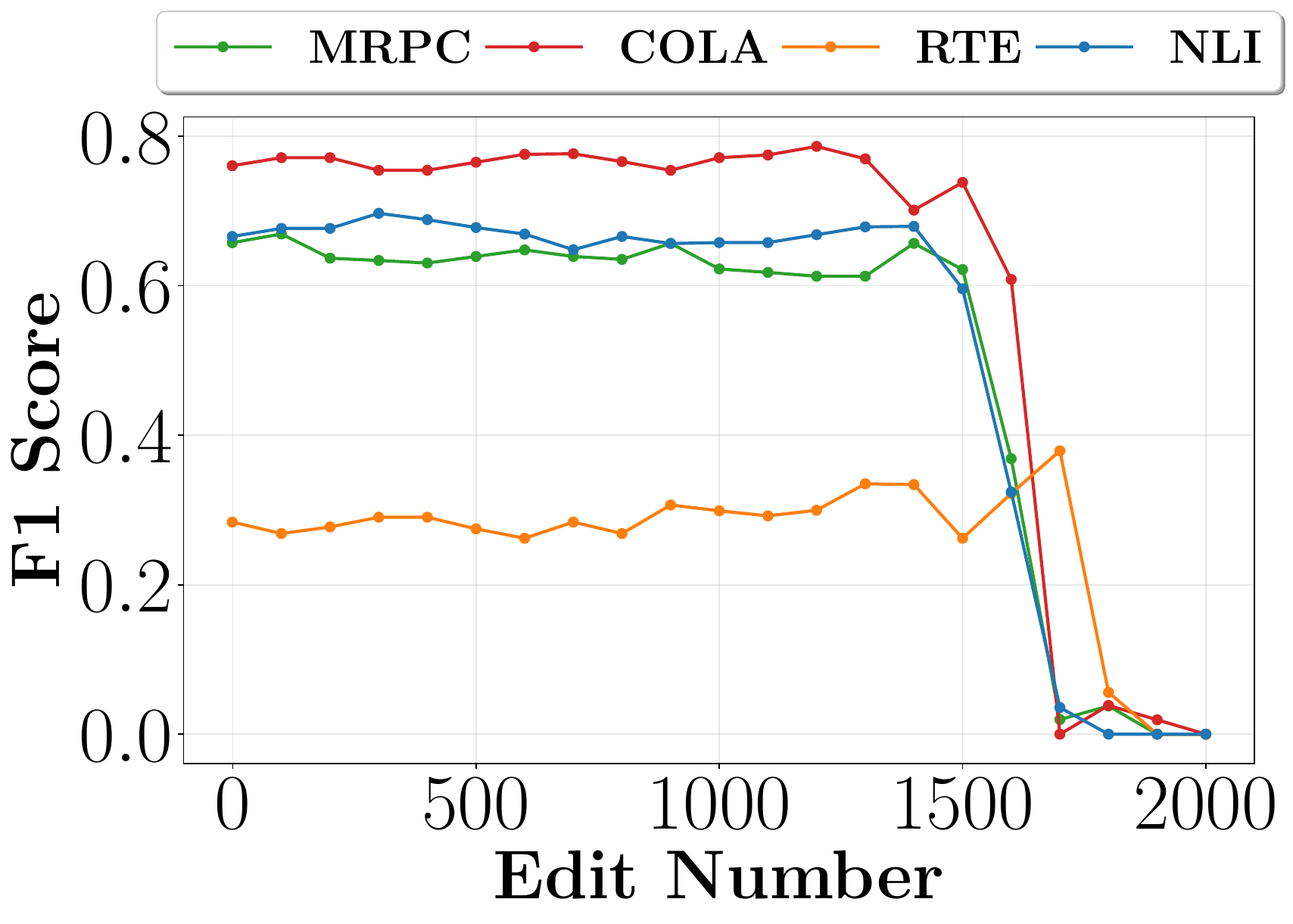}
        \caption{General Ability Performance on GLUE vs. Edit Number.}
        \label{fig:general}
    \end{minipage}
\end{figure}

Having established that general abilities are concentrated in a fragile dominant singular subspace, we now analyze how these critical subspaces degrade during sequential editing progress. We apply $2,000$ edits from \textsc{Counterfact} dataset to LLaMA3 using \textsc{MEMIT}, organized into $20$ rounds with $100$ edits per round. After each editing round, we evaluate (i) editing performance using Efficacy Score and Paraphrase Score, (ii) general abilities using the GLUE benchmark, and (iii) the spectral metrics introduced below. This allows us to analyze the relationship between behavioral degradation and structural changes in model parameters. We conduct the same analytical experiments on AlphaEdit; due to space constraints, these results are reported in Appendix~\ref{app:specalphaedit}. 

\paragraph{Spectrum-based metrics.}
To characterize the stability of the dominant singular subspace (top 10\% components by singular value energy), we introduce two complementary metrics that capture degradation at different granularities.
\begin{itemize}[leftmargin=*]
    \item \textbf{Low-rank Subspace Similarity (LS)} measures the macroscopic drift of the dominant singular subspace. It is defined as the Frobenius cosine similarity between the low-rank reconstructions of the original matrix $\mathbf{\hat{W}}_0$ and the edited matrix $\mathbf{\hat{W}}_t$ at editing round $t$:
\begin{equation}
    \text{LS}_{t} = \tfrac{{\langle \mathbf{\hat{W}}_t, \mathbf{\hat{W}}_0 \rangle}_F}{\| \mathbf{\hat{W}}_t \|_F \cdot \| \mathbf{\hat{W}}_0 \|_F}.
\end{equation}
\item \textbf{Singular Vector Similarity (SS)} provides a microscopic view by quantifying how individual dominant singular vectors rotate during editing. We compute the cosine similarity between a dominant singular vector $\mathbf{v}_t$ of the edited matrix $\mathbf{W}_t$ and each original singular vector $\mathbf{v}_j$ of $\mathbf{W}_0$ (a vector basis): $\text{SS}_t^j = \langle \mathbf{v}_{t}, \mathbf{v}_j \rangle.$ 
Results for left singular vectors show the same trend and are reported in Appendix~\ref{leftss}.
\end{itemize}

\noindent \textbf{Finding: Sequential editing progressively degrades the dominant singular subspace, accompanying model collapse.} This finding is supported by consistent evidence across behavioral, macroscopic, and microscopic analyses. At the \emph{behavioral level}, both editing performance and general ability remain stable in early rounds but deteriorate rapidly after round~$10$, collapsing almost entirely by round $20$ (Figure~\ref{fig:eff-vs-run} and~\ref{fig:general}). At the \emph{macroscopic level}, this behavioral degradation is closely mirrored by a marked decline in the Low-rank Subspace Similarity (LS) metric (Figure~\ref{fig:ss_combined}). LS remains near its initial value in early rounds, indicating that the dominant singular subspace is largely preserved, before drifting and dropping sharply after roughly round~$15$. At the \emph{microscopic level}, the Singular Vector Similarity (SS) metric (Figure~\ref{fig:ls_vs_rounds}) reveals that individual dominant singular directions gradually rotate away from their original directions, becoming nearly orthogonal by round $20$. This gradual rotation reflects a systematic corruption of the learned input--output mappings represented by the dominant singular subspace. Taken together, these observations provide strong empirical evidence that collapse under sequential editing is structurally associated with the progressive degradation of the dominant singular subspace.

\section{Methodology}
\label{method}

Our analysis in Section~\ref{sec:analysis} show that collapse under sequential editing is closely associated with the cumulative degradation of the dominant singular subspace, which is strongly linked to a model’s general performance. This motivates explicitly preserving this subspace to stabilize long-horizon sequential editing.

To this end, we introduce a \textbf{\underline{R}}obust s\textbf{\underline{E}}quential editing \textbf{\underline{V}}ia dom\textbf{\underline{I}}nant subspace preser\textbf{\underline{V}}ation fram\textbf{\underline{E}}work (\themodel), a plug-and-play framework that operates directly on parameter updates produced by existing editing methods.
Rather than modifying editing objectives or imposing method-specific constraints, \themodel analyzes each update in the singular-vector basis of the original weight matrix, providing a unified spectral coordinate system for characterizing how edits interact with intrinsic functional directions. Within this representation, \themodel identifies update components that interfere with dominant singular directions and removes them before applying the update. An overview of the framework is shown in Figure~\ref{fig:method}, and the complete algorithm is provided in Appendix~\ref{alg:dsp}.



\subsection{Spectral Decomposition of Parameter Updates}
\label{method:svd-align}

To explicitly characterize how a parameter update interacts with different spectral components of the original weight matrix, \themodel analyzes parameter updates through spectral decomposition. 

Recall from Section~\ref{anal:spect} that the SVD of the original weight matrix $\mathbf{W}$ defines sets of left and right singular vectors $\{\mathbf{u}_i\}$ and $\{\mathbf{v}_j\}$. The rank-one outer products $\{\mathbf{u}_i \mathbf{v}_j^\top\}_{i,j}$ constitute an orthogonal basis for the matrix space $\mathbb{R}^{m \times n}$; a formal proof is provided in Appendix~\ref{app:basisproof}. Using this basis, any update matrix $\Delta \mathbf{W}$ produced by an editing method can be expressed as
\begin{equation}
\Delta \mathbf{W}
= \sum_{i=1}^{m} \sum_{j=1}^{n} \alpha_{ij} \, \mathbf{u}_i \mathbf{v}_j^\top ,
\end{equation}
where the coefficients are given by $\alpha_{ij}
= \langle \Delta \mathbf{W}, \mathbf{u}_i \mathbf{v}_j^\top \rangle_F$.
This spectral decomposition resolves a parameter update into its effects on individual input--output mappings of the original weight matrix. Each coefficient $\alpha_{ij}$ quantifies the contribution of the update along the mapping from input singular direction $\mathbf{v}_j$ to output singular direction $\mathbf{u}_i$. As a result, update components aligned with dominant spectral directions can be distinguished from those operating in lower-energy regions.


\subsection{Dominant Subspace Protection}
\label{method:DSP}
Building on the spectral decomposition above, \themodel introduces a {Dominant Subspace Protection} (DSP) mechanism that constrains parameter updates to avoid interference with the dominant singular subspace. DSP consists of two components:  Dominant Subspace Identification and Safe Update Construction. 

\begin{figure}
\includegraphics[width=\linewidth]{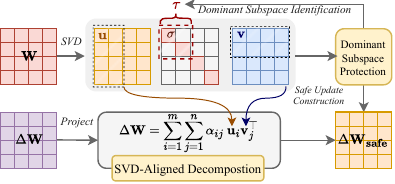}
\caption{Overview of the \themodel. An update matrix $\Delta\mathbf{W}$ is represented in the singular vector basis of the original weight matrix. The dominant subspace, identified by an energy threshold $\tau$, is then used to filter out update components that would interfere with this protected region. The resulting safe update $\Delta\mathbf{W}_\text{safe}$ preserves the dominant subspace while allowing effective knowledge modification.}
\label{fig:method}
\vspace{-1.em}
\end{figure}

\paragraph{\textbf Dominant Subspace Identification.} To identify the critical components for preservation, we adopt an energy-based criterion. Given a \textbf{singular-value energy threshold} $\tau \in (0,1)$ (analyzed in Section~\ref{hyper}), we select the smallest index $k$ such that the cumulative energy of the top-$k$ singular values exceeds $\tau$:
\begin{equation}
\frac{\sum_{i=1}^{k} \sigma_i}{\sum_{i=1}^{r} \sigma_i} \geq \tau.
\end{equation}
The corresponding singular vectors $\{\mathbf{u}_i\}_{i=1}^{k}$ and  $\{\mathbf{v}_i\}_{i=1}^{k}$ span the dominant input and output subspaces, respectively.

\paragraph{\textbf Safe Update Construction.} Once the dominant subspace is identified, we construct a {safe update} by removing update components associated with this protected region. Concretely, the coefficient $\alpha_{ij}$ is set to zero whenever the corresponding input direction $\mathbf{v}_j$ or output direction $\mathbf{u_i}$ lies in the dominant subspace, i.e., $\alpha_{ij}$  $=$ $0$ if  $i \leq k \text{ or } j \leq k$. The resulting safe update is given by
\begin{equation}
\Delta \mathbf{W}_{\text{safe}} 
= \sum_{i > k} \sum_{j > k} \alpha_{ij} \, \mathbf{u}_i \mathbf{v}_j^\top.
\end{equation}
By construction, $\Delta \mathbf{W}_{\text{safe}}$ operates exclusively in lower-energy spectral directions, thereby limiting interference with dominant components that are critical for preserving general abilities.

\begin{figure*}[t]
  \begin{subfigure}{0.27\textwidth}
    \includegraphics[width=\linewidth]{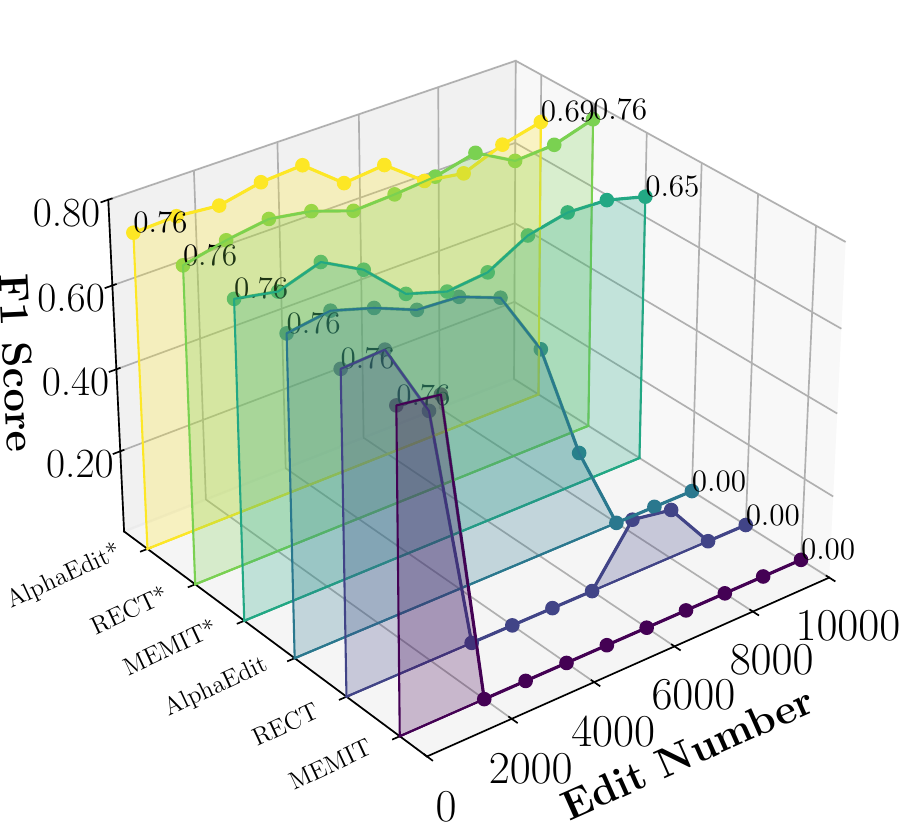}
    \caption{COLA}
  \end{subfigure}\hfill
  \begin{subfigure}{0.27\textwidth}
    \includegraphics[width=\linewidth]{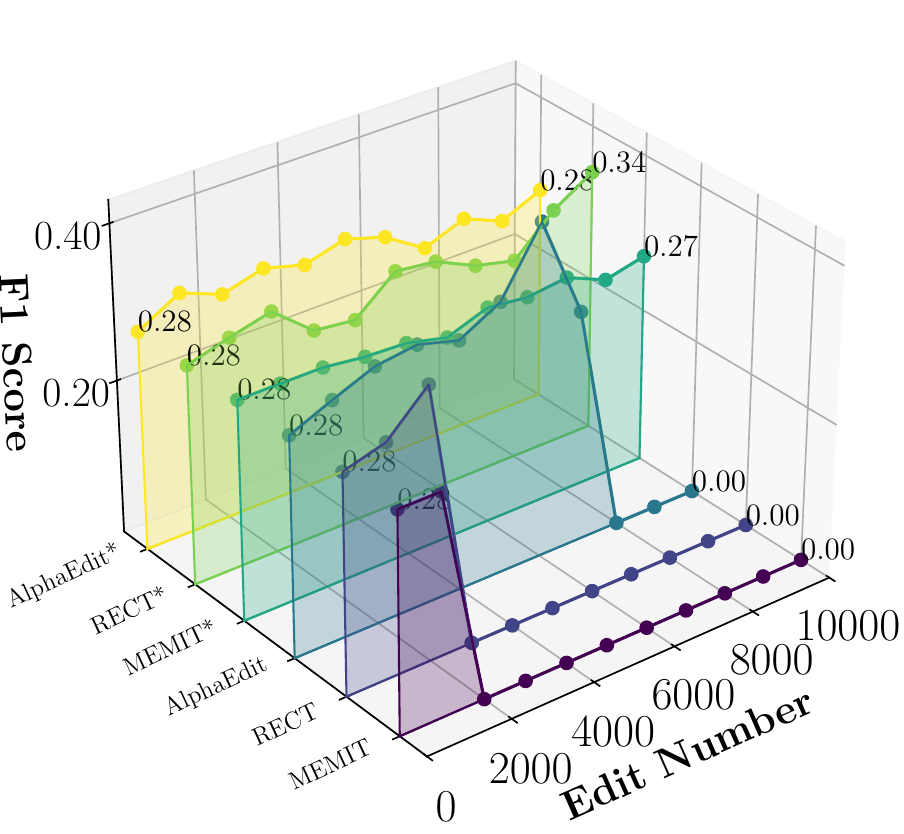}
    \caption{RTE}
  \end{subfigure}\hfill
  \begin{subfigure}{0.27\textwidth}
    \includegraphics[width=\linewidth]{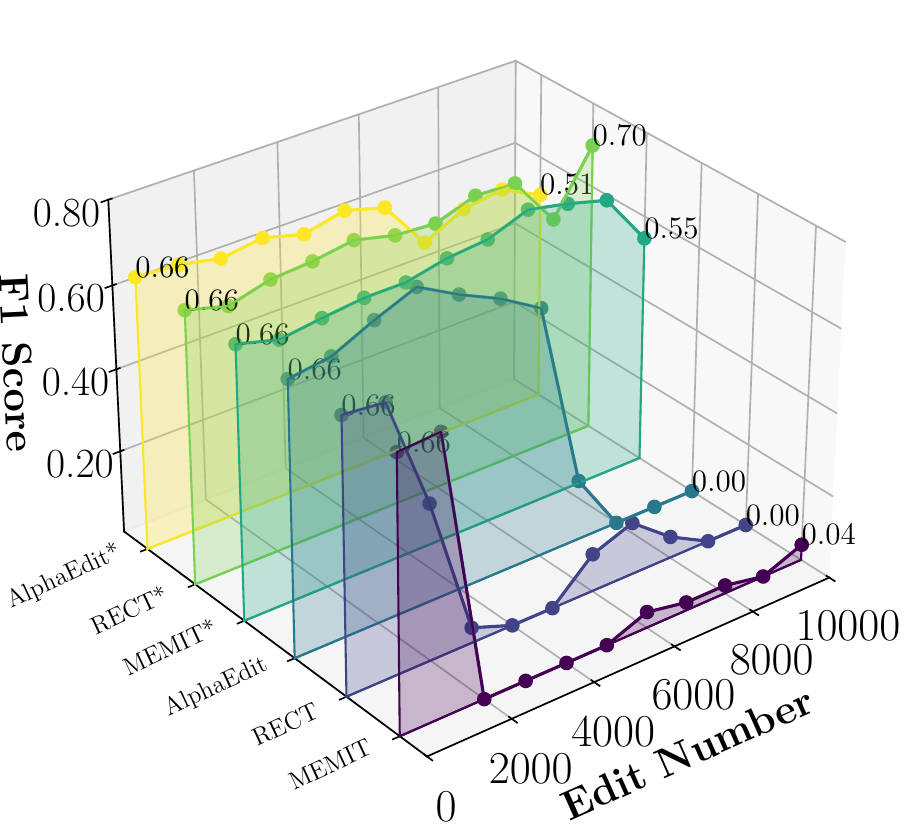}
    \caption{MRPC}
  \end{subfigure}
  \caption{Performance of baselines and their REVIVE-enhanced versions (*) on GLUE.}
  \label{fig:three-by-two-3d}
  \vspace{-1.1em}
\end{figure*}

Importantly, the protection mechanism of \themodel operates as a plug-and-play wrapper that can be applied to existing parameter-modifying knowledge editing methods, without altering how edits are generated.


\section{Experiments}

\begin{table*}[t]
\centering
\caption{\footnotesize 
Performance on sequential editing over 10,000 Samples. 
The abbreviations \textit{Eff.} (Efficacy Success), \textit{Para.} (Paraphrase Success), 
\textit{Neigh.} (Neighborhood Success), \textit{Flu.} (Generation Entropy), and 
\textit{Consis.} (Reference Score) denote respective evaluation metrics. 
Relative improvements (\%) are shown in \blue{blue} and decreases in \orange{orange}. 
 $\uarbig$ indicates a large improvement where the baseline score was near zero.
}\label{tab:table1}
\fontsize{8pt}{10pt}\selectfont

\setlength{\tabcolsep}{1pt} 
\renewcommand{\arraystretch}{1} 

\resizebox{1\textwidth}{!}{%
\begin{tabular}{c|ccccc|ccc}
\toprule[1.5pt]
\textbf{Method}  & \multicolumn{5}{c|}{\textbf{Counterfact}} & \multicolumn{3}{c}{\textbf{ZsRE}} \\
\cmidrule(lr){2-6} \cmidrule(lr){7-9}
& \textbf{Eff.$\uparrow$} & \textbf{Para.$\uparrow$} & \textbf{Neigh.$\uparrow$} & \textbf{Flu.$\uparrow$} & \textbf{Consis.$\uparrow$} & \textbf{Eff.$\uparrow$} & \textbf{Para.$\uparrow$} & \textbf{Neigh.$\uparrow$} \\
\cmidrule{1-9}

\textbf{LLaMA3} & 7.02  & 9.44 & 89.73 & 635.47 & 24.24 & 35.67 & 34.81 & 31.83 \\
\cmidrule{1-9}
MEMIT & 62.3 & 55.02 & 48.11 & 522.1 & 4.4 & 0.08 & 0.08 & 1.36 \\
+\themodel & 95.62 $\uar{53.5\%}$ & 84.60 $\uar{53.8\%}$ & 62.17 $\uar{29.2\%}$ & 603.22 $\uar{15.5\%}$ & 29.39 $\uar{568.0\%}$ & 83.45 $\uarbig$ & 79.90 $\uarbig$ & 32.01 $\uarbig$ \\
\cmidrule{1-9}
PRUNE & 59.98 & 55.72 & 48.56 & 571.27 & 1.89 & 0.00 & 0.00 & 0.08 \\
+\themodel & 80.57 $\uar{34.3\%}$ & 69.54 $\uar{24.7\%}$ & 54.76 $\uar{12.8\%}$ & 570.85 $\dar{0.1\%}$ & 28.49 $\uarbig$ & 56.61 $\uarbig$ & 53.30 $\uarbig$ & 27.74 $\uarbig$ \\
\cmidrule{1-9}
RECT & 60.23 & 54.9 & 50.56 & 441.61 & 5.08 & 0.00 & 0.00 & 0.00 \\
+\themodel & 92.69 $\uar{53.9\%}$ & 79.95 $\uar{45.6\%}$ & 63.09 $\uar{24.8\%}$ & 600.13 $\uar{35.9\%}$ & 29.28 $\uar{476.8\%}$ & 84.20 $\uarbig$ & 80.27 $\uarbig$ & 31.92 $\uarbig$ \\
\cmidrule{1-9}
AlphaEdit & 62.48 & 56.9 & 52.31 & 505.5 & 4.25 & 90.57 & 85.66 & 30.5 \\
+\themodel & 98.74 $\uar{58.0\%}$ & 90.08 $\uar{58.4\%}$ & 60.19 $\uar{15.1\%}$ & 615.97 $\uar{21.9\%}$ & 32.66 $\uar{668.5\%}$ & 93.40 $\uar{3.1\%}$ & 89.31 $\uar{4.3\%}$ & 31.72 $\uar{4.0\%}$ \\
\cmidrule{1-9}
NSE & 77.59 & 44.42 & 86.12 & 607.86 & 23.31 & 45.61 & 45.04 & 31.27 \\
+\themodel & 98.89 $\uar{27.4\%}$ & 92.28 $\uar{107.8\%}$ & 65.72 $\dar{23.6\%}$ & 618.66 $\uar{1.8\%}$ & 32.74 $\uar{40.5\%}$ & 94.37 $\uar{107.0\%}$ & 90.57 $\uar{101.2\%}$ & 32.17 $\uar{2.9\%}$ \\

\cmidrule[1pt]{1-9}
\textbf{GPT-J} & 15.22 & 17.65 & 83.50 & 622.01 & 29.61 & 26.45 & 25.74 & 27.04 \\
\cmidrule{1-9}
MEMIT & 54.03 & 52.66 & 53.63 & 594.16 & 5.17 & 0.10 & 0.10 & 0.17 \\
+\themodel & 97.63 $\uar{80.7\%}$ & 87.76 $\uar{66.6\%}$ & 66.52 $\uar{24.1\%}$ & 616.47 $\uar{3.8\%}$ & 40.69 $\uar{687.4\%}$ & 88.88 $\uarbig$ & 83.22 $\uarbig$ & 27.87 $\uarbig$ \\
\cmidrule{1-9}
PRUNE & 52.92 & 51.47 & 53.91 & 576.95 & 5.14 & 0.03 & 0.02 & 0.05 \\
+\themodel & 86.95 $\uar{64.3\%}$ & 81.03 $\uar{57.5\%}$ & 64.21 $\uar{19.1\%}$ & 583.05 $\uar{1.1\%}$ & 35.73 $\uar{595.7\%}$ & 63.08 $\uarbig$ & 58.90 $\uarbig$ & 26.03 $\uarbig$ \\
\cmidrule{1-9}
RECT & 63.60 & 55.33 & 56.69 & 404.13 & 4.49 & 23.60 & 22.02 & 12.44 \\
+\themodel & 94.96 $\uar{49.4\%}$ & 77.27 $\uar{39.6\%}$ & 67.78 $\uar{19.6\%}$ & 612.76 $\uar{51.62\%}$ & 38.69 $\uar{761.7\%}$ & 81.28 $\uar{244.4\%}$ & 74.78 $\uar{239.6\%}$ & 28.20 $\uar{126.7\%}$ \\
\cmidrule{1-9}
AlphaEdit & 96.51 & 86.76 & 60.80 & 544.18 & 19.33 & 87.84 & 78.65 & 22.31 \\
+\themodel & 99.50 $\uar{3.1\%}$ & 93.92 $\uar{8.3\%}$ & 67.35 $\uar{10.8\%}$ & 600.64 $\uar{10.4\%}$ & 40.63 $\uar{110.3\%}$ & 97.53 $\uar{11.0\%}$ & 91.33 $\uar{16.1\%}$ & 23.40 $\uar{4.9\%}$ \\
\cmidrule{1-9}
NSE & 88.95 & 69.69 & 75.46 & 611.35 & 33.31 & 44.03 & 42.39 & 24.86 \\
+\themodel & 
94.88 $\uar{6.7\%}$ & 
89.49 $\uar{28.4\%}$ & 
64.06 $\dar{15.1\%}$ & 
608.12 $\dar{0.5\%}$ & 
40.18 $\uar{20.6\%}$ & 
97.80 $\uar{122.1\%}$ & 
91.75 $\uar{116.4\%}$ & 
26.84 $\uar{8.0\%}$ \\

\bottomrule[1.5pt]
\end{tabular}}
\end{table*}
\subsection{Experiment Setup}
\textbf{Base Models.} We conduct experiments on three LLMs that are widely used in prior knowledge editing work: GPT2-XL (1.5B) \citep{gpt2}, GPT-J (6B) \citep{gptj}, and LLaMA3 (8B) \citep{llama3}. Due to space constraints, results on GPT2-XL are reported in Appendix \ref{app:gpt2-xl}. 

\paragraph{\textbf{Baselines.}} We compare \themodel against a suite of strong baselines, including the canonical \textsc{MEMIT} \citep{memit} , as well as four state-of-the-art methods designed for sequential editing: \textsc{PRUNE} \citep{prune}, \textsc{RECT} \citep{rect}, \textsc{AlphaEdit} \citep{alphaedit},  \textsc{DeltaEdit} \citep{deltaedit}, and \textsc{NSE} \citep{nsee}. Due to differences in experimental settings, comparison with DeltaEdit can be found in Appendix \ref{app:deltacomp}. Additional implementation details and baseline comparisons are provided in Appendix~\ref{app:baselines}.

\paragraph{\textbf{Datasets and Metrics.}} We evaluate all methods on two standard factual knowledge editing benchmarks, \textsc{Counterfact} \citep{memit} and \textsc{ZsRE} \citep{zsre}. For \textsc{ZsRE}, we report Efficacy, Paraphrase, and Neighborhood Scores. For \textsc{Counterfact}, we additionally report Fluency and Consistency metrics. Detailed definitions of all datasets and evaluation metrics are provided in Appendix~\ref{app:dataset}, \ref{app:zsreme}, and \ref{app:ctfme}. 

\subsection{Results and Analysis}
\label{resultandnalysis}
This section presents a comprehensive evaluation of \themodel. We first assess its performance under extended sequential editing, measuring both editing success and the preservation of general abilities. We then analyze robustness properties, including scalability to $20{,}000$ edits, sensitivity to hyperparameters, the stability of internal representations using t-SNE visualizations, and matrix norm growth. Additionally, Appendix~\ref{app:specalphaedit_revive} provides a direct verification that \themodel preserves the dominant singular directions during long editing sequences. Further efficiency analyses are reported in Appendix~\ref{app:timeanalysis}. 

\paragraph{Sequential Editing Performance.}
We evaluate the effectiveness of \themodel over an extended sequence of $10{,}000$ edits, applied in $100$ rounds of $100$ edits each, on the \textsc{CounterFact} and \textsc{ZsRE} benchmarks. As shown in Table \ref{tab:table1}, incorporating \themodel as a plug-and-play module consistently improves performance across all base methods and models. The improvements are particularly pronounced on the more challenging \textsc{ZsRE} benchmark, where several baselines, including \textsc{MEMIT} and \textsc{RECT}, degrade rapidly when applied sequentially without protection. When augmented with \themodel, these methods remain stable and achieve substantially higher efficacy (e.g., $83.45\%$ for \textsc{MEMIT}+\themodel on LLaMA3). Notably, the standard \textsc{MEMIT}+\themodel combination consistently outperforms specialized sequential baselines such as \textsc{PRUNE} and \textsc{RECT}, suggesting that proactively constraining harmful update directions is more effective than post-hoc regularization. For certain methods such as \textsc{NSE}, applying \themodel leads to a moderate decrease in Neighborhood Success. This behavior is likely attributable to the fact that high neighborhood scores can arise trivially when edits fail to modify the model. In contrast, \themodel achieves large gains in Efficacy and Paraphrase while maintaining strong Neighborhood performance, reflecting a more meaningful form of successful editing.

\paragraph{Preservation of General Abilities.}
To evaluate the preservation of general abilities, we assess edited LLaMA3 models on the GLUE benchmark after $10{,}000$ sequential edits. For clarity, Figure~\ref{fig:three-by-two-3d} reports results on three representative tasks, with full results provided in Appendix~\ref{app:gluefull}. All tasks exhibit consistent trends. As shown in Figure \ref{fig:three-by-two-3d}, baseline methods without protection experience rapid degradation in general performance. \textsc{MEMIT} and \textsc{RECT} collapse to near-zero performance after only $3,000$ edits, and even the more robust \textsc{AlphaEdit} eventually suffers a complete collapse after $8,000$ edits. In contrast, \themodel enhanced methods maintains an overall average ${86.34\%}$ of its performance across all tasks after $10,000$ edits. These results clearly demonstrate that shielding the dominant singular subspace is a highly effective strategy for preserving a model's general abilities during sequential editing.

\begin{figure*}[t] \label{threshfigure}
    \centering
    \includegraphics[width=\linewidth]{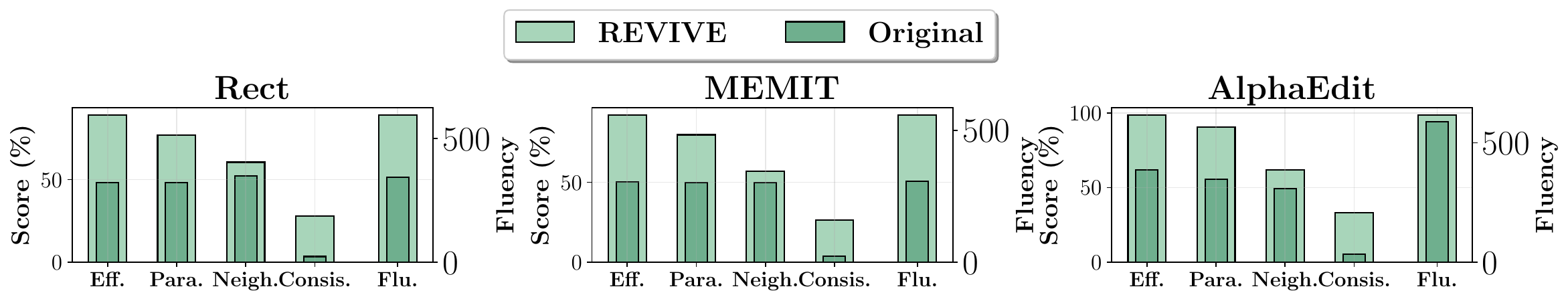}
    \caption{REVIVE vs. original methods under 20,000 sequential edits on \textsc{Counterfact}.}
    \label{fig:20000figure}
\end{figure*}

\paragraph{Scalability under Extreme Sequential Editing.}
\label{app:extreme20000}
We further evaluate the scalability of \themodel under substantially extended sequential editing on LLaMA3. Specifically, we apply $20{,}000$ sequential edits on \textsc{CounterFact}, organized into $200$ rounds of $100$ edits each, as well as the full set of $19{,}086$ edits on \textsc{ZsRE}. As shown in Figure~\ref{fig:20000figure}, REVIVE continues to deliver substantial gains over the original base methods. On \textsc{CounterFact}, \themodel achieves average improvements of $+75.1\%$ in Efficacy and $+53.1\%$ in Fluency. Complete results for all metrics and methods, provided in Appendix~\ref{app:20000editsresults}. These results indicate that \themodel effectively maintains editing performance even when the number of edits is significantly scaled up.

\begin{figure}[thbp]
    \centering
    \begin{minipage}[t]{0.24\textwidth}
        \centering
        \includegraphics[width=1\textwidth]{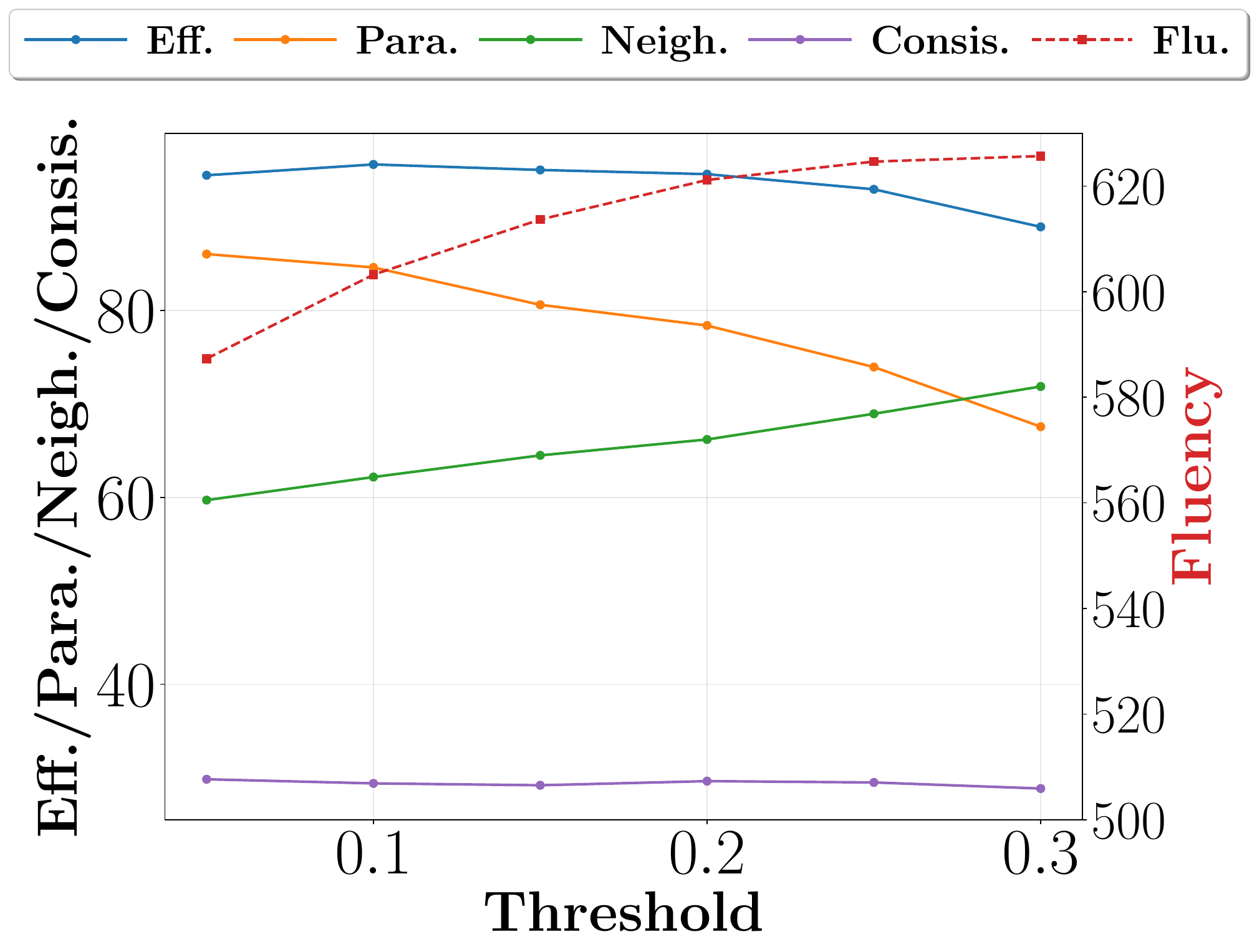}
        \caption{Performance of MEMIT-REVIVE on LLaMA3 (CounterFact) under different thresholds.}
        \label{fig:sensitivity}
    \end{minipage}
    \hfill
    \begin{minipage}[t]{0.23\textwidth}
        \centering
        \includegraphics[width=0.95\textwidth]{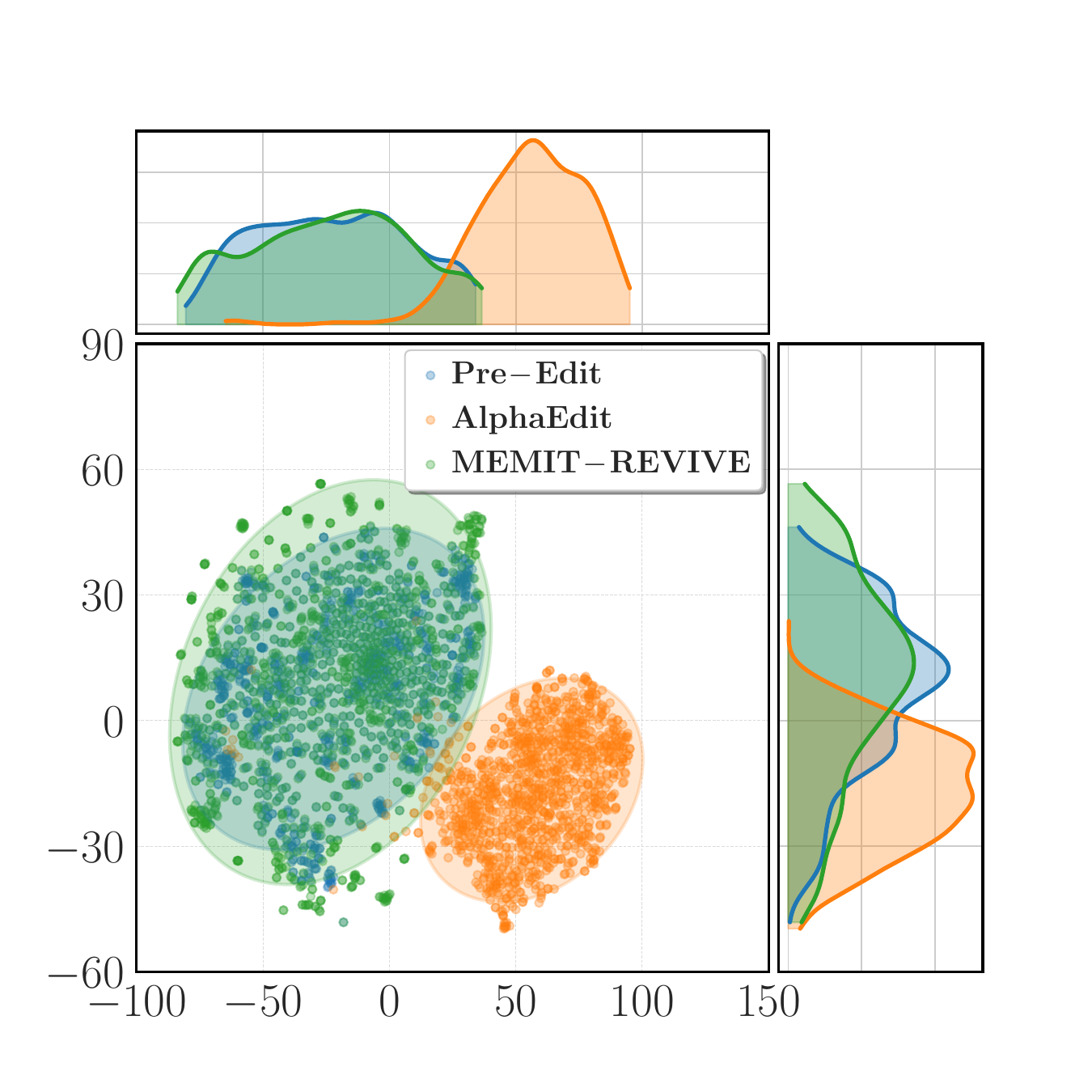}
        \caption{t-SNE visualization of LLaMA3 representations before and after $20{,}000$ sequential edits.}
        \label{fig:tal}
    \end{minipage}
 
\end{figure}

\paragraph{Sensitivity Analysis.}\label{hyper}
We evaluate the robustness of \themodel with respect to its only intrinsic hyperparameter, the singular value energy threshold $\tau$, defined in Section~\ref{method:DSP}. This threshold controls the size of the dominant singular subspace that is protected from modification: larger values of $\tau$ emphasize stronger preservation of general abilities, while smaller values permit more flexible updates by exposing a larger portion of the parameter space to editing.
Figure \ref{fig:sensitivity} shows that \themodel exhibits strong robustness, maintaining high performance across a wide range of $\tau$ values. This behavior indicates that the effectiveness of \themodel is not sensitive to the precise choice of the dominant subspace boundary, and that stable performance can be achieved without careful hyperparameter tuning. Further hyperparameter results for all models are reported in Appendix~\ref{app:threshold}.

\paragraph{Visualization of Representational Stability.}
To qualitatively examine how extended sequential editing affects internal representations, we apply t-SNE~\citep{tsne} to visualize the representations of 1,000 factual prompts from LLaMA3, both before and after applying 20,000 edits.  As illustrated in Figure~\ref{fig:tal}, a strong baseline like \textsc{AlphaEdit} causes a noticeable distributional shift, where post-edit representations drift away from their original positions. In contrast, when combined with \themodel, \textsc{MEMIT} exhibits considerably smaller shifts, with post-edit representations remaining closely aligned with their original clusters. This visualization is consistent with our quantitative findings and suggests that preserving the dominant singular subspace helps maintain the overall representational geometry of the model under extended sequential editing.

\paragraph{Matrix Norm Growth Analysis}
Prior studies \citep{norm1,xu2025constraining} have consistently observed that parameter matrix norms tend to exhibit abnormal growth during sequential editing, a phenomenon often linked to model instability and catastrophic collapse. To verify whether preserving the dominant singular subspace inherently mitigates this issue, we examine the evolution of weight magnitudes during long-horizon editing.

Specifically, we track the $L_1$ and $L_2$ norms of the parameter matrix at Layer 3 of the GPT-J (6B) model over a sequence of 10,000 edits on the \textsc{CounterFact}, utilizing MEMIT as the base editor. The results are summarized in Table~\ref{tab:norm_growth}.
As edits accumulate, the unprotected baseline exhibits severe norm inflation, with the $L_2$ norm surging from 105.51 to 20946.66. In contrast, REVIVE substantially constrains and stabilizes this growth. This empirical evidence confirms that by addressing the interference in the dominant subspace, REVIVE naturally suppresses the abnormal weight inflation associated with long-horizon degradation.

\begin{table}[htbp]
\centering
\small
\setlength{\tabcolsep}{4.5pt}

\caption{Evolution of parameter matrix norms in GPT-J (Layer 3) over $10,000$ sequential edits.}
\label{tab:norm_growth}

\begin{tabular}{lcccc}
\toprule
\multirow{2}{*}{\textbf{Edits}} & \multicolumn{2}{c}{\textbf{$L_1\ Norm$}} & \multicolumn{2}{c}{\textbf{$L_2\ Norm$}} \\
\cmidrule(lr){2-3} \cmidrule(lr){4-5}
      & \textbf{MEMIT} & \textbf{+REVIVE} & \textbf{MEMIT} & \textbf{+REVIVE} \\
\midrule
0     & $6.88 \times 10^5$ & $6.88 \times 10^5$ & 105.51   & 105.51 \\
2000  & $1.37 \times 10^6$ & $7.47 \times 10^5$ & 5590.36  & 153.57 \\
4000  & $1.75 \times 10^6$ & $7.71 \times 10^5$ & 8649.95  & 156.45 \\
6000  & $2.29 \times 10^6$ & $7.96 \times 10^5$ & 13012.32 & 159.32 \\
8000  & $2.51 \times 10^6$ & $8.13 \times 10^5$ & 13871.72 & 161.41 \\
10000 & $3.92 \times 10^6$ & $8.30 \times 10^5$ & 20946.66 & 163.47 \\
\bottomrule
\end{tabular}
\end{table}

\section{Related Work}
Our work focuses on \emph{parameter-modifying methods} for knowledge editing. Early methods such as \textsc{MEND} and \textsc{MEMIT}~\citep{mend,memit} demonstrate strong performance for isolated edits, but often degrade under sequential editing due to the accumulation of interfering parameter updates. 

To mitigate interference across edits, several sequential editing methods introduce mechanisms that restrict or regularize parameter updates during editing~\citep{alphaedit,prune,rect,deltaedit}. A common strategy among these approaches is to construct \emph{protection subspaces} that constrain how updates are applied. Notably, \textbf{these subspaces are typically derived from empirical statistics over {external data}, such as covariance estimates computed from sampled factual triples~\citep{alphaedit}, or from {historical editing directions} accumulated during previous updates~\citep{deltaedit}.} While effective in reducing empirically observed conflicts, such approaches are largely driven by external data distributions or past editing trajectories, and are not explicitly grounded in the intrinsic organization of the model’s parameter space.

By contrast, our approach directly analyzes the spectral structure of the model’s weight matrices, identifies degradation of dominant functional subspaces as the primary failure mechanism under sequential editing, and introduces a plug-and-play intervention to preserve these subspaces during updates. A more comprehensive discussion of related work, including detailed comparisons with SVD-based editing methods, is provided in Appendix~\ref{app:fullrelatedwork} and Appendix~\ref{app:svd_comparison}.



\section{Conclusion}
In this work, we investigated the mechanisms behind model collapse in sequential knowledge editing from a spectral perspective. Our analysis reveals that repeated parameter updates progressively distort dominant singular directions of pretrained weight matrices, which are crucial for maintaining a model’s general abilities. Based on this insight, we propose \themodel, a plug-and-play framework that protects this intrinsic structure by filtering update components that interfere with dominant singular directions. Extensive experiments across multiple models and benchmarks show that \themodel substantially improves editing performance while preserving general abilities, even under large numbers of sequential edits. Overall, our work provides a structural explanation for sequential editing failure and offer a principled approach for maintaining the long-term stability of edited LLMs.

\section*{Acknowledgements}
This work was supported by the National Natural Science Foundation of China (Nos.~62502286, 62472261, 62576339, 62372275). It was also Funded by Basic Research Program of Jiangsu (BK20250429), the Shandong Provincial Natural Science Foundation (ZR2024QF203), the Technology Innovation Guidance Program of Shandong Province (YDZX2024088), and the Provincial Key R\&D Program of Shandong Province (Nos.~2024CXGC010108, 2025CXGC010112).


\section*{Limitations}

While our results demonstrate the effectiveness of preserving the dominant singular subspace for stabilizing sequential knowledge editing, several limitations remain and suggest directions for future work.

First, the dominant subspace in \themodel is identified using a singular-value energy criterion. While this choice is empirically stable and effective across models and tasks, it is not theoretically optimal in a strict sense. Exploring alternative criteria for defining task-relevant or functionally critical spectral components remains an open direction.


Second, our analysis and experiments primarily focus on the feed-forward network (FFN) layers, which are known to play a central role in factual knowledge storage. However, other architectural components, such as attention mechanisms, may also contribute to knowledge representation and could exhibit different spectral behaviors under sequential editing. Extending our spectral analysis and protection strategy to these components remains an important avenue for future work.

Third, while recent inspiring works~\citep{liu2025model,liu2026we} have questioned the adequacy of existing evaluation protocols, this study primarily focuses on mitigating model collapse. Consequently, we did not benchmark our method against these emerging metrics at this stage. We are committed to investigating these evaluation-related issues in our subsequent research.

\section*{Usage of large language models} 
In the preparation of this work, large language models (LLMs) have been utilized to assist in several stages of writing. In particular, LLMs played a significant role in polishing the manuscript by improving readability and correcting grammatical issues.  Moreover, they provided valuable assistance in certain aspects of data visualization, such as generating and refining plotting scripts, which streamlined the experimental analysis process. 
\section*{Ethics Statement}
This work investigates knowledge editing techniques for large language models to enable targeted updates of factual behaviors. All experiments are conducted on publicly available datasets and do not involve the collection of personal data or interaction with human subjects. We emphasize that knowledge editing methods should be applied with caution, as they could potentially be misused to inject harmful, misleading, or inappropriate information into model parameters. Our approach is intended solely for controlled research settings, such as correcting erroneous knowledge and analyzing model behavior, and should not be deployed without appropriate safeguards and oversight. We acknowledge that edited models may still inherit biases present in the original training data.


\bibliography{custom}

\appendix
\newpage

\section{Algorithm Details}\label{alg:dsp}
\begin{algorithm}[H]
\caption{REVIVE}

\begin{algorithmic}[1]
\Require Current weight matrix $\mathbf{W} \in \mathbb{R}^{m \times n}$; 
         update matrix $\Delta \mathbf{W}$; 
         singular-value energy threshold $\tau \in (0,1)$
\Ensure Safe update $\Delta \mathbf{W}_{\text{safe}}$

\State \textbf{SVD-Aligned Decomposition:} 
\State  $\{\mathbf{u}_i\}_{i=1}^{m}$, $\{\sigma_i\}_{i=1}^r$, $\{\mathbf{v}_i\}_{i=1}^n$ = SVD($\mathbf{W}$)
\State Construct orthogonal basis 
       $\{\mathbf{u}_i \mathbf{v}_j^\top \mid i=1,\ldots,m;\, j=1,\ldots,n\}$
\For{$i = 1$ to $m$}
    \For{$j = 1$ to $n$}
        \State $\alpha_{ij} \gets \langle \Delta \mathbf{W}, \mathbf{u}_i \mathbf{v}_j^\top \rangle_F$
    \EndFor
\EndFor
\State Represent update as 
       $\Delta \mathbf{W} = \sum_{i=1}^{m} \sum_{j=1}^{n} \alpha_{ij} \mathbf{u}_i \mathbf{v}_j^\top$

\State \textbf{Dominant Subspace Identification:}
\State Find smallest $k$ s.t.\ $\frac{\sum_{i=1}^k \sigma_i}{\sum_{i=1}^r \sigma_i} \geq \tau$
\State Define dominant subspace 
       $\{\mathbf{u}_1,\ldots,\mathbf{u}_k\}$, 
       $\{\mathbf{v}_1,\ldots,\mathbf{v}_k\}$

\State \textbf{Safe Update Construction:}
\State Initialize $\Delta \mathbf{W}_{\text{safe}} \gets 0$
\For{$i = 1$ to $m$}
    \For{$j = 1$ to $n$}
        \If{$i > k$ \textbf{and} $j > k$}
            \State $\Delta \mathbf{W}_{\text{safe}} \gets 
                   \Delta \mathbf{W}_{\text{safe}} + \alpha_{ij} \mathbf{u}_i \mathbf{v}_j^\top$
        \EndIf
    \EndFor
\EndFor

\State \Return $\Delta \mathbf{W}_{\text{safe}}$
\end{algorithmic}
\end{algorithm}

\section{Proof of the SVD-aligned Matrix Basis}
\label{app:basisproof}
This section provides the formal proof for the claim made in Section \ref{method:svd-align} that the set of rank-one outer products $\{\mathbf{u}_i\mathbf{v}_j^\top\}_{i,j}$ derived from the singular vectors of a matrix $\mathbf{W}$, forms an orthonormal basis for the space of matrices $\mathbb{R}^{m \times n}$.

\begin{theorem}[Outer-product bases from two orthonormal vector bases]

Let $\{\mathbf{u}_1,\dots,\mathbf{u}_m\}\subset\mathbb{R}^m$ and $\{\mathbf{v}_1,\dots,\mathbf{v}_n\}\subset\mathbb{R}^n$ be orthonormal bases of $\mathbb{R}^m$ and $\mathbb{R}^n$ respectively.  Consider the set of $m n$ matrices
\begin{equation}
\mathcal{B} = \left\{ \mathbf{u}_p \mathbf{v}_q^\top : p=1, \dots, m, \; q=1, \dots, n \right\}.
\end{equation}

Then $\mathcal{B}$ forms an orthonormal basis of the real vector space $\mathbb{R}^{m\times n}$ with respect to the Frobenius inner product $\langle \mathbf{X}, \mathbf{Y} \rangle_F = \mathrm{tr}(\mathbf{X}^\top \mathbf{Y})$. 

In particular, every $\mathbf{X} \in \mathbb{R}^{m \times n}$ admits the unique expansion

\begin{equation}
\mathbf{X} = \sum_{p=1}^m \sum_{q=1}^n c_{pq} \, \mathbf{u}_p \mathbf{v}_q^\top,
\end{equation}

where the coefficients are given by

\begin{equation}
c_{pq} = \langle \mathbf{X}, \mathbf{u}_p \mathbf{v}_q^\top \rangle_F = \mathbf{u}_p^\top \mathbf{X} \mathbf{v}_q.
\end{equation}

\end{theorem}

\begin{proof}

We split the proof into three parts: (i) orthogonality, (ii) spanning (completeness), and (iii) uniqueness / coefficient formula.

\paragraph{(i) Orthogonality.}

Take two generic elements $\mathbf{u}_p \mathbf{v}_q^\top$ and $\mathbf{u}_{p'} \mathbf{v}_{q'}^\top$ from $\mathcal{B}$. Their Frobenius inner product is

\begin{equation}
\begin{aligned}
\langle \mathbf{u}_p \mathbf{v}_q^\top,\,
\mathbf{u}_{p'} \mathbf{v}_{q'}^\top \rangle_F
&= (\mathbf{u}_p^\top \mathbf{u}_{p'})
   (\mathbf{v}_q^\top \mathbf{v}_{q'}).
\end{aligned}
\end{equation}

Since $\{\mathbf{u}_p\}$ and $\{\mathbf{v}_q\}$ are orthonormal bases, we have

\begin{equation}
\mathbf{u}_p^\top \mathbf{u}_{p'} = \delta_{pp'}, \qquad \mathbf{v}_q^\top \mathbf{v}_{q'} = \delta_{qq'},
\end{equation}

where $\delta_{ij}$ is the Kronecker delta:

\begin{equation}
\delta_{ij} = 
\begin{cases}
1, & \text{if } i = j, \\
0, & \text{if } i \neq j.
\end{cases}
\end{equation}

Therefore,

\begin{equation}
\langle \mathbf{u}_p \mathbf{v}_q^\top, \mathbf{u}_{p'} \mathbf{v}_{q'}^\top \rangle_F = \delta_{pp'} \delta_{qq'}.
\end{equation}

In particular, if either $p \neq p'$ or $q \neq q'$, then one of the Kronecker deltas vanishes, making the inner product equal to 0. This proves that distinct basis elements are orthogonal.

\paragraph{(ii) Completeness.}

The vector space $\mathbb{R}^{m \times n}$ has dimension $mn$. We have produced $mn$ elements in $\mathcal{B}$, which are mutually orthonormal. Mutual orthonormality implies linear independence. Since we have exactly $mn$ linearly independent matrices in an $mn$-dimensional space, $\mathcal{B}$ must span $\mathbb{R}^{m \times n}$, and therefore forms a basis. 

For a constructive argument, let $\mathbf{X} \in \mathbb{R}^{m \times n}$ be arbitrary. Define coefficients

\begin{equation}
c_{pq} = \langle \mathbf{X}, \mathbf{u}_p \mathbf{v}_q^\top \rangle_F = \mathbf{u}_p^\top \mathbf{X} \mathbf{v}_q.
\end{equation}

Form the matrix

\begin{equation}
\widehat{\mathbf{X}} = \sum_{p=1}^m \sum_{q=1}^n c_{pq} \mathbf{u}_p \mathbf{v}_q^\top.
\end{equation}

For any fixed indices $(p', q')$, we compute

\begin{equation}
\begin{aligned}
\langle \widehat{\mathbf{X}}, \mathbf{u}_{p'} \mathbf{v}_{q'}^\top \rangle_F
&= \sum_{p,q} c_{pq} \langle \mathbf{u}_p \mathbf{v}_q^\top, \mathbf{u}_{p'} \mathbf{v}_{q'}^\top \rangle_F \\
&= \sum_{p,q} c_{pq} \delta_{pp'} \delta_{qq'} \\
&= c_{p'q'} .
\end{aligned}
\end{equation}

But by definition, $c_{p'q'} = \langle \mathbf{X}, \mathbf{u}_{p'} \mathbf{v}_{q'}^\top \rangle_F$, hence
\begin{equation}
\langle \widehat{\mathbf{X}} - \mathbf{X}, \mathbf{u}_{p'} \mathbf{v}_{q'}^\top \rangle_F = 0 \quad \text{for all } p', q'.
\end{equation}

Since $\mathcal{B}$ spans the space, the only matrix orthogonal to every basis element is the zero matrix. Therefore, $\widehat{\mathbf{X}} - \mathbf{X} = \mathbf{0}$, proving $\mathbf{X} = \widehat{\mathbf{X}}$. This provides an explicit expansion of any matrix in the basis $\mathcal{B}$, proving completeness.

\paragraph{(iii) Uniqueness.}

Orthogonality immediately gives that the coefficients are unique and equal to the Frobenius inner products:

\begin{equation}
c_{pq} = \langle \mathbf{X}, \mathbf{u}_p \mathbf{v}_q^\top \rangle_F = \mathbf{u}_p^\top \mathbf{X} \mathbf{v}_q.
\end{equation}

This completes the proof. Therefore, using the $\mathbf{u}$ and $\mathbf{v}$ matrices obtained from the SVD of a matrix to construct such outer-product basis matrices is valid and well-founded.

\end{proof}

\section{Experimental Details}
\subsection{Baselines}
\label{app:baselines}
\begin{itemize}[leftmargin=*]
    \item \textbf{MEMIT} \citep{memit} is the first method to support large-scale knowledge injection across multiple layers. It exploits the key–value structure \citep{keyvalue} of FFNs and improves upon ROME by restricting updates to a contiguous set of layers, allowing thousands of new facts to be inserted in one pass. However, MEMIT does not consider sequential editing, leaving space for later improvements.
    \item \textbf{RECT} \citep{rect}RECT is designed to mitigate the degradation of general abilities during sequential editing. 
It observes that general ability performance declines as more edits are applied, and addresses this by updating parameters based on the magnitude of change in individual weights. 
However, as our earlier analysis suggests, general abilities are governed by mappings between directions rather than individual parameters. 
Consequently, RECT remains too \textbf{localized} at the parameter level and fails to effectively preserve general abilities in long-horizon sequential editing.

\item \textbf{PRUNE} \citep{prune} is specifically designed for sequential editing with the goal of protecting the general abilities of LLMs. 
From the perspective of matrix conditioning, it constrains the maximum singular value of the update matrix so that it does not exceed that of the original parameter matrix, thereby reducing the risk of collapse. However, unlike our method, PRUNE does not filter the directions associated with large singular values, which may weaken knowledge retention. Moreover, its constraint only limits singular values to remain below a threshold, effectively attenuating but not eliminating the influence of noise. 

\item \textbf{NSE} \citep{nsee} is a method specifically designed for sequential knowledge editing. It preserves the original parameters during update computation, ensuring that each new edit does not interfere with previously injected knowledge. Inspired by the key–value view of FFN layers \citep{keyvalue}, NSE uses activation values to identify those neurons most relevant to the current update, restricting parameter changes within this subset. 
While this reduces unnecessary disturbance to the model, neuron-level selection alone cannot fully protect general abilities due to the problem of \emph{superposition}, where a single neuron may encode multiple orthogonal directions. As a result, NSE still fails to maintain general abilities under long-horizon sequential editing.

\item \textbf{AlphaEdit} \citep{alphaedit} is a method specifically designed for sequential knowledge editing. It constructs a protection subspace for previously stored knowledge by collecting 100K $(\text{subject}, \text{relation}, \text{object})$ triples from Wikipedia. During subsequent edits, parameter updates are projected onto the null space of this protection subspace to prevent interference with existing knowledge. 
However, based on our earlier analysis, sequential editing primarily perturbs the subspace associated with general abilities rather than factual knowledge alone. Thus, the choice of protection subspace in AlphaEdit is not sufficiently precise. As shown in our experiments, AlphaEdit can withstand more editing steps compared to other baselines, but eventually still suffers from a collapse of general abilities.

\item \textbf{DeltaEdit} \cite{deltaedit} is a sequential knowledge editing method that addresses the superimposed noise accumulation problem in LLMs. It mitigates performance degradation by identifying and reducing conflicts between consecutive edits. Specifically, it employs a dynamic orthogonal constraint strategy to ensure that new knowledge updates remain orthogonal to previous ones, thereby minimizing interference between different editing instances and maintaining model stability over long-term sequences.

\item \textbf{MEND} \cite{mend} is a hypernetwork-based knowledge editing method that addresses the scalability and efficiency challenges of updating massive LLMs. It enables fast, localized modifications by learning to transform standard fine-tuning gradients into targeted parameter updates. Specifically, it employs a low-rank gradient decomposition strategy to parameterize the editor network, thereby reducing the immense computational overhead and allowing rapid fact corrections without degrading the model's general capabilities.

\end{itemize}
\subsection{Datasets}\label{app:dataset}
\begin{itemize}[leftmargin=*]
    \item \textbf{ZsRE} \cite{zsre} is a question-answering (QA) dataset. Each sample contains a subject string and a corresponding answer, which serve as the editing target to assess \textbf{Efficacy}. To evaluate \textbf{Paraphrase}, it utilizes rephrased questions generated through back-translation. Following prior work, it employs out-of-scope Natural Questions to measure \textbf{Neighborhood} (also referred to as \textbf{Locality}).

    \item \textbf{Counterfact} \cite{memit} is a more challenging dataset that contrasts Counterfactual with factual statements. Each record is derived from an entry in the PARAREL dataset \cite{elazar2021measuring}, with all entities originating from WikiData. It uses metrics similar to ZsRE to evaluate \textbf{Efficacy Score}, \textbf{Paraphrase Score}, and \textbf{Neighborhood Score}. For its out-of-scope data, it replaces the subject entity with an approximate entity that shares the same predicate. Furthermore, Counterfact uniquely includes multiple generation prompts with the same meaning to test the \textbf{Fluency}(Generation Entropy) and \textbf{Consistency}(Reference Score) of the generated text.
\end{itemize}

\subsection{ZsRE Metrics} \label{app:zsreme}
In line with prior work \citep{rome,memit}, we define the evaluation metrics on the ZSRE dataset. 
Given a language model $f_\theta$, a factual prompt $(s_i, r_i)$, its target output $o_i$, and the model’s pre-edit prediction $o_i^c$, the following metrics are used:
\begin{itemize}[leftmargin=*]
    \item \textbf{Efficacy}: This metric reflects the model’s accuracy on the edited samples, computed as the average top-1 success rate:
    \begin{equation}
    \mathbb{E}_i  \left\{ o_i = \arg\max_o \mathbb{P}_{f_\theta}(o \mid (s_i,r_i) \right\}.
    \end{equation}
    
    \item \textbf{Paraphrase}: This measures how well the model transfers the edit to paraphrased forms of $(s_i,r_i)$, denoted as $N((s_i,r_i))$. It is defined as the average top-1 accuracy over these rephrasings:
    \begin{equation}
    \mathbb{E}_i \left\{ o_i = \arg\max_o \mathbb{P}_{f_\theta}(o \mid N((s_i,r_i)) \right\}.
    \end{equation}
    
    \item \textbf{Neighborhood}: This evaluates whether unrelated prompts $O(s_i,r_i)$ remain unaffected by the edit. It is measured as the proportion of cases where predictions for such prompts stay consistent:
    \begin{equation}
    \mathbb{E}_i \left\{ o_i^c = \arg\max_o \mathbb{P}_{f_\theta}(o \mid O((s_i,r_i)) \right\}.
    \end{equation}
\end{itemize}

\subsection{Counterfact Metrics} \label{app:ctfme}
In line with prior work \citep{rome,memit}, we further define the metrics used in the Counterfact benchmark. Given a language model $f_\theta$, a factual prompt $(s_i, r_i)$, a target output $o_i$, and the model’s pre-edit prediction $o_i^c$, we define:

\begin{itemize}[leftmargin=*]
    \item \textbf{Efficacy Score}: The fraction of cases where, for the prompt $(s_i,r_i)$, the target $o_i$ receives higher probability than the original output $o_c^i$:
    \begin{equation}
    \mathbb{E}_i \left[ \mathbb{P}_{f_\theta}[o_i \mid (s_i, r_i)] > \mathbb{P}_{f_\theta}[o_c^i \mid (s_i, r_i)] \right]. 
    \end{equation}
    \item \textbf{Paraphrase Score}: The proportion of paraphrased prompts $N((s_i, r_i))$ where the edited output $o_i$ is ranked higher than the original response $o_i^c$:
    \begin{equation}
    \mathbb{E}_i \left[ \mathbb{P}_{f_\theta}[o_i \mid N((s_i, r_i))] > \mathbb{P}_{f_\theta}[o_c^i \mid N((s_i, r_i))] \right]. 
    \end{equation}
    \item \textbf{Neighborhood Score}: The proportion of semantically related but distinct prompts $O((s_i, r_i))$ where the model maintains correct predictions, assigning higher probability to $o_i$ over $o_c^i$:
    \begin{equation}
    \mathbb{E}_i \left[ \mathbb{P}_{f_\theta}[o_i \mid O((s_i, r_i))] > \mathbb{P}_{f_\theta}[o_c^i \mid O((s_i, r_i))] \right]. 
    \end{equation}
    \item \textbf{Fluency}: A measure of output repetition, defined using the entropy of the n-gram distribution:
    \begin{equation}
    - \frac{2}{3} \sum_k g_2(k) \log_2 g_2(k) + \frac{4}{3} \sum_k g_3(k) \log_2 g_3(k),
    \end{equation}
    where $g_n(\cdot)$ denotes the frequency distribution over $n$-grams.
    \item \textbf{Consistency}: This evaluates how consistent the model’s generations are with external references. Given a subject $s$, we compute the cosine similarity between TF-IDF embeddings of the model’s text and the corresponding Wikipedia article about $o$.
\end{itemize}

\subsection{Details of GLUE}\label{app:glue}
GLUE \citep{GLUE} is a comprehensive benchmark, and this paper leverages the following six subtasks:
\begin{itemize}[leftmargin=*]
    \item \textbf{SST} \citep{SST} (Stanford Sentiment Treebank): A single-sentence classification task based on movie reviews, where the goal is to predict binary sentiment labels.  
    
    \item \textbf{MRPC} \citep{MRPC} (Microsoft Research Paraphrase Corpus): A sentence-pair task aiming to identify whether two sentences are semantically equivalent.  
    
    \item \textbf{MMLU} \citep{MMLU} (Massive Multi-task Language Understanding): A broad benchmark covering diverse subjects, designed to evaluate models under zero-shot and few-shot conditions.  
    
    \item \textbf{RTE} \citep{RTE} (Recognizing Textual Entailment): A natural language inference task where the objective is to determine if a premise entails its corresponding hypothesis.  
    
    \item \textbf{CoLA} \citep{CoLA} (Corpus of Linguistic Acceptability): A single-sentence classification benchmark that tests whether sentences are grammatically acceptable.  
    
    \item \textbf{NLI} \citep{NLI} (Natural Language Inference): A task requiring models to infer the logical relationship between a premise and a hypothesis. 
\end{itemize}

\subsection{Method Implementation Details}\label{impledetail}
All experiments based on GPT-J and LLaMA3 are conducted on NVIDIA A800 GPUs (80GB), while experiments on GPT2-XL are performed on NVIDIA RTX 4090 GPUs (24GB). For baselines, we adopt the official implementations of \textsc{AlphaEdit} and \textsc{NSE} and keep their default hyperparameter configurations unchanged. The only additional hyperparameter introduced by our method is the singular value projection threshold. Following Appendix~\ref{app:threshold}, we set this threshold to preserve the top 10\% dominant singular components for all models. We use the same threshold across baselines because the protected structure is model-dependent and determined solely by the spectrum of the underlying pretrained weights. For SVD computation, we use \texttt{torch.linalg.svd}. Before each update, we perform SVD on the current parameter matrix to construct the spectral basis used by our projection.

\section{Related Work (Full Version)}
\label{app:fullrelatedwork}
Knowledge editing methods for large language models can be broadly categorized into \emph{parameter-preserving} and \emph{parameter-modifying} approaches, depending on whether the original model parameters are directly updated. These two lines of work differ fundamentally in their design philosophy, flexibility, and susceptibility to interference under sequential editing. Below we review representative methods in each category, with an emphasis on techniques relevant to sequential knowledge editing.

\textbf{Parameter-Preserving Methods.}  
Parameter-preserving approaches maintain the base model’s parameters unchanged and instead incorporate external mechanisms to realize edits. A common direction is to attach additional modules that store and retrieve edited knowledge. For example, SERAC~\citep{serac} introduces an auxiliary memory with a Counterfactual model, CaliNet~\citep{calinet} and T-Patcher~\citep{tpatcher} insert neuron-based units, and GRACE~\citep{grace} organizes edits in a dynamic codebook. MELO~\citep{melo} uses additional LoRA-style adapters to preserve original parameters, while WISE~\citep{wang2024wise} improves stability and general ability with dual-memory and conflict-free sharding. Another line of work performs edits through prompting: MemPrompt~\citep{memprompt} and IKE~\citep{ike} rely on injecting new facts into the input context. More recent efforts combine symbolic structures with neural editing, such as OneEdit~\citep{oneedit}, which leverages knowledge graphs for collaborative knowledge updates.

\textbf{Parameter-Modifying Methods.}  
Parameter-modifying methods directly update the model’s weights to encode new knowledge. Meta-learning based techniques predict parameter shifts through hypernetworks, including MEND~\citep{mend}, MALMEN~\citep{MALMEN}, and InstructEdit~\citep{instructedit}. Locate-then-edit methods first determine the locations where knowledge is stored and then apply targeted modifications. Typical examples are ROME~\citep{rome}, which computes updates using closed-form equations, MEMIT~\citep{memit}, which scales editing to batches, GLAME~\citep{glame}, which integrates knowledge graphs, and AnyEdit~\citep{anyedit}, which recursively edits knowledge of arbitrary structure.  

When parameter-modifying edits are applied sequentially, additional difficulties arise due to the accumulation of interfering updates. To address these issues, several methods introduce explicit mechanisms to regularize or constrain updates during editing. RECT~\citep{rect} enforces sparsity at the level of individual parameters, PRUNE~\citep{prune} controls the conditioning of update matrices, AlphaEdit~\citep{alphaedit} restricts edits to a null space derived from previously stored knowledge, and DeltaEdit~\citep{deltaedit} mitigates edit--edit interference by projecting new updates onto the orthogonal complement of historical update directions. NSE~\citep{nsee} selects editing locations based on neuron activation patterns that are deemed most relevant for knowledge storage. While these methods alleviate certain forms of interference, they do not explicitly account for how sequential updates interact with the intrinsic structure of pretrained model parameters, which is the focus of our work.

\begin{table*}[t]
\centering
\caption{Comparison between REVIVE and representative SVD-based or subspace-constrained editing methods.}
\label{tab:svd_comparison}
\begin{tabular}{l  l l}
\hline
\textbf{Method} &
\textbf{Protected Subspace} &
\textbf{Failure Mode Addressed} \\
\hline
PRUNE &
Singular value magnitude constraint &
Large-magnitude updates \\
AlphaEdit &
Feature covariance from external factual data &
Interference across factual activations  \\
Delta-Edit &
Historical update directions &
Edit--edit interference\\
\textbf{\themodel} &
\textbf{Dominant singular vectors of weight matrices} &
\textbf{Dominant subspace corruption} \\
\hline
\end{tabular}
\end{table*}

\section{Comparison with SVD-based Editing Methods}
\label{app:svd_comparison}
This section clarifies the conceptual and methodological differences between \themodel and prior knowledge editing methods that employ SVD or subspace-based constraints, including {PRUNE} \cite{prune}, {AlphaEdit} \cite{alphaedit}, and {DeltaEdit} \cite{deltaedit}. Although these approaches share a common mathematical tool (e.g., singular value decomposition or subspace projection), \textbf{they differ fundamentally in how protected subspaces are defined, which failure modes they are intended to mitigate, and what aspects of model degradation they address.}

\paragraph{PRUNE.}
PRUNE constrains parameter updates by limiting the singular values of the update matrix relative to those of the original weights. This design primarily regulates the \emph{magnitude} of updates and provides a form of spectral regularization.  However, it does not distinguish between different singular directions and therefore does not explicitly preserve functionally meaningful subspaces of the pretrained model.
Consequently, PRUNE may suppress useful update components while failing to preserve the functional subspace that supports general model abilities.

\paragraph{AlphaEdit.}
AlphaEdit constructs a protected subspace based on the covariance of feature activations induced by a large external factual dataset (e.g., 100k triples from wikipedia). This approach aims to reduce interference between factual representations during sequential edits. As a result, the protected subspace is \emph{extrinsic} to the model and reflects correlations arising from the sampled data distribution rather than the intrinsic structure of the pretrained weight matrices. While effective at mitigating interference between edits, this design does not explicitly safeguard the dominant spectral components of the model parameters. Empirically, AlphaEdit exhibits severe degradation after long edit sequences, with both editing efficacy and general abilities collapsing after approximately 8k edits.

\paragraph{DeltaEdit.}
DeltaEdit focuses on preventing \emph{edit--edit interference} by tracking historical update directions and projecting new edits to be orthogonal to them. The protected subspace is thus derived entirely from past edits, without reference to the pretrained model’s internal parameter structure. Consequently, DeltaEdit addresses interference between edits but does not target model-level spectral drift, and cannot prevent degradation of general abilities that may arise even when consecutive edits are semantically independent.

\paragraph{REVIVE (ours).}
REVIVE is motivated by a failure mode that has received limited attention in prior sequential editing work: the progressive degradation of the \emph{dominant singular subspace} of weight matrices over long edit sequences. Our spectral analyses suggest that high-energy singular directions are strongly associated with general abilities. REVIVE therefore protects an \emph{intrinsic}, model-derived subspace defined by the dominant singular vectors of the pretrained weights. By selectively filtering update components that would interfere with this subspace, REVIVE mitigates spectral degradation and supports stable editing over long horizons (up to 20k sequential edits in our experiments).

\paragraph{Summary.}
As summarized in Table \ref{tab:svd_comparison}, existing methods employ subspace constraints to mitigate different forms of interference, they are primarily designed to control interactions between edits or between data instances. In contrast, REVIVE identifies degradation of the model’s intrinsic dominant singular subspace as a fundamental cause of failure in sequential editing and intervenes by selectively filtering update components that interfere with the dominant singular subspace.
\begin{figure}[htbp]
    \centering
    \begin{subfigure}{0.48\textwidth}
        \centering
        \includegraphics[width=\linewidth]{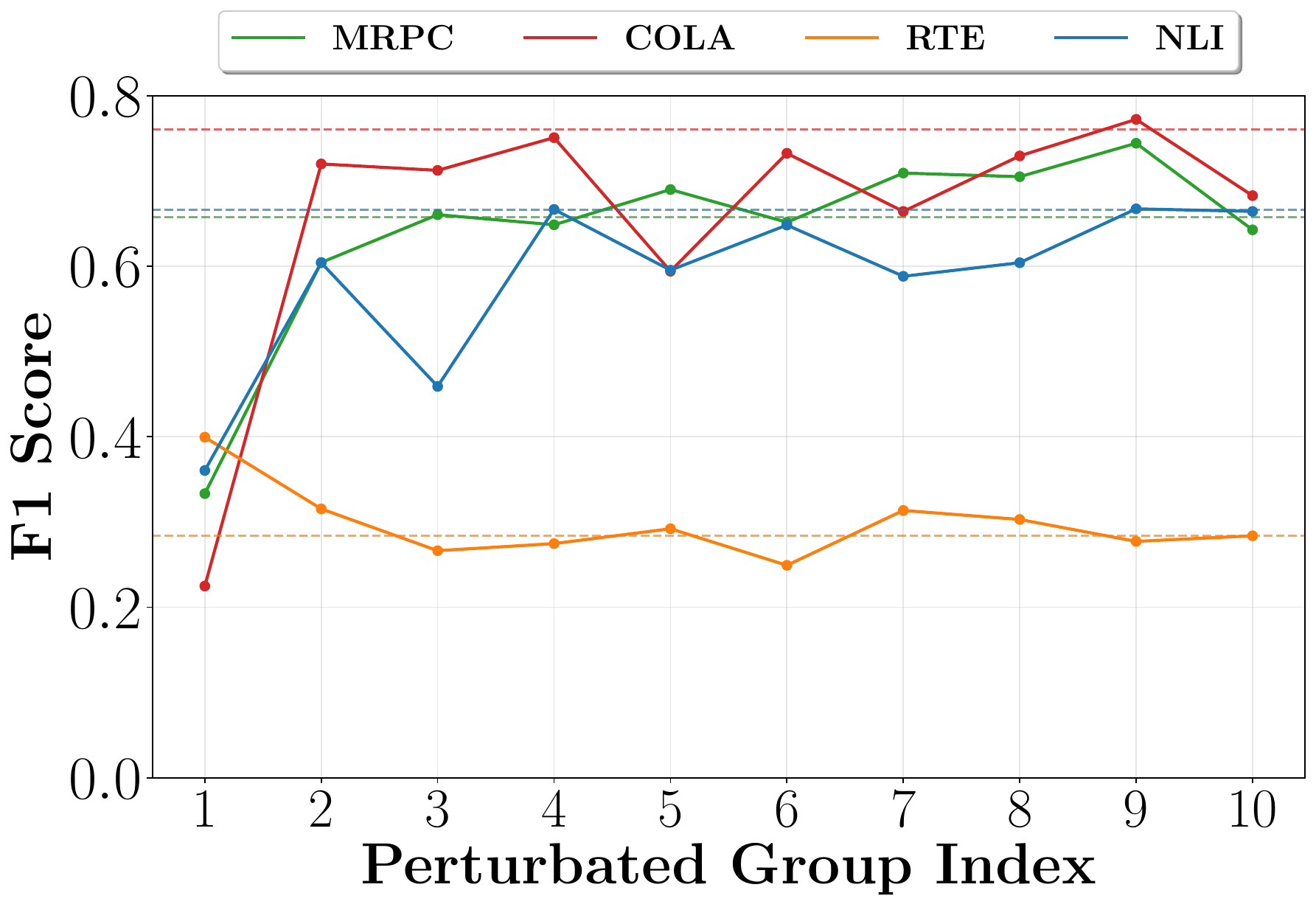}
    \end{subfigure}
    \caption{Sensitivity of general ability to perturbations across different output side (left singular vector) spectral groups.}
    \label{app:outputsidepic}
\end{figure}

\section{Supplementary Spectral Analyses for Section~\ref{sec:analysis}}
\subsection{Robustness to Output-Side Perturbations}
\label{app:outputsidepertuabtion}
To complement the input-side perturbation analysis presented in Section~\ref{anal:how}, we conduct a symmetric study on \emph{output-side perturbations}, examining how perturbations aligned with different groups of \emph{left} singular vectors affect model robustness.

Specifically, we partition the left singular vectors $\{\mathbf{u}_i\}$ of the weight matrix $\mathbf{W}$ into groups based on the cumulative energy of their corresponding singular values, using the same grouping strategy as in the main text. For a selected group $\mathcal{H}$, we construct structured perturbations that modify the contribution of these output directions while mixing them with all input directions:
\[
\mathbf{\Delta} = \sum_{i \in \mathcal{H}} \sum_{j=1}^r \beta_{i,j}\, \mathbf{u}_i \mathbf{v}_j^\top,
\beta_{i,j} \sim \mathcal{N}(0,1).
\]
To ensure comparability across groups, the perturbation is normalized and rescaled to a fixed magnitude:
\[
\mathbf{\tilde\Delta} = \varepsilon \cdot \frac{\mathbf{\Delta}}{\|\mathbf{\Delta}\|_F}.
\]
The perturbed weight matrix is given by $\mathbf{W}' = \mathbf{W} + \mathbf{\tilde\Delta}$. This construction can be interpreted as altering the \emph{input representation} of selected outputs $\{\mathbf{u}_i\}_{i \in \mathcal{H}}$ (left singular vector) into random mixtures of all inputs $\{\mathbf{v}_j\}_{j=1}^{r}$ (right singular vector). 

Figure~\ref{app:outputsidepic} reports the robustness curves under output-side perturbations for different singular value groups.
The observed trends closely mirror those of the input-side experiments: perturbations aligned with dominant singular directions lead to substantially larger degradation in model performance than perturbations confined to lower-energy components.

\subsection{Spectral Dynamics of AlphaEdit under Sequential Editing}
\label{app:specalphaedit}

This section supplements the analysis in Section~\ref{sec:memitlongsequenceediting} by examining the spectral dynamics of the strongest existing baseline, \textsc{AlphaEdit}, under long-sequence editing. We consider the same experimental setting as in the main text: $10{,}000$ sequential edits (100 edits per round for 100 rounds) on the \textsc{CounterFact} dataset using the \textsc{LLaMA3} model.
We track the evolution of the Singular Vector Similarity (SS) metric for dominant singular vectors throughout the editing process.

\begin{figure*}[thbp]
    \centering
    \captionsetup[subfigure]{font=footnotesize, skip=2pt}
    \small
    \begin{subfigure}[t]{0.19\textwidth}
        \includegraphics[width=\textwidth]{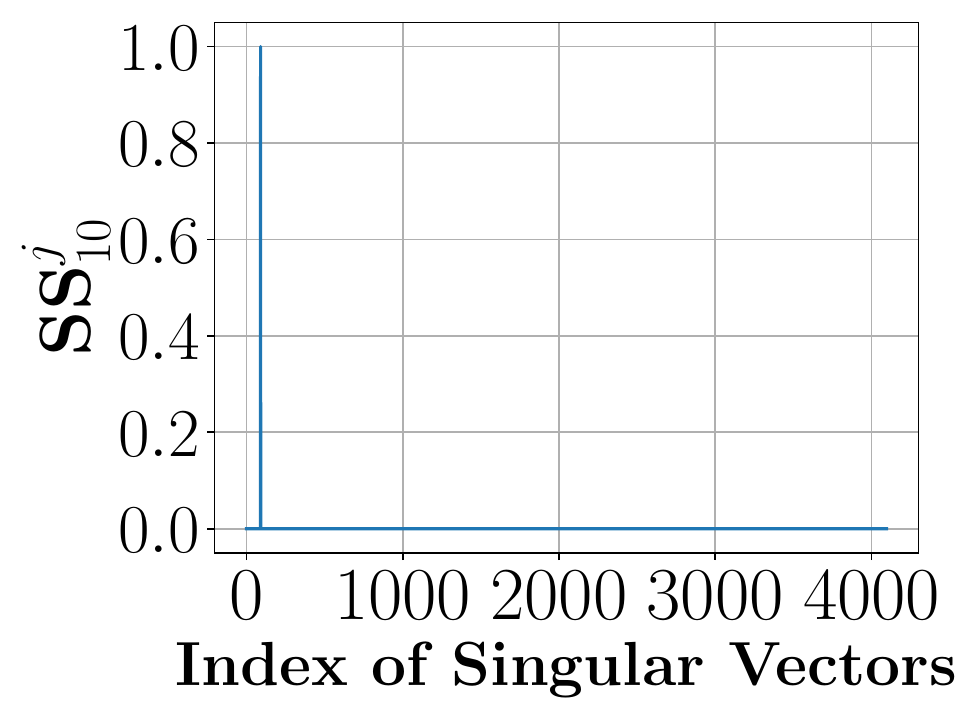}
    \end{subfigure}
    \begin{subfigure}[t]{0.19\textwidth}
        \includegraphics[width=\textwidth]{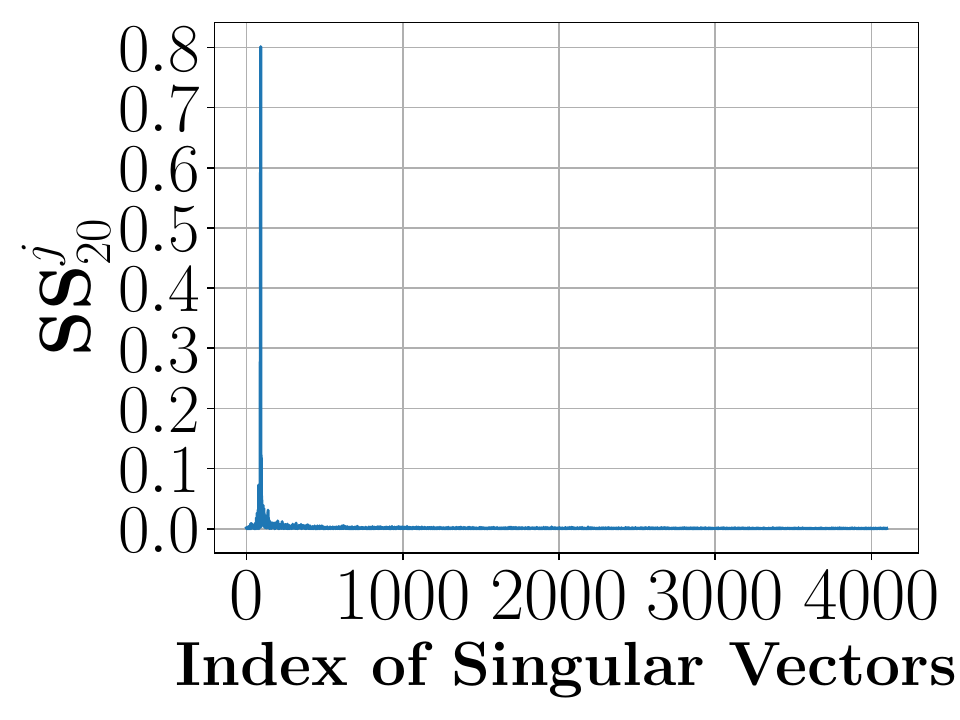}
    \end{subfigure}
    \begin{subfigure}[t]{0.19\textwidth}
        \includegraphics[width=\textwidth]{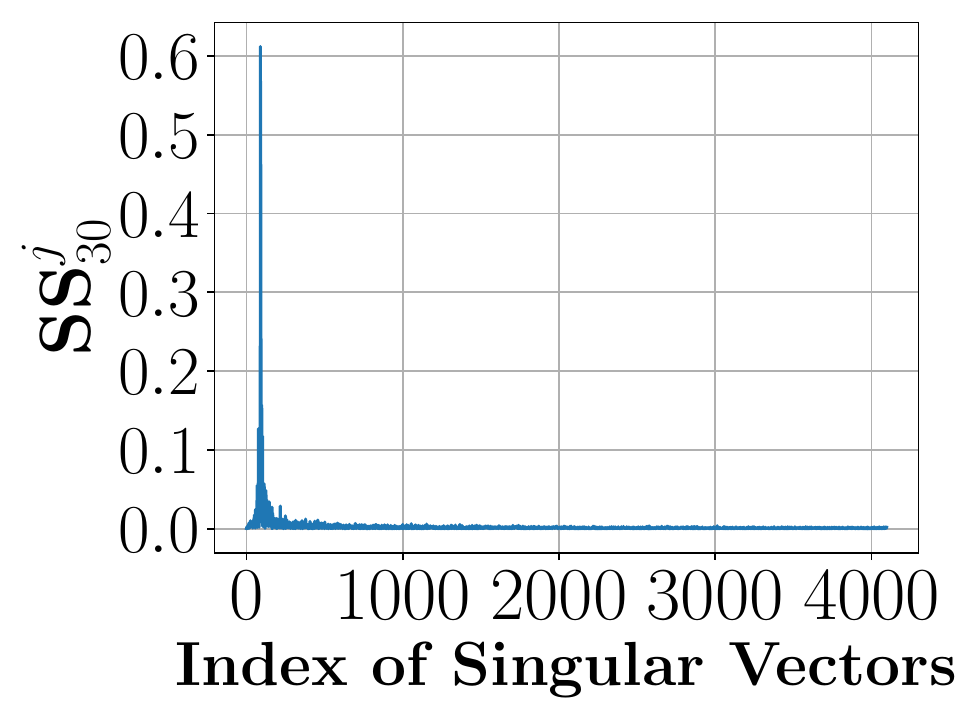}
    \end{subfigure}
    \begin{subfigure}[t]{0.19\textwidth}
        \includegraphics[width=\textwidth]{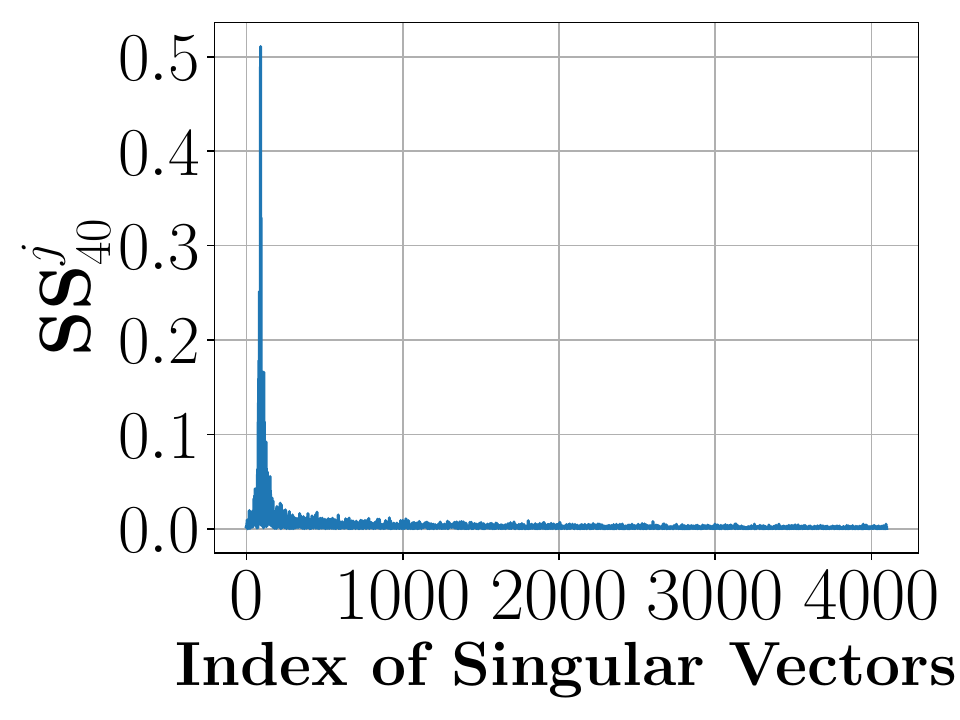}
    \end{subfigure}
    \begin{subfigure}[t]{0.19\textwidth}
        \includegraphics[width=\textwidth]{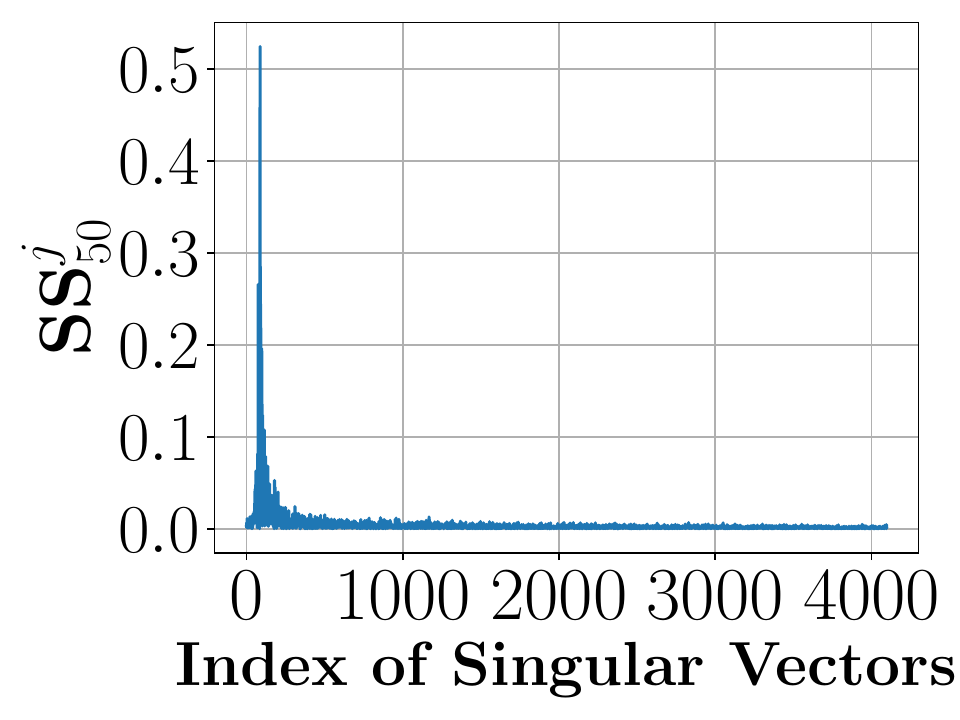}
    \end{subfigure}

    \par  

    \begin{subfigure}[t]{0.19\textwidth}
        \includegraphics[width=\textwidth]{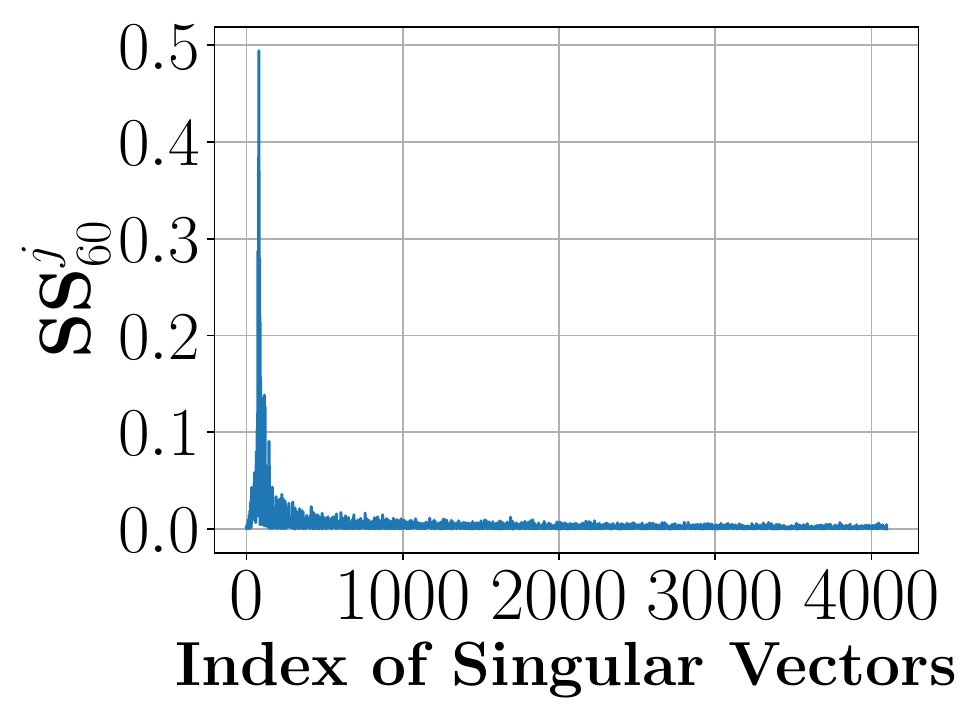}
    \end{subfigure}
    \begin{subfigure}[t]{0.19\textwidth}
        \includegraphics[width=\textwidth]{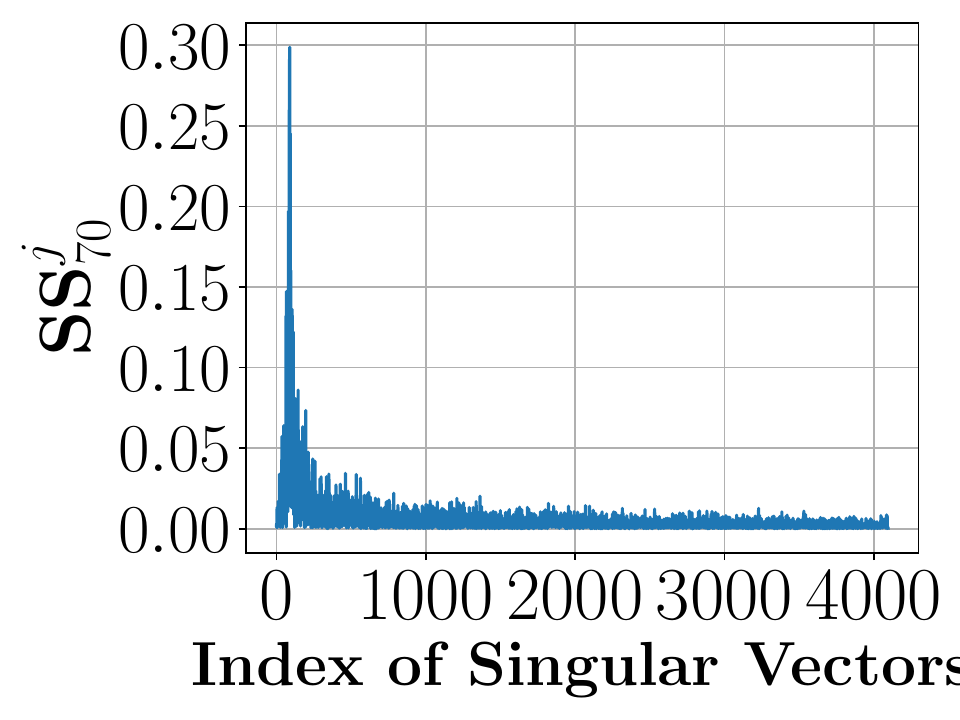}
    \end{subfigure}
    \begin{subfigure}[t]{0.19\textwidth}
        \includegraphics[width=\textwidth]{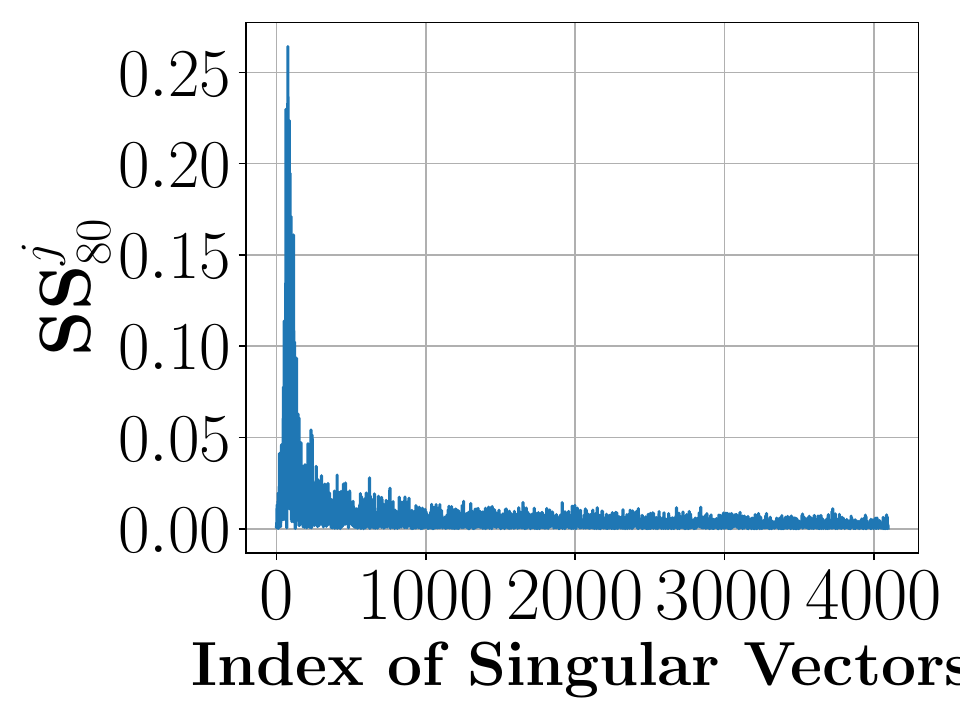}
    \end{subfigure}
    \begin{subfigure}[t]{0.19\textwidth}
        \includegraphics[width=\textwidth]{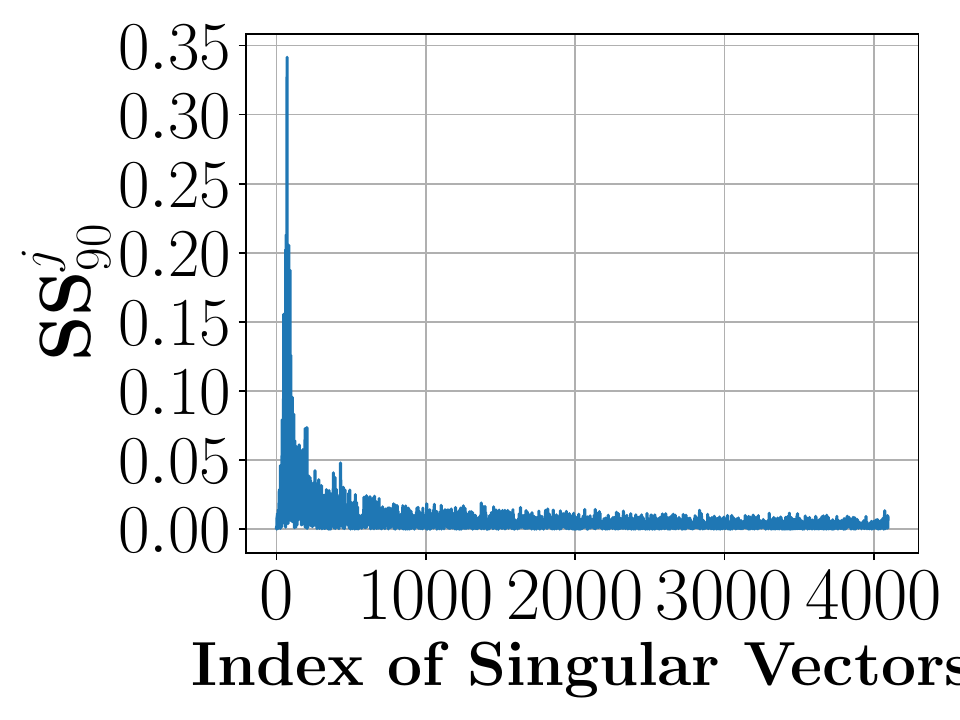}
    \end{subfigure}
    \begin{subfigure}[t]{0.19\textwidth}
        \includegraphics[width=\textwidth]{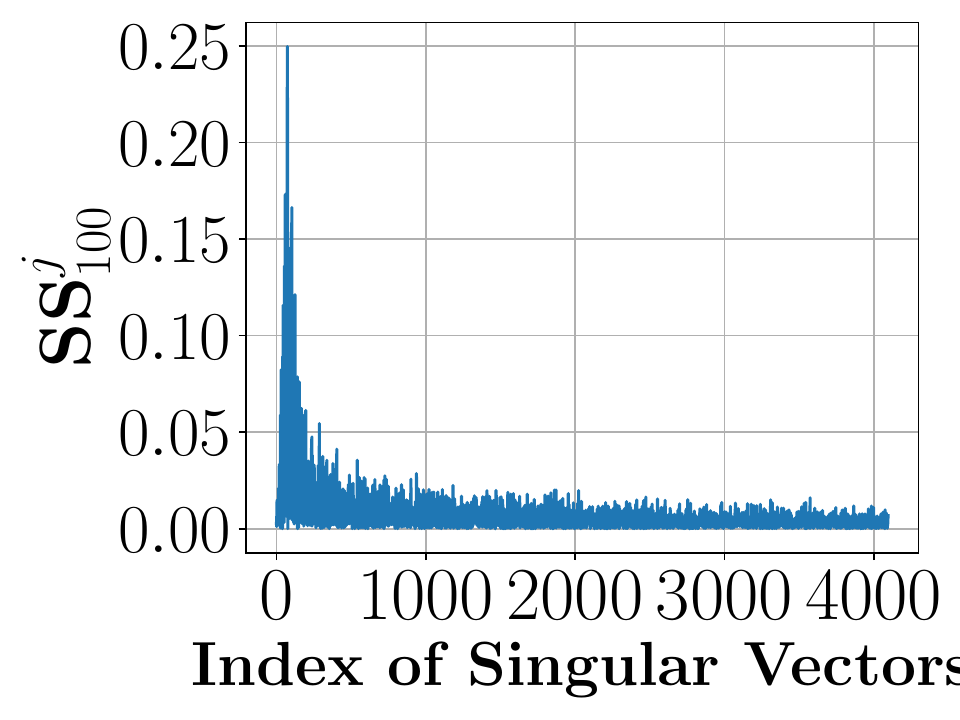}
    \end{subfigure}

\caption{Evolution of SS of AlphaEdit over sequential editing, from $\text{SS}_{10}$ to $\text{SS}_{100}$ with step size 10.}

    \label{fig:cosine-sim-alphaedit}
\end{figure*}

As shown in Figure~\ref{fig:cosine-sim-alphaedit}, \textsc{AlphaEdit} exhibits relatively small deviations in dominant singular vectors during the early editing rounds. However, as edits accumulate, the maximum SS value steadily decreases, indicating progressive rotation of critical singular directions.
By the end of the editing sequence, the maximum SS drops below $0.3$, suggesting substantial distortion of the dominant singular subspace. This spectral degradation closely mirrors the collapse in general abilities observed on the \textsc{GLUE} benchmark (Figure~\ref{fig:three-by-two-3d}). These results further support the conclusion that corruption of the dominant singular subspace is a general failure mode of long-sequence knowledge editing, rather than a method-specific artifact.



\begin{figure*}[thbp]
    \centering
    \captionsetup[subfigure]{font=footnotesize, skip=2pt}
    \small
    \begin{subfigure}[t]{0.19\textwidth}
        \includegraphics[width=\textwidth]{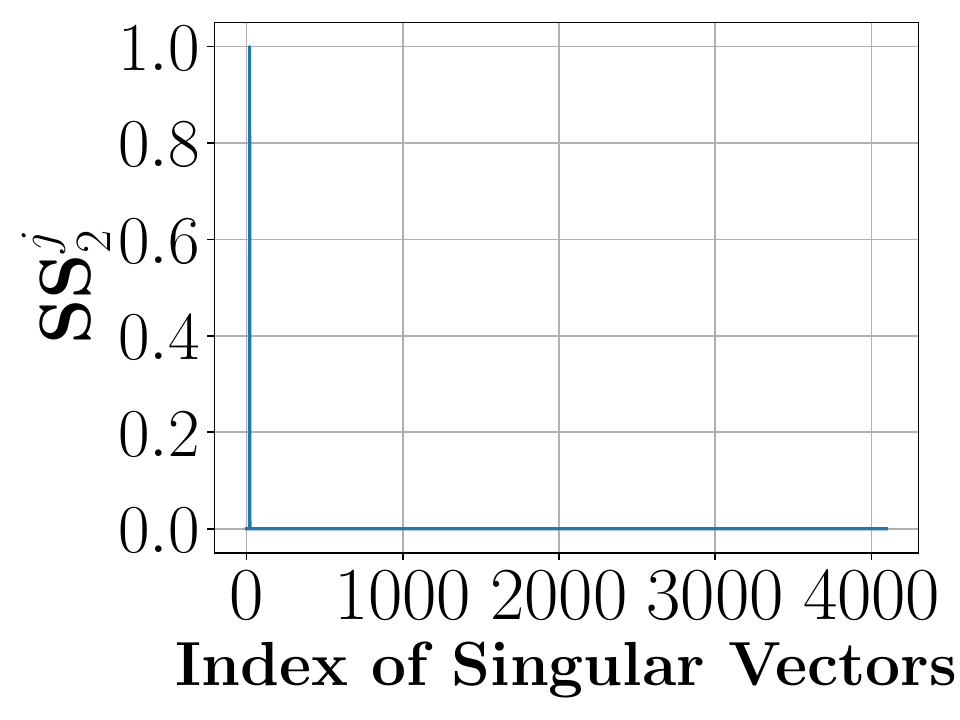}
        
    \end{subfigure}
    \begin{subfigure}[t]{0.19\textwidth}
        \includegraphics[width=\textwidth]{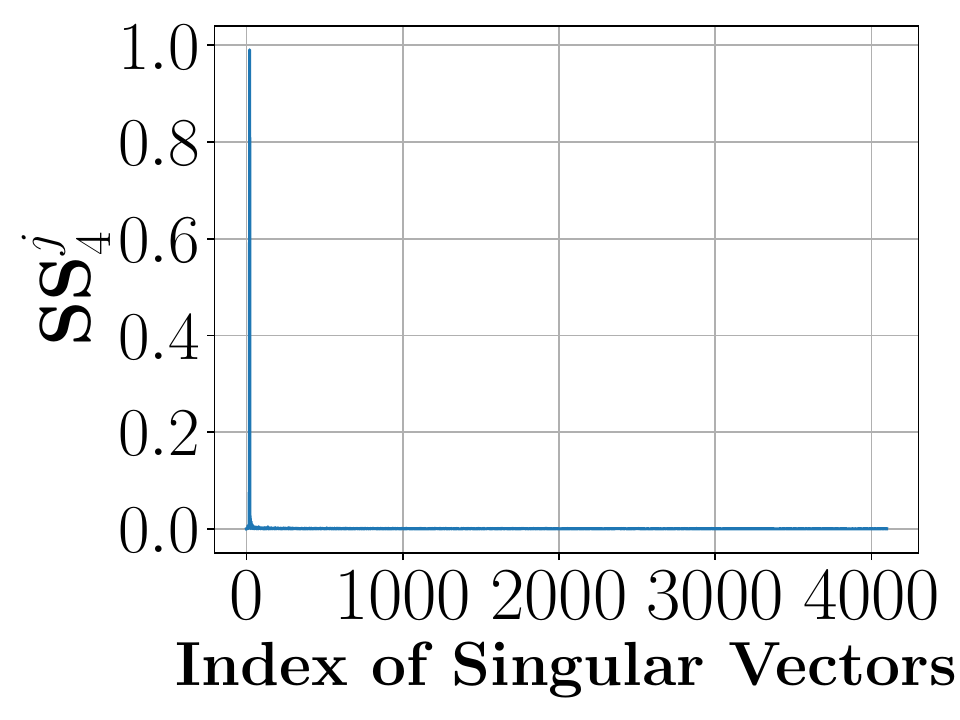}
        
    \end{subfigure}
    \begin{subfigure}[t]{0.19\textwidth}
        \includegraphics[width=\textwidth]{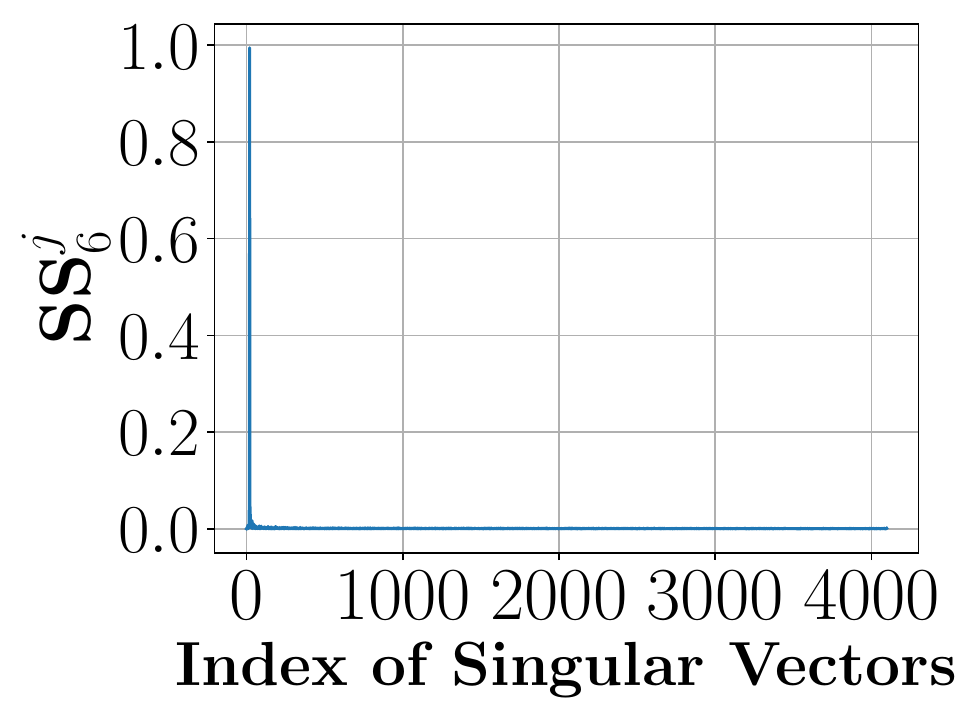}
        
    \end{subfigure}
    \begin{subfigure}[t]{0.19\textwidth}
        \includegraphics[width=\textwidth]{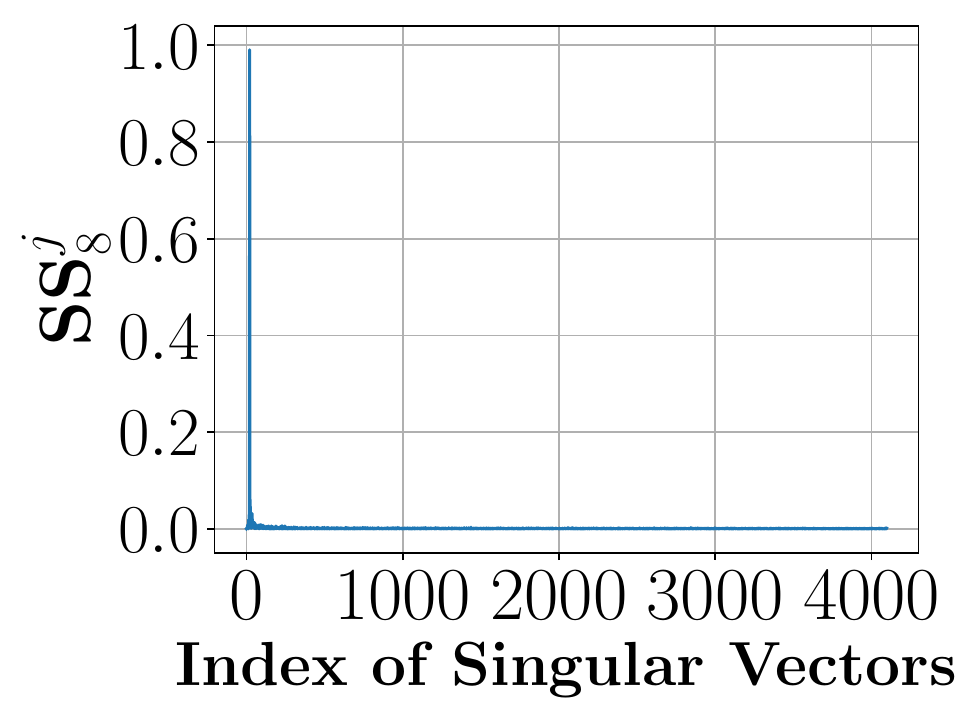}
        
    \end{subfigure}
    \begin{subfigure}[t]{0.19\textwidth}
        \includegraphics[width=\textwidth]{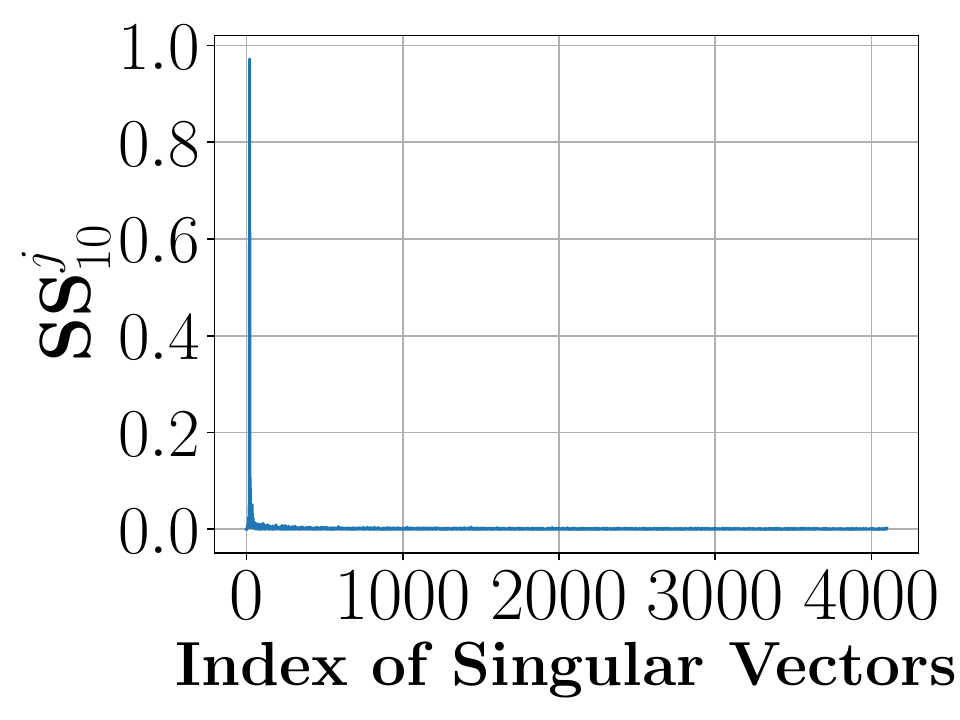}
        
    \end{subfigure}

    \par  

    \begin{subfigure}[t]{0.19\textwidth}
        \includegraphics[width=\textwidth]{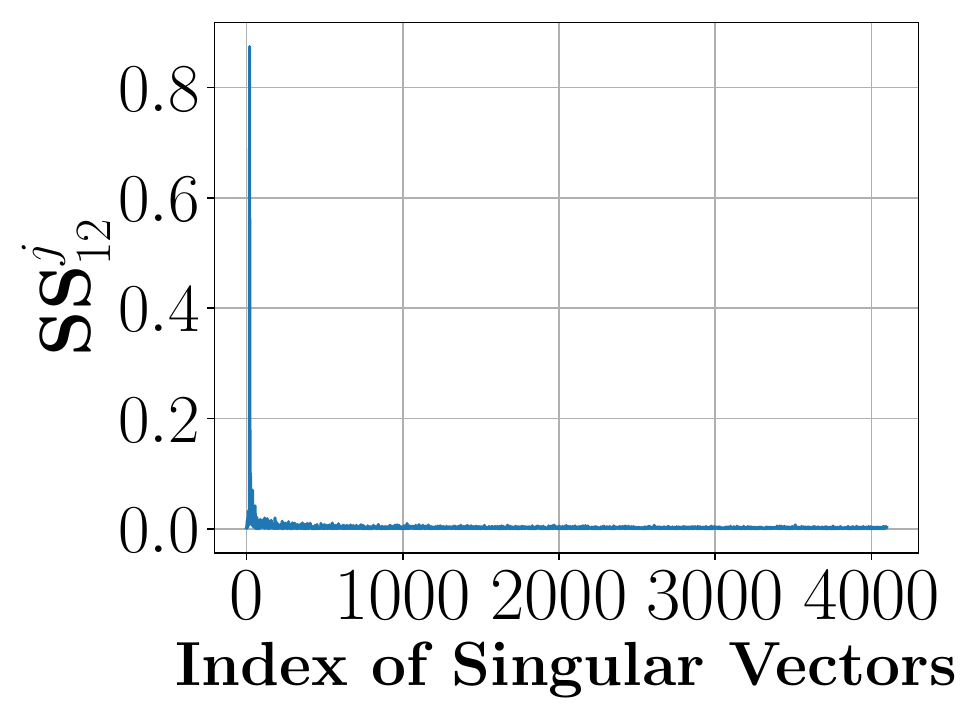}
       
    \end{subfigure}
    \begin{subfigure}[t]{0.19\textwidth}
        \includegraphics[width=\textwidth]{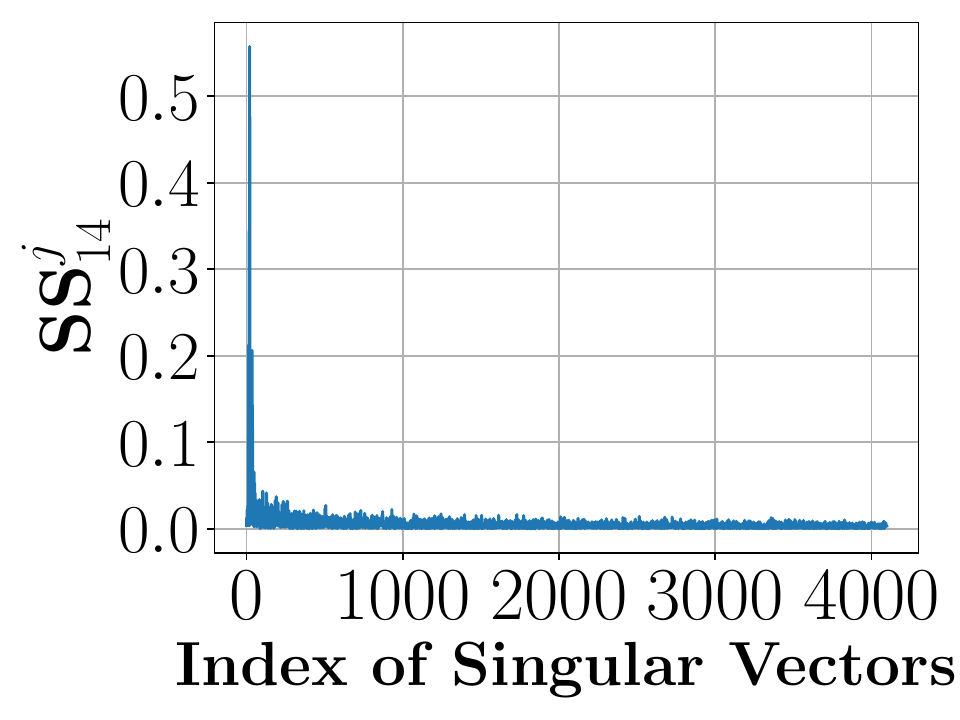}
        
    \end{subfigure}
    \begin{subfigure}[t]{0.19\textwidth}
        \includegraphics[width=\textwidth]{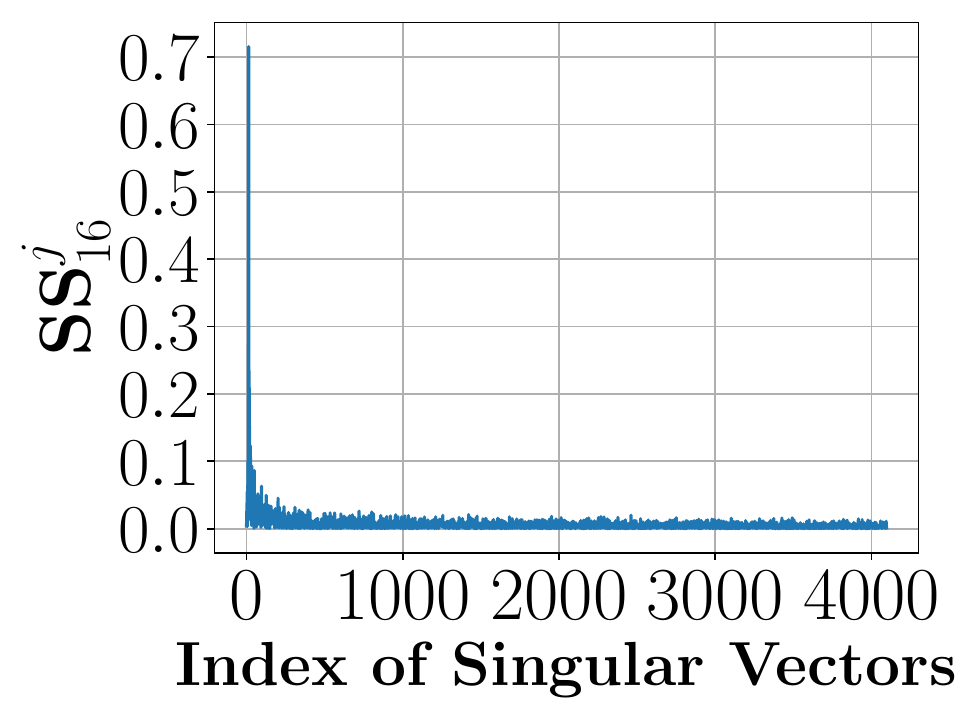}
        
    \end{subfigure}
    \begin{subfigure}[t]{0.19\textwidth}
        \includegraphics[width=\textwidth]{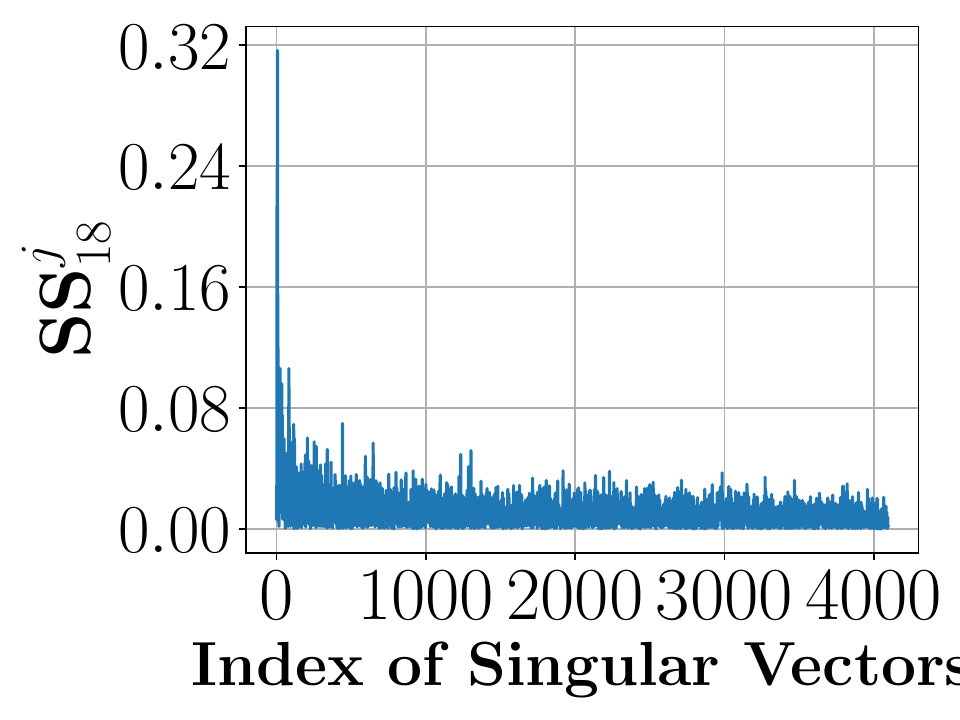}
        
    \end{subfigure}
    \begin{subfigure}[t]{0.19\textwidth}
        \includegraphics[width=\textwidth]{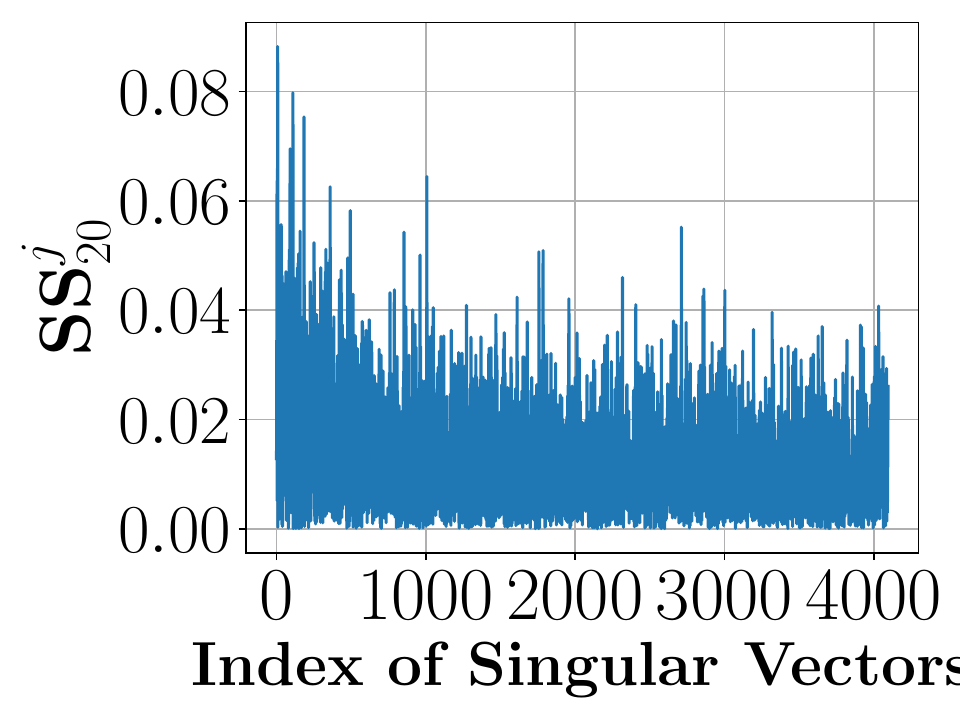}
        
    \end{subfigure}

    \caption{Evolution of Left Singular Vector Similarity (SS) over sequential editing.from $\text{SS}_{2}$ to $\text{SS}_{20}$ with step size 2.}
    \label{fig:leftvector}
\end{figure*}

\subsection{Dynamics of Left Singular Vectors}
\label{leftss}
In Section~\ref{sec:memitlongsequenceediting}, we focus on the evolution of right singular vectors to characterize microscopic degradation of the dominant subspace during sequential editing.  Here, we provide a complementary analysis on the corresponding \emph{left} singular vectors. As shown in Figure~\ref{fig:leftvector}, left singular vectors exhibit the same qualitative behavior as their right counterparts: dominant singular directions progressively rotate away from their original orientations as edits accumulate.
This consistency confirms that sequential editing induces a symmetric degradation of input–output mappings, rather than an artifact specific to one side of the decomposition.

\begin{table*}[t]
\centering
\vspace{-0.3cm}
\caption{
Performance on sequential editing over 10,000 Samples. 
The abbreviations \textit{Eff.} (Efficacy Success), \textit{Para.} (Paraphrase Success), 
\textit{Neigh.} (Neighborhood Success), \textit{Flu.} (Generation Entropy), and 
\textit{Consis.} (Reference Score) denote respective evaluation metrics. 
Relative improvements (\%) are shown in \blue{blue} and decreases in \orange{orange}. 
 $\uarbig$ indicates a large improvement where the baseline score was near zero.
}\label{tab:gpt2-xlresults}
\selectfont

\setlength{\tabcolsep}{1pt} 
\renewcommand{\arraystretch}{1} 

\resizebox{1\textwidth}{!}{%
\begin{tabular}{c|ccccc|ccc}
\toprule[1.5pt]
\textbf{Method}  & \multicolumn{5}{c|}{\textbf{Counterfact}} & \multicolumn{3}{c}{\textbf{ZsRE}} \\
\cmidrule(lr){2-6} \cmidrule(lr){7-9}
& \textbf{Eff.$\uparrow$} & \textbf{Para.$\uparrow$} & \textbf{Neigh.$\uparrow$} & \textbf{Flu.$\uparrow$} & \textbf{Consis.$\uparrow$} & \textbf{Eff.$\uparrow$} & \textbf{Para.$\uparrow$} & \textbf{Neigh.$\uparrow$} \\
\cmidrule{1-9}

\cmidrule[1pt]{1-9}
\textbf{GPT2-XL} & 21.82 & 24.16 & 78.32 & 626.69 & 31.34 & 22.17 & 21.28 & 24.20 \\
\cmidrule{1-9}
MEMIT & 70.56 & 62.42 & 55.94 & 516.26 & 8.74 & 53.00 & 46.27 & 12.76 \\
+\themodel & 90.46 $\uar{28.2\%}$ & 75.88 $\uar{21.5\%}$ & 63.83 $\uar{14.1\%}$ & 598.21 $\uar{15.9\%}$ & 34.32 $\uar{292.7\%}$ & 68.20 $\uar{28.7\%}$ & 60.80 $\uar{31.4\%}$ & 27.09 $\uar{112.4\%}$ \\
\cmidrule{1-9}
PRUNE & 57.61 & 54.01 & 52.87 & 596.56 & 6.93 & 0.21 & 0.19 & 2.06 \\
+\themodel & 82.00 $\uar{42.4\%}$ & 70.90 $\uar{31.3\%}$ & 62.82 $\uar{18.8\%}$ & 600.99 $\uar{0.7\%}$ & 34.55 $\uar{398.1\%}$ & 40.92 $\uarbig$ & 37.61 $\uarbig$ & 25.29 $\uar{1127.2\%}$ \\
\cmidrule{1-9}
RECT & 86.52 & 69.50 & 55.71 & 499.64 & 11.41 & 29.80 & 27.17 & 6.94 \\
+\themodel & 82.99 $\dar{4.1\%}$ & 69.20 $\dar{0.4\%}$ & 65.60 $\uar{17.7\%}$ & 595.69 $\uar{19.2\%}$ & 34.05 $\uar{198.3\%}$ & 62.45 $\uar{109.6\%}$ & 55.17 $\uar{103.0\%}$ & 26.20 $\uar{278.0\%}$ \\
\cmidrule{1-9}
AlphaEdit & 92.13 & 76.80 & 56.85 & 581.49 & 31.72 & 53.00 & 46.27 & 12.76 \\
+\themodel & 94.48 $\uar{2.6\%}$ & 78.70 $\uar{2.5\%}$ & 62.87 $\uar{10.6\%}$ & 587.94 $\uar{1.1\%}$ & 38.51 $\uar{21.5\%}$ & 68.10 $\uar{28.5\%}$ & 57.17 $\uar{23.5\%}$ & 20.35 $\uar{59.4\%}$ \\
\cmidrule{1-9}
NSE & 69.22 & 54.54 & 69.26 & 596.41 & 28.87 & 33.71 & 32.31 & 22.70 \\
+\themodel & 96.12 $\uar{38.8\%}$ & 84.49 $\uar{54.9\%}$ & 64.17 $\dar{7.4\%}$ & 592.71 $\dar{0.6\%}$ & 37.74 $\uar{30.9\%}$ & 77.83 $\uar{131.0\%}$ & 70.55 $\uar{118.4\%}$ & 24.84 $\uar{9.4\%}$ \\
\bottomrule[1.5pt]
\end{tabular}}
\vspace{-0.2cm}
\end{table*}

\section{Supplementary Experimental Results}

\subsection{Performance on GPT2-XL under Sequential Editing}
\label{app:gpt2-xl}
In this subsection, we additionally report the results on GPT2-XL (1.5B). Table~\ref{tab:gpt2-xlresults} summarizes the performance of REVIVE under sequential editing on this model. Across all evaluated baselines, incorporating REVIVE consistently improves editing performance and stability. These results indicate that the effectiveness of REVIVE generalizes well across different model sizes.

\begin{table*}[t!]
\centering
\caption{Comparison between AlphaEdit-REVIVE and DeltaEdit.}
\label{deltaeditcom}
\begin{tabularx}{\textwidth}{llXXXXXXXX}
\toprule
\multirow{2}{*}{\textbf{Model}} & \multirow{2}{*}{\textbf{Method}} 
& \multicolumn{3}{c}{\textbf{CounterFact}} 
& \multicolumn{3}{c}{\textbf{ZsRE}} \\ 
\cmidrule(lr){3-5} \cmidrule(lr){6-8}
& 
& Eff.$\uparrow$ & Para.$\uparrow$ & Neigh.$\uparrow$ 
& Eff.$\uparrow$ & Para.$\uparrow$ & Neigh.$\uparrow$ \\ 
\midrule

\multirow{6}{*}{\textbf{Llama3-8B}} 
    & MEMIT 
    & 50.54 & 50.50 & 49.46 
    & 0.05 & 0.05 & 1.41 \\
    & PRUNE 
    & 50.33 & 50.40 & 49.63 
    & 0 & 0 & 0 \\
    & RECT 
    & 49.02 & 49.23 & 51.00 
    & 0 & 0 & 0 \\
    & DeltaEdit 
    & 98.63 & 83.20 & \textbf{78.67} 
    & 94.94 & 91.19 & 30.08 \\
    \cmidrule(lr){2-8}
    & AlphaEdit 
    & 93.83 & 84.75 & 64.44 
    & 94.49 & 91.52 & 29.77 \\
    & +REVIVE 
    & \textbf{99.37} & \textbf{93.50} & 64.40 
    & \textbf{95.19} & \textbf{91.58} & \textbf{32.17} \\

\bottomrule
\end{tabularx}
\end{table*}

\begin{table*}[!htbp]
\centering
\caption{ Performance results of REVIVE enhanced Baseilnes under extreme sequential editing (20,000 edits).}
\label{20000edit}

\renewcommand{\arraystretch}{0.9}
\resizebox{\textwidth}{!}{%
\begin{tabular}{c c|ccccc|ccc}
\toprule[1.5pt]
\textbf{Model} & \textbf{Method}  & \multicolumn{5}{c|}{\textbf{Counterfact}} & \multicolumn{3}{c}{\textbf{ZsRE}} \\
\cmidrule(lr){3-7} \cmidrule(lr){8-10}
&& \textbf{Eff.$\uparrow$} & \textbf{Para.$\uparrow$} & \textbf{Neigh.$\uparrow$} & \textbf{Flu.$\uparrow$} & \textbf{Consis.$\uparrow$} & \textbf{Eff.$\uparrow$} & \textbf{Para.$\uparrow$} & \textbf{Neigh.$\uparrow$} \\
\cmidrule{1-10}
\multirow{4}{*}{LLaMA3} 

&  MEMIT-REVIVE & 91.94 & 79.67 & 56.90 & 557.61 & 26.44 & 84.11 & 79.85 & 32.92 \\
\cmidrule{2-10}
&  RECT-REVIVE & 89.00 & 76.78 & 60.54 & 594.38 & 27.93 & 79.35 & 76.35 & 30.24 \\
\cmidrule{2-10}
& AlphaEdit-REVIVE & 97.50 & 87.24 & 57.65 & 613.22 & 32.77 & 92.62 & 88.25 & 31.31 \\
\cmidrule{2-10}
&  NSE-REVIVE & 98.50 & 90.38 & 61.78 & 615.65 & 33.23 & 93.91 & 89.67 & 31.58 \\
\bottomrule[1.5pt]
\end{tabular}}
\end{table*}

\subsection{Performance Comparison with DeltaEdit}
\label{app:deltacomp}

DeltaEdit~\cite{deltaedit} is evaluated under an experimental protocol that differs substantially from our main setting.
Specifically, DeltaEdit performs one edit per sample and reports results over $3{,}000$ edited samples, whereas our primary experiments apply batched updates ($100$ edits per round) over a total of $10{,}000$ edits.
To ensure a fair and meaningful comparison, we therefore conduct a separate evaluation of DeltaEdit under its original experimental setting. Moreover, since DeltaEdit is proposed as an extension of AlphaEdit, we compare it against \textsc{AlphaEdit}+\themodel to ensure a fair and meaningful evaluation.


Table~\ref{deltaeditcom} reports the results. Both DeltaEdit and the \themodel-enhanced model substantially improve over the AlphaEdit baseline across most metrics. Notably, \textsc{AlphaEdit}+\themodel achieves the strongest overall performance, outperforming DeltaEdit on all metrics except for \emph{Neighborhood Score} on \textsc{CounterFact}. We further note that \themodel and DeltaEdit address long-horizon degradation from different perspectives. \themodel prevents collapse by protecting a model-derived dominant structure during parameter updates, which yields broad and consistent improvements across metrics. In contrast, DeltaEdit primarily improves edit quality by reducing interference between edits. Overall, \themodel delivers more pronounced gains in our evaluation.


\subsection{Full Glue Results}\label{app:gluefull}
\begin{figure*}[t]\label{remainglue}

  \begin{subfigure}{0.32\textwidth}
    \includegraphics[width=\linewidth]{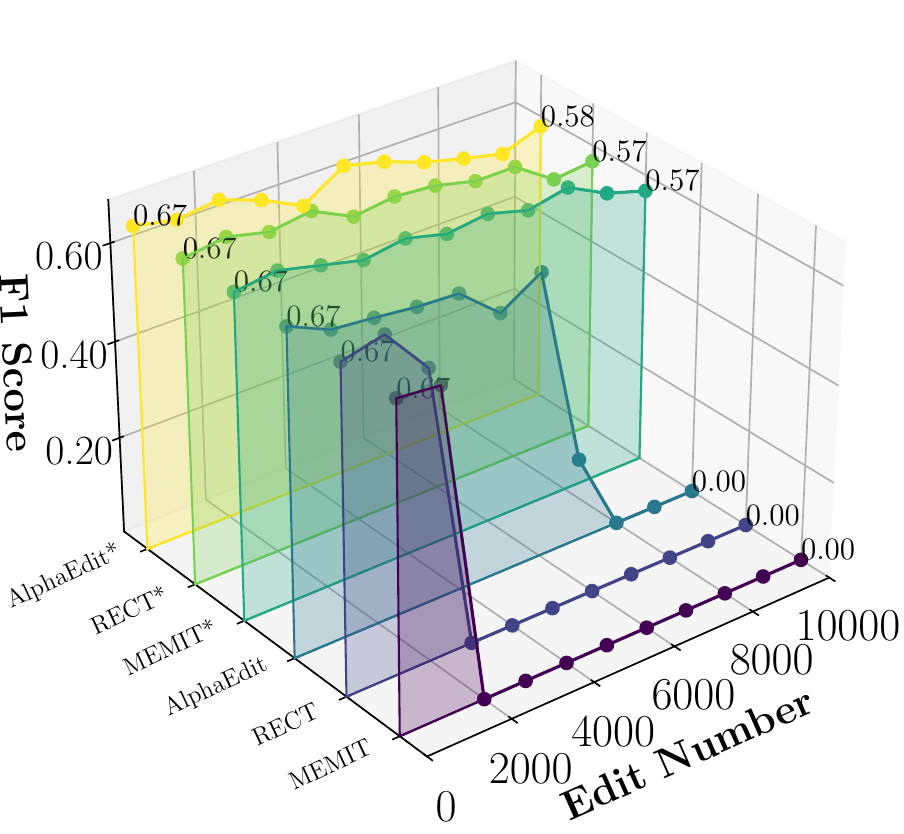}
    \caption{NLI}
  \end{subfigure}\hfill
  \begin{subfigure}{0.32\textwidth}
    \includegraphics[width=\linewidth]{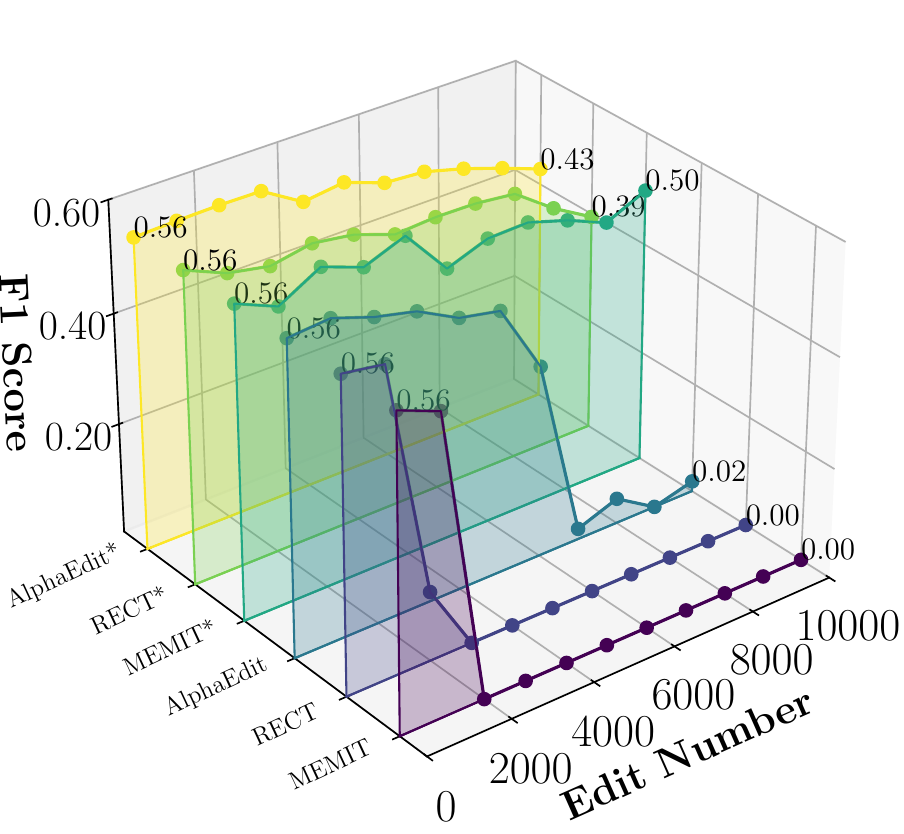}
    \caption{MMLU}
  \end{subfigure}\hfill
  \begin{subfigure}{0.32\textwidth}
    \includegraphics[width=\linewidth]{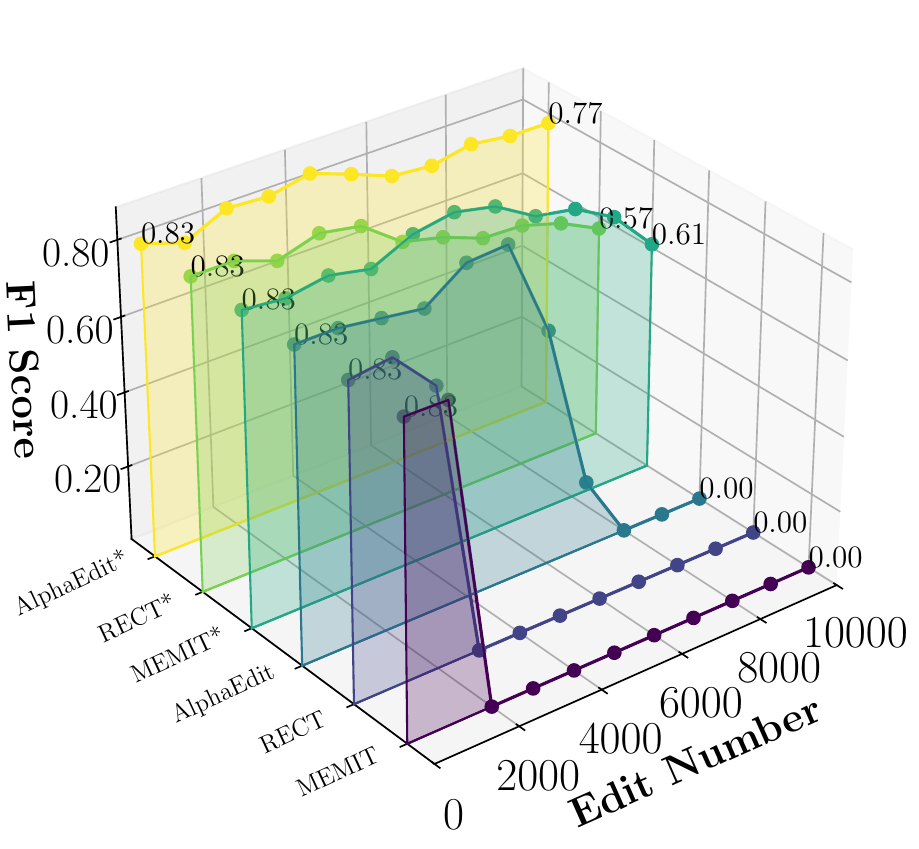}
    \caption{SST}
  \end{subfigure}
  \caption{Baseline and corresponding REVIVE version(*) performance on GLUE across datasets.}
  \label{fig:rest-3d}
\end{figure*}

This section reports the remaining GLUE evaluation results that were omitted from Section~\ref{resultandnalysis} due to space constraints. Results for the remaining three GLUE tasks are shown in Figure~\ref{fig:rest-3d}.

\begin{figure*}[h]
  \centering
  \begin{subfigure}{0.32\textwidth}
    \includegraphics[width=\linewidth]{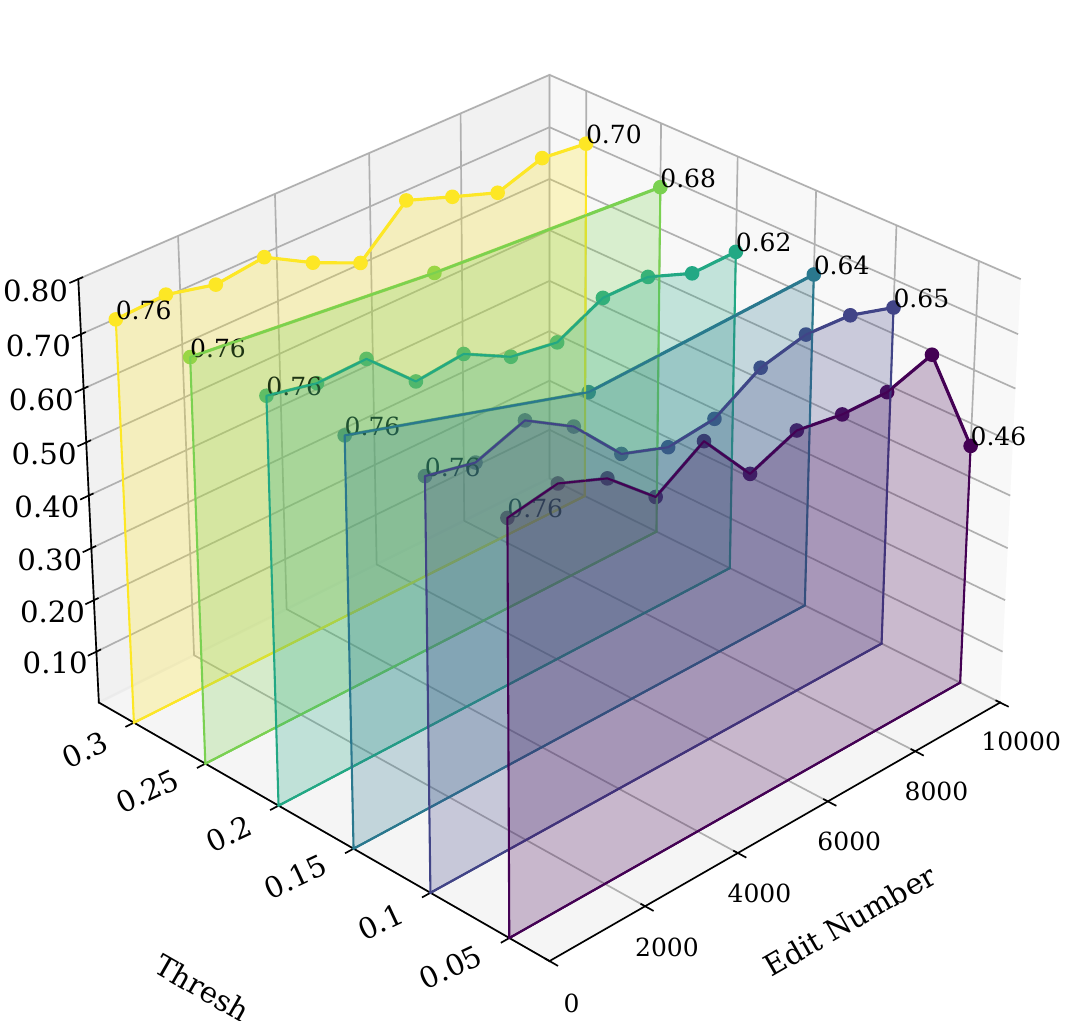}
    \caption{COLA}
  \end{subfigure}\hfill
  \begin{subfigure}{0.32\textwidth}
    \includegraphics[width=\linewidth]{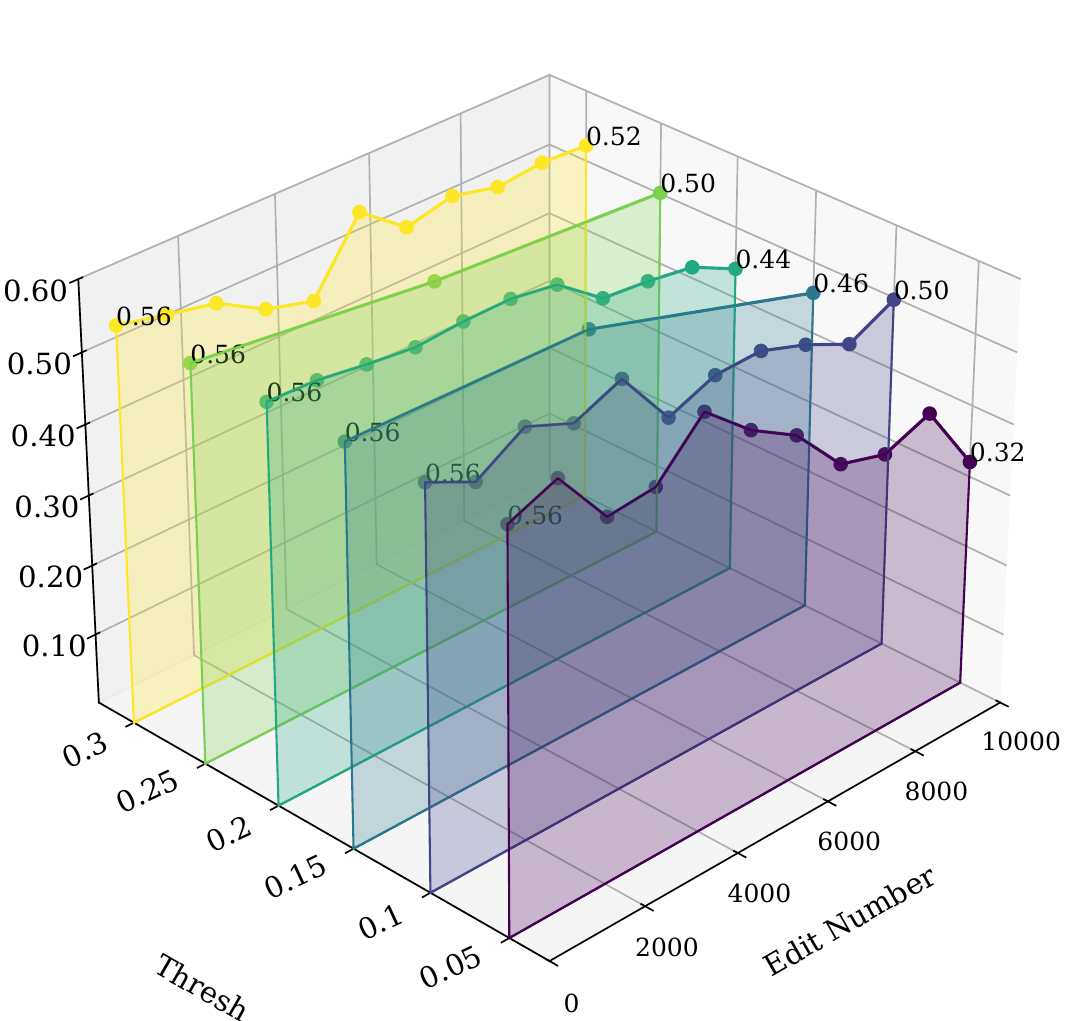}
    \caption{MMLU}
  \end{subfigure}\hfill
  \begin{subfigure}{0.32\textwidth}
    \includegraphics[width=\linewidth]{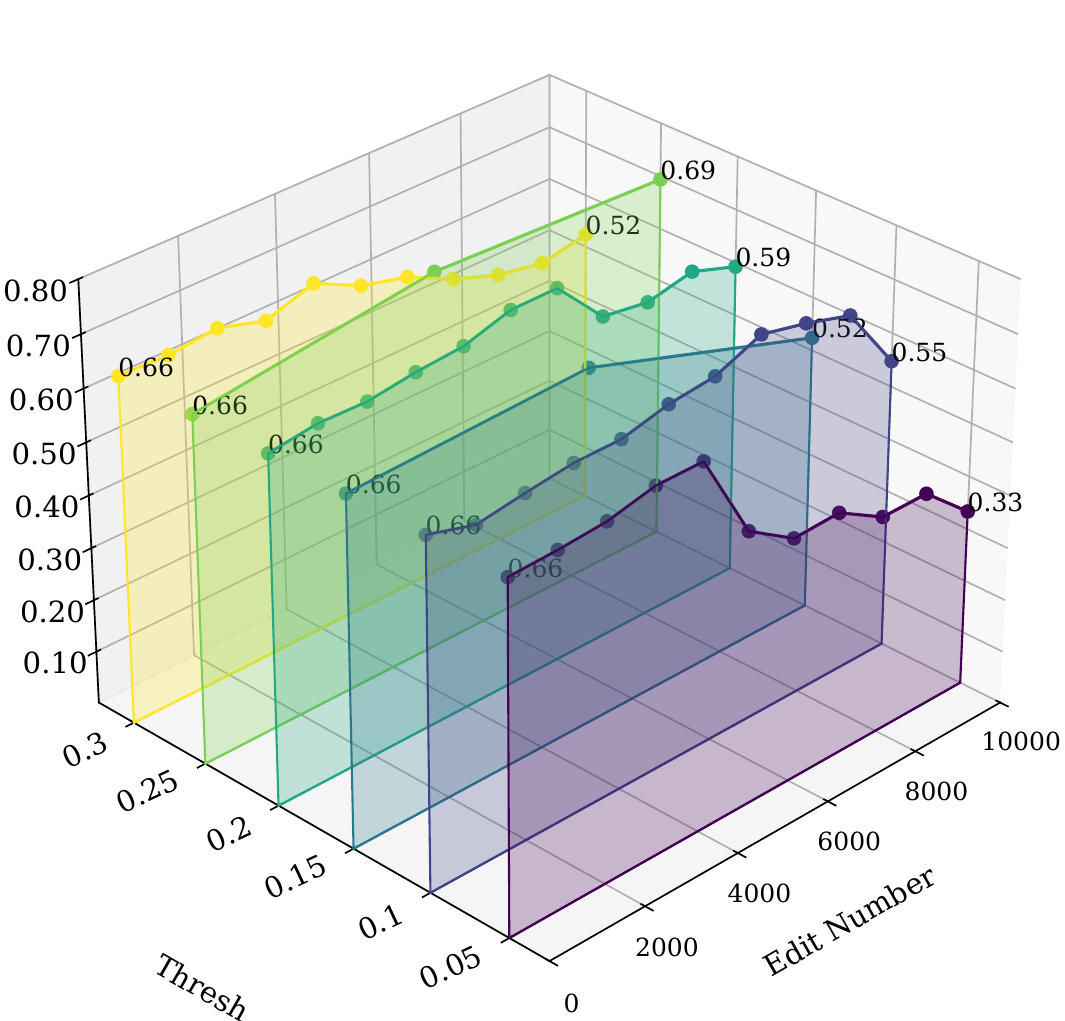}
    \caption{MRPC}
  \end{subfigure}

  \begin{subfigure}{0.32\textwidth}
    \includegraphics[width=\linewidth]{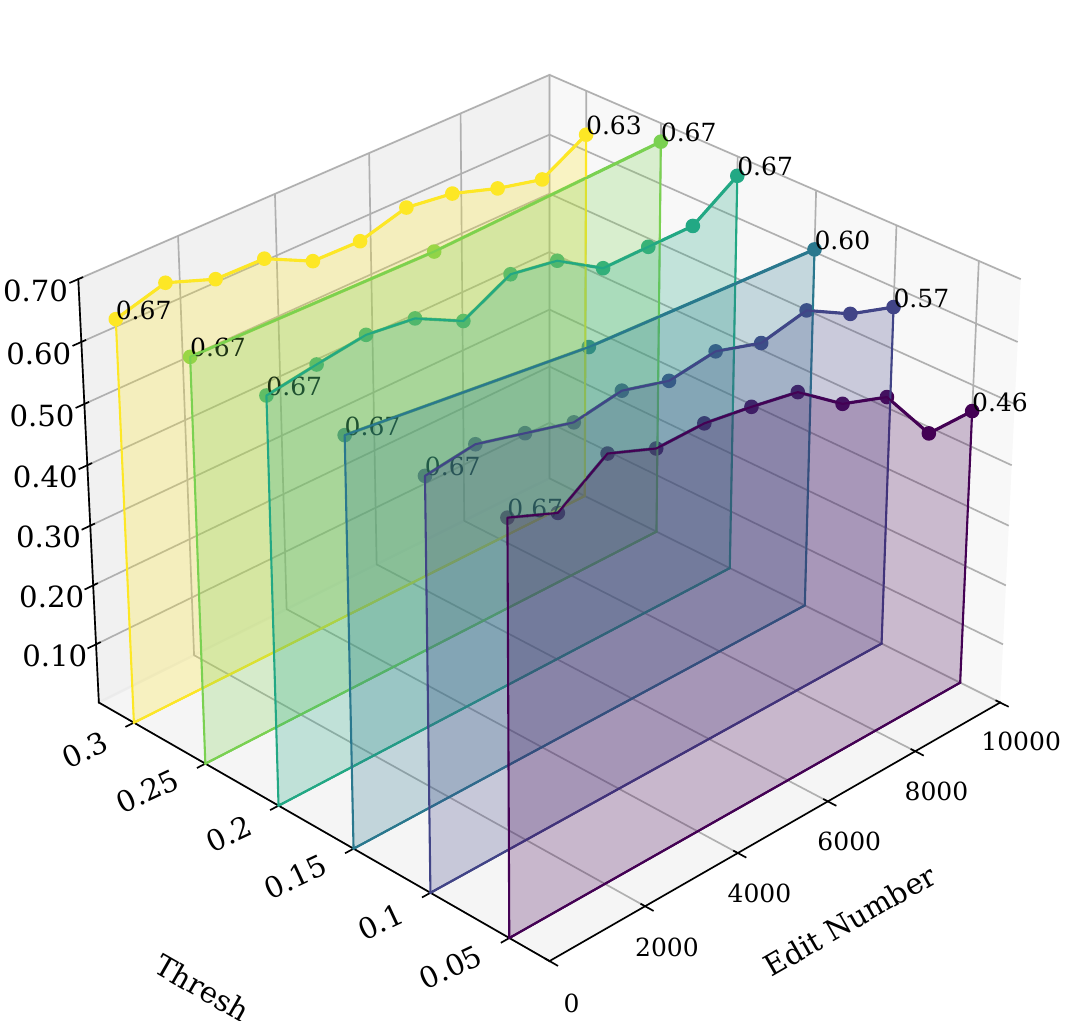}
    \caption{NLI}
  \end{subfigure}\hfill
  \begin{subfigure}{0.32\textwidth}
    \includegraphics[width=\linewidth]{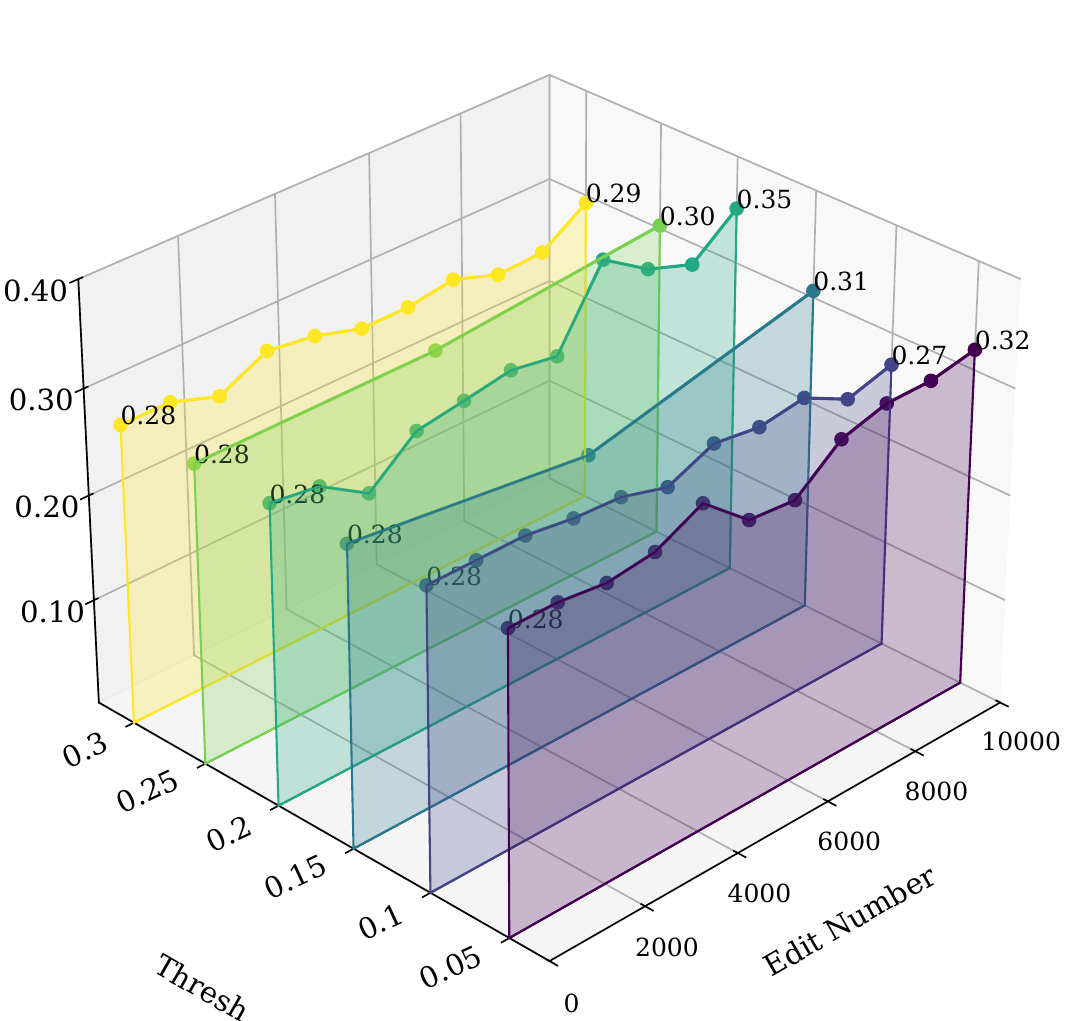}
    \caption{RTE}
  \end{subfigure}\hfill
  \begin{subfigure}{0.32\textwidth}
    \includegraphics[width=\linewidth]{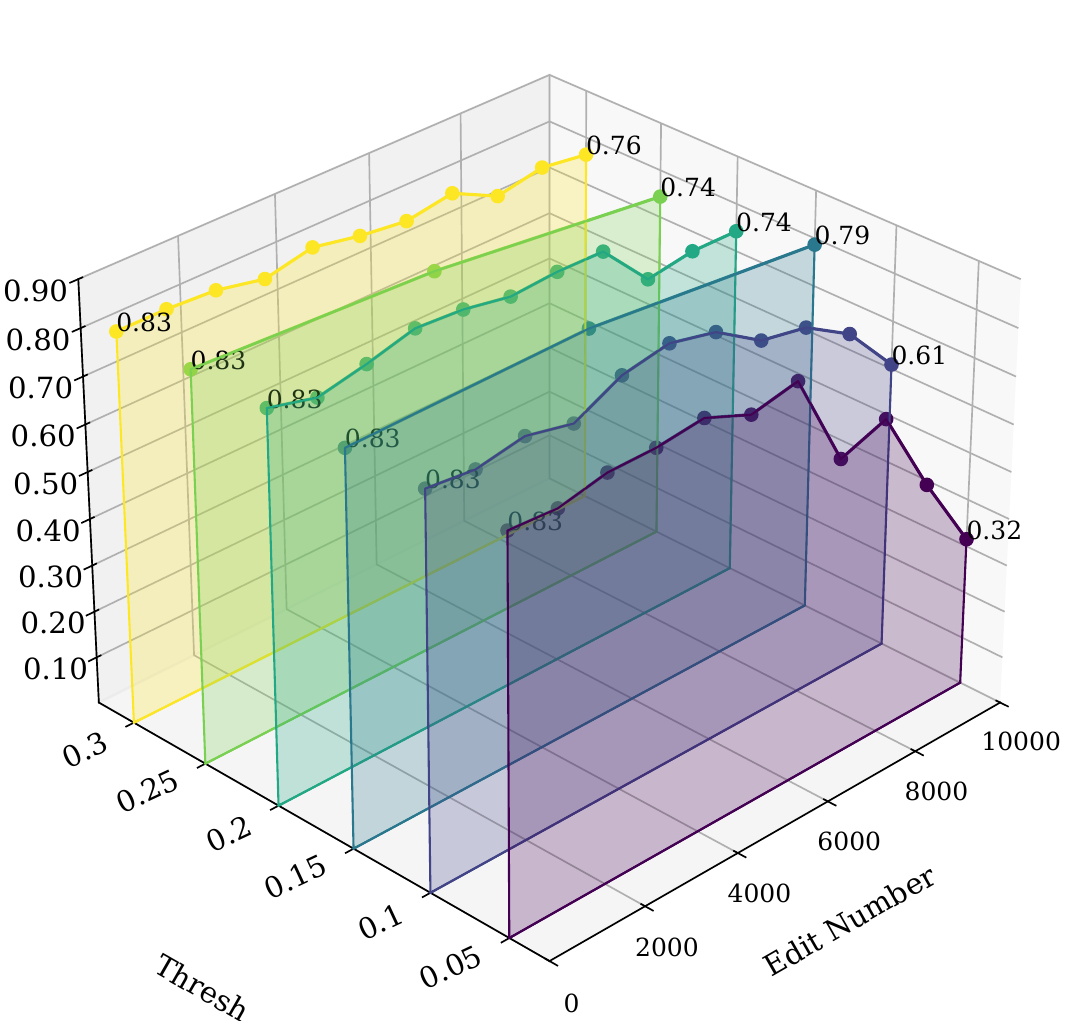}
    \caption{SST}
  \end{subfigure}

  \caption{GLUE evaluation results on LLaMA3 after 10,000 edits on the CounterFact dataset using MEMIT-REVIVE with different protection thresholds.}
  \label{fig:threshglue3d}
\end{figure*}

\subsection{REVIVE Enhanced Baselines Under Extreme Settings}
\label{app:20000editsresults}
This subsection reports the complete results of REVIVE-enhanced methods under extreme sequential editing settings. Table~\ref{20000edit} presents detailed performance metrics, including results on the \textsc{ZsRE} dataset that were omitted from the main text due to space constraints.

\subsection{Full Threshold Experiments Results}
\label{app:threshold}

This section provides detailed results that complement the sensitivity analysis reported in Section~\ref{hyper}. We present extended experiments on the singular value energy threshold $\tau$, which determines the size of the dominant singular subspace protected by \themodel. We report results for MEMIT+\themodel across three base models. Table~\ref{threshmetric} summarizes the variations of editing-related metrics under different threshold values. In addition, Figure~\ref{fig:threshglue3d} reports the corresponding GLUE performance on LLaMA3 for different choices of $\tau$. Consistent with the observations in the main text, the results indicate that \themodel maintains stable performance across a broad range of threshold values. Results on GPT-J and GPT2-XL for GLUE are omitted, as these models exhibit limited GLUE performance prior to editing, making threshold-dependent variations less informative in this setting.


\begin{table*}[t]
\centering
\caption{\footnotesize Performance results of MEMIT-REVIVE on sequential editing task under different singular value energy thresholds (10,000 Samples from CounterFact). }\label{threshmetric}
\renewcommand{\arraystretch}{0.9}
\resizebox{\textwidth}{!}{%
\begin{tabular}{c c|ccccc|ccc}
\toprule[1.5pt]
\textbf{Model} & \textbf{Thresh}  & \multicolumn{5}{c|}{\textbf{Counterfact}} & \multicolumn{3}{c}{\textbf{ZsRE}} \\
\cmidrule(lr){3-7} \cmidrule(lr){8-10}
&& \textbf{Eff.$\uparrow$} & \textbf{Para.$\uparrow$} & \textbf{Neigh.$\uparrow$} & \textbf{Flu.$\uparrow$} & \textbf{Consis.$\uparrow$} & \textbf{Eff.$\uparrow$} & \textbf{Para.$\uparrow$} & \textbf{Neigh.$\uparrow$} \\
\cmidrule{1-10}
\multirow{6}{*}{LLaMA3} 
& 0.05 & 94.46 & 86.03 & 59.70 & 587.30 & 29.83 & 78.66 & 75.89 & 29.29 \\ \cmidrule{2-10}
& 0.10 & 95.62 & 84.60 & 62.17 & 603.22 & 29.39 & 86.56 & 83.07 & 31.88 \\ \cmidrule{2-10}
& 0.15 & 95.03 & 80.60 & 64.49 & 613.66 & 29.19 & 87.10 & 83.36 & 32.41 \\ \cmidrule{2-10}
& 0.20 & 94.58 & 78.38 & 66.19 & 621.15 & 29.63 & 86.85 & 83.46 & 32.75 \\ \cmidrule{2-10}
& 0.25 & 92.96 & 73.94 & 68.94 & 624.63 & 29.49 & 83.85 & 80.23 & 33.27 \\ \cmidrule{2-10}
& 0.30 & 88.94 & 67.56 & 71.86 & 625.68 & 28.84 & 81.18 & 77.81 & 33.02 \\ 
\midrule[1pt]
\multirow{6}{*}{GPT-J} 
& 0.05 & 91.23 & 83.72 & 57.26 & 596.20 & 33.29 & 78.50 & 73.19 & 27.44 \\ \cmidrule{2-10}
& 0.10 & 97.09 & 87.01 & 67.10 & 616.15 & 40.00 & 83.87 & 77.28 & 29.77 \\ \cmidrule{2-10}
& 0.15 & 96.74 & 81.20 & 69.98 & 617.42 & 39.63 & 88.57 & 82.87 & 29.15 \\ \cmidrule{2-10}
& 0.20 & 94.95 & 76.59 & 72.13 & 621.01 & 38.36 & 85.83 & 79.97 & 29.27 \\ \cmidrule{2-10}
& 0.25 & 92.84 & 69.42 & 74.15 & 621.61 & 37.19 & 81.67 & 74.67 & 27.59 \\ \cmidrule{2-10}
& 0.30 & 88.49 & 64.30 & 74.94 & 623.53 & 36.66 & 81.32 & 73.54 & 28.56 \\ 
\midrule[1pt]
\multirow{6}{*}{GPT2-XL} 
& 0.05 & 91.89 & 80.72 & 61.13 & 575.14 & 32.12 & 62.13 & 55.40 & 25.90 \\ \cmidrule{2-10}
& 0.10 & 90.82 & 77.24 & 63.73 & 595.36 & 34.28 & 63.34 & 55.29 & 25.93 \\ \cmidrule{2-10}
& 0.15 & 87.82 & 73.39 & 65.89 & 607.06 & 35.33 & 66.19 & 58.40 & 27.13 \\ \cmidrule{2-10}
& 0.20 & 83.10 & 66.95 & 68.44 & 615.17 & 35.46 & 64.53 & 57.45 & 26.60 \\ \cmidrule{2-10}
& 0.25 & 78.77 & 61.82 & 69.66 & 618.28 & 35.17 & 58.11 & 51.80 & 26.89 \\ \cmidrule{2-10}
& 0.30 & 73.03 & 57.28 & 71.12 & 621.46 & 34.60 & 57.05 & 51.15 & 26.42 \\ 
\bottomrule[1.5pt]
\end{tabular}}
\end{table*}



\subsection{Preserving Dominant Singular Directions with REVIVE}
\label{app:specalphaedit_revive}
\begin{figure*}[thbp]

    \centering
    \captionsetup[subfigure]{font=footnotesize, skip=2pt}
    \small
    \begin{subfigure}[t]{0.19\textwidth}
        \includegraphics[width=\textwidth]{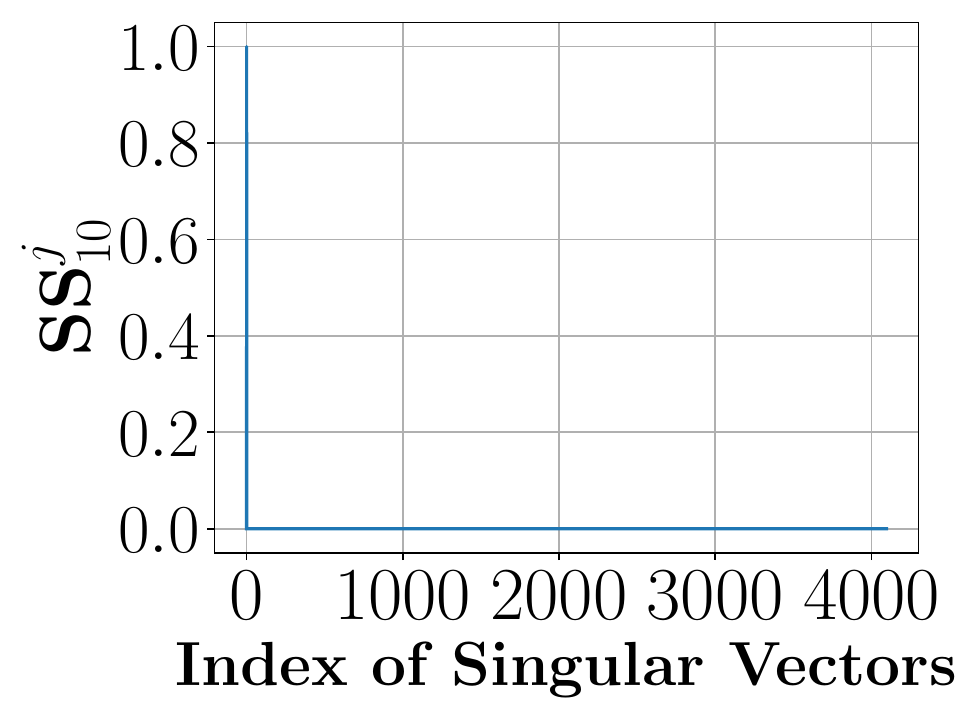}
    \end{subfigure}
    \begin{subfigure}[t]{0.19\textwidth}
        \includegraphics[width=\textwidth]{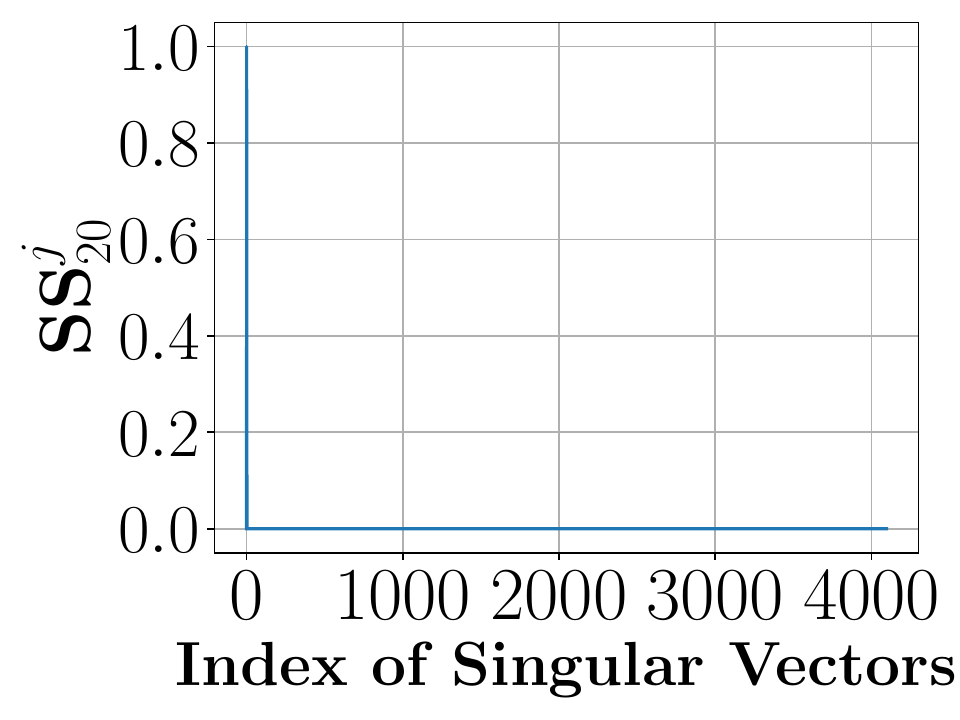}
    \end{subfigure}
    \begin{subfigure}[t]{0.19\textwidth}
        \includegraphics[width=\textwidth]{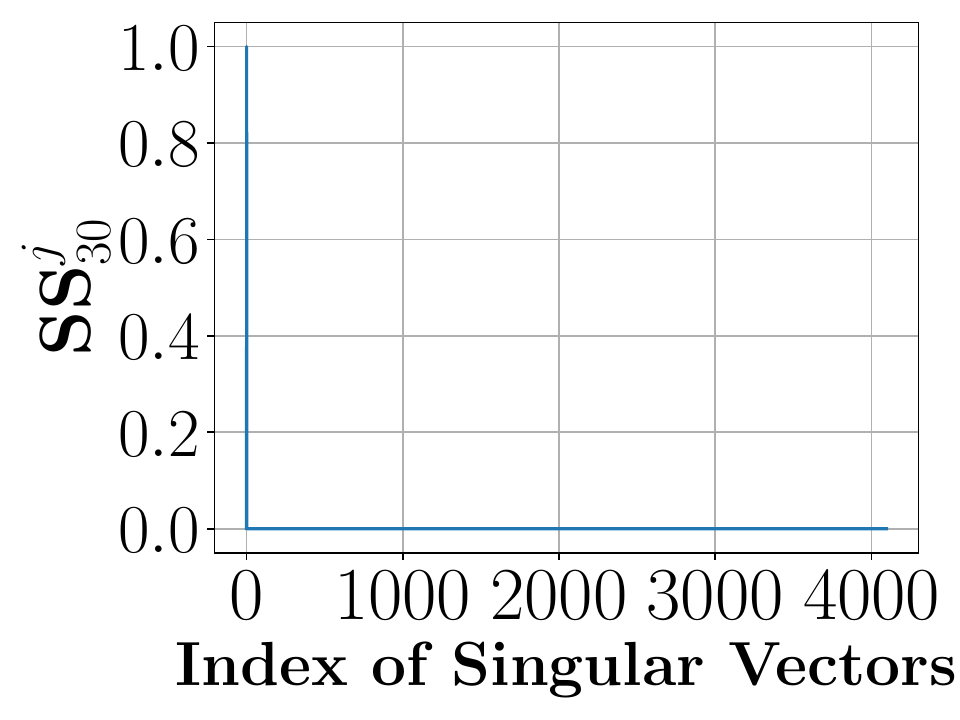}
    \end{subfigure}
    \begin{subfigure}[t]{0.19\textwidth}
        \includegraphics[width=\textwidth]{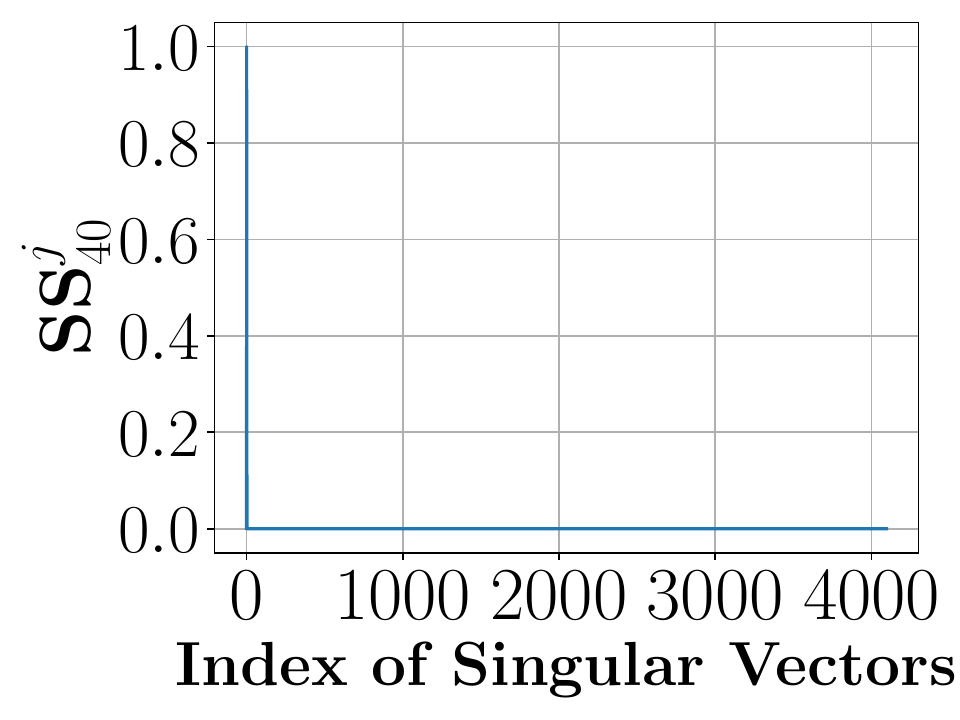}
    \end{subfigure}
    \begin{subfigure}[t]{0.19\textwidth}
        \includegraphics[width=\textwidth]{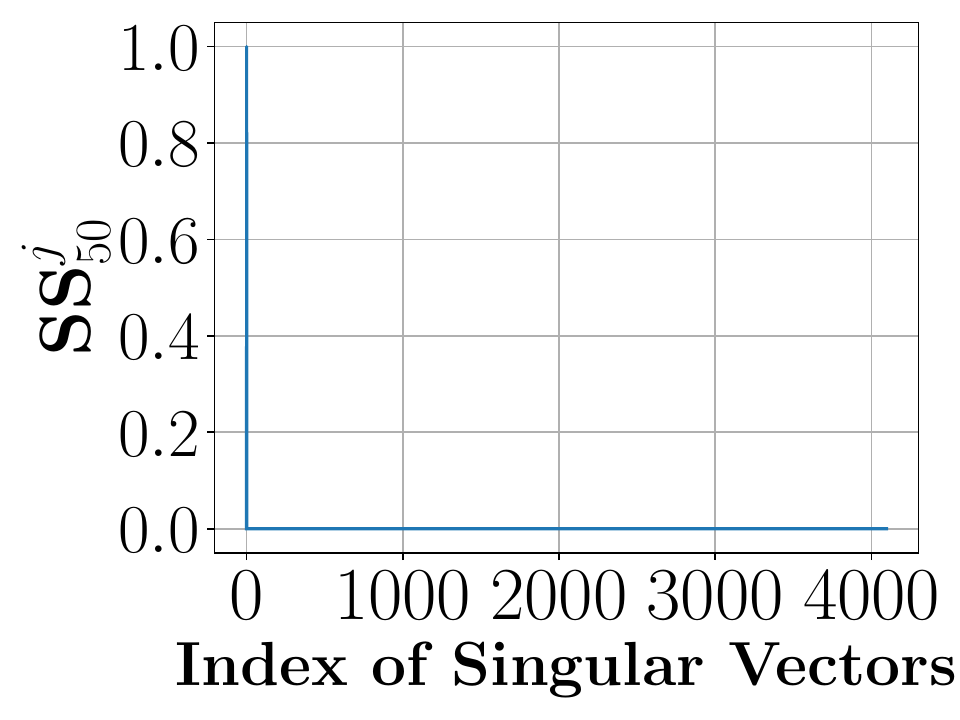}
    \end{subfigure}

    \par  

    \begin{subfigure}[t]{0.19\textwidth}
        \includegraphics[width=\textwidth]{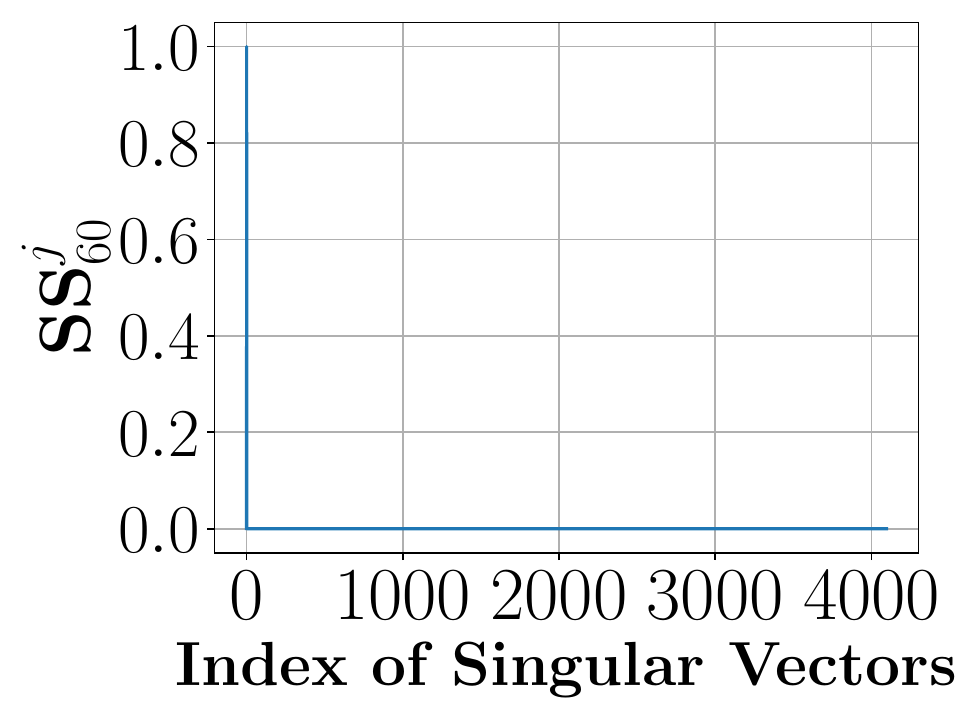}
    \end{subfigure}
    \begin{subfigure}[t]{0.19\textwidth}
        \includegraphics[width=\textwidth]{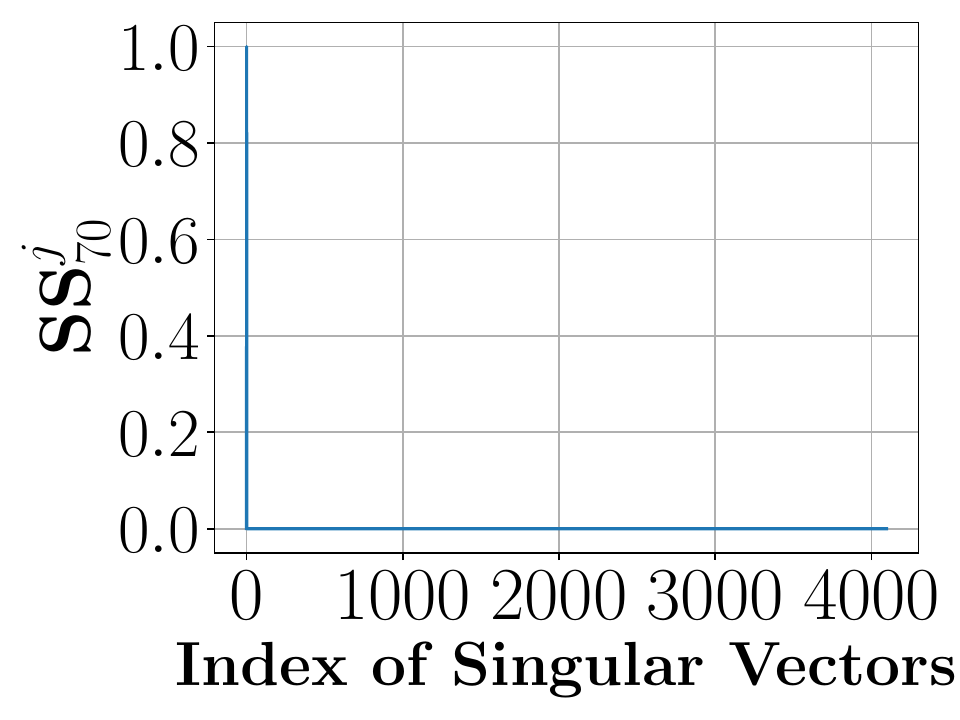}
    \end{subfigure}
    \begin{subfigure}[t]{0.19\textwidth}
        \includegraphics[width=\textwidth]{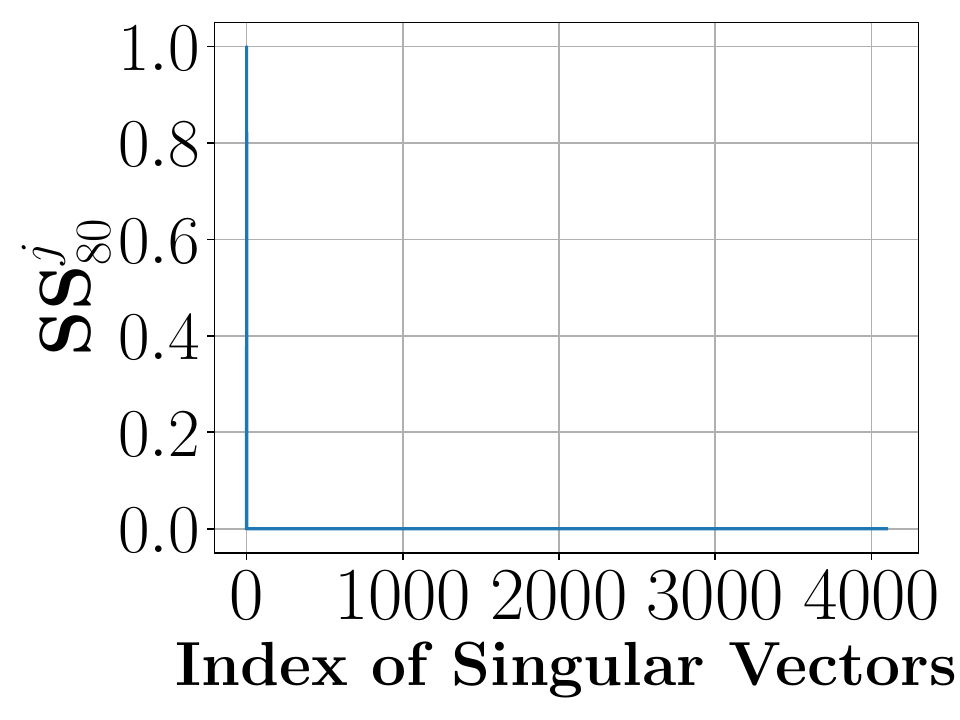}
    \end{subfigure}
    \begin{subfigure}[t]{0.19\textwidth}
        \includegraphics[width=\textwidth]{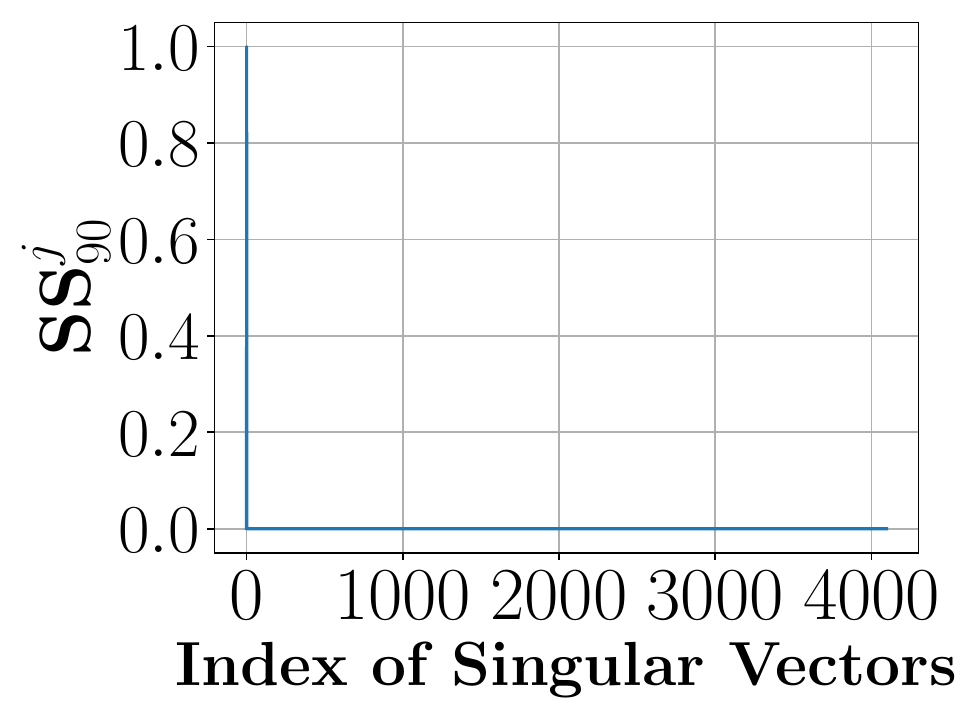}
    \end{subfigure}
    \begin{subfigure}[t]{0.19\textwidth}
        \includegraphics[width=\textwidth]{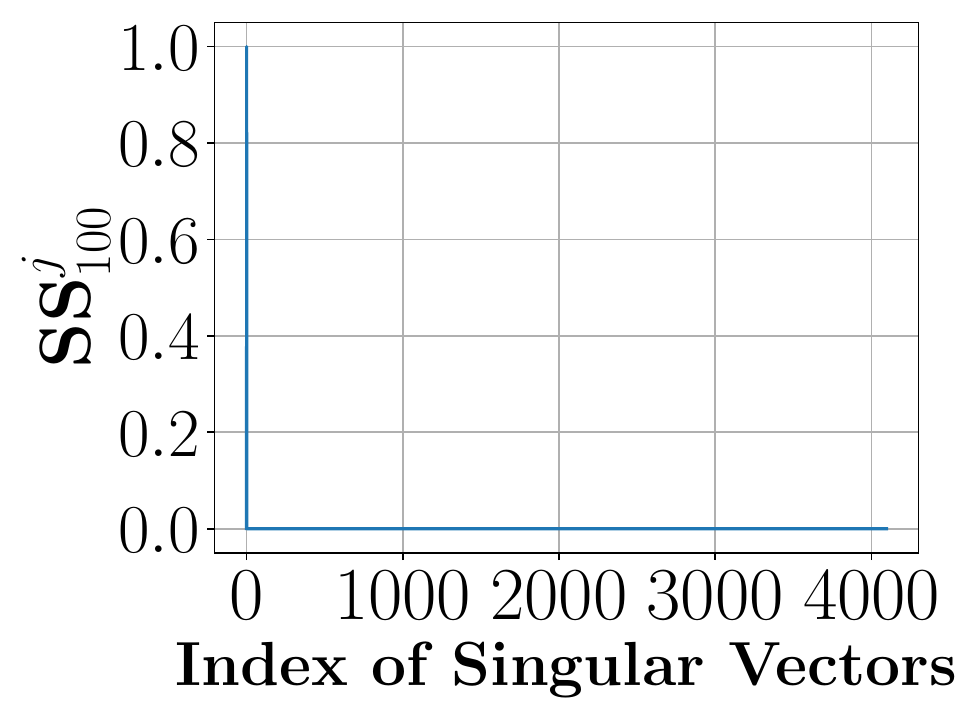}
    \end{subfigure}

\caption{Evolution of SS of MEMIT-REVIVE over sequential editing, from $\text{SS}_{10}$ to $\text{SS}_{100}$ with step size 10.}

    \label{fig:cosine-sim-MEMITREVIVE}
\end{figure*}

We now investigate whether the spectral degradation observed above can be mitigated by explicitly protecting the dominant singular subspace.
To this end, we compare the singular vector dynamics of \textsc{MEMIT} with those of \textsc{MEMIT-REVIVE}. All experiments follow the same setup as in Section~\ref{sec:memitlongsequenceediting}. Figure~\ref{fig:cosine-sim-MEMITREVIVE} reports the evolution of the maximum SS value for \textsc{MEMIT-REVIVE} throughout the entire editing sequence.

In contrast to \textsc{MEMIT}, \textsc{MEMIT-REVIVE} consistently maintains a maximum SS value of $1$ across all rounds, indicating that the dominant singular vectors remain aligned with their original directions throughout the editing process. This absence of spectral drift demonstrates that \textsc{REVIVE} effectively shields the critical subspace from cumulative perturbations. Importantly, this spectral stability directly corresponds to the preserved general performance observed on the \textsc{GLUE} benchmark.
Together with the results in Section~\ref{sec:memitlongsequenceediting}, these findings provide strong evidence that explicitly protecting the dominant singular subspace is sufficient to prevent the collapse of general abilities under long-sequence knowledge editing.

\subsection{Generalization to Meta-Learning Methods}
\label{sec:meta_learning}

To evaluate whether REVIVE generalizes beyond locate-then-edit methods to meta-learning-based editors, we conduct additional experiments using MEND \citep{mend} on the GPT-J model. Because meta-learning methods compute parameter updates independently for each sample, we evaluate sequential editing with a batch size of 1 and scale the sequence up to $1,000$ edits.

As shown in Table~\ref{tab:mend_generalization}, the standard MEND model suffers from rapid catastrophic collapse under sequential updates. For instance, on the ZSRE dataset, its Efficacy drops to $2.40\%$ after just $50$ edits and completely collapses to $0.00\%$ by $1,000$ edits. In sharp contrast, applying REVIVE as a plug-and-play module effectively stabilizes the process. After $1,000$ sequential edits, MEND equipped with REVIVE maintains $81.20\%$ Efficacy on ZSRE and $86.60\%$ on COUNTERFACT, while generative metrics such as Fluency and Consistency remain stable. 

This strong generalization stems from REVIVE's structural design. REVIVE operates purely on the resulting update matrix within the spectral basis of the pretrained weights, making no assumptions about the underlying algorithm or hypernetwork generating the update. Consequently, whether the update is derived from a closed-form editor or a meta-learning method, REVIVE can consistently filter out harmful components aligned with the dominant singular subspace, thereby preserving the model's general abilities.

\begin{table*}[t] \centering
\caption{Performance comparison of MEND and MEND + REVIVE on the GPT-J model under sequential editing. Results demonstrate that REVIVE successfully prevents catastrophic collapse for meta-learning methods.} 
\begin{tabular}{ll ccccc ccc} \toprule \multirow{2}{*}{\textbf{Edits}} & \multirow{2}{*}{\textbf{Method}} & \multicolumn{5}{c}{\textbf{COUNTERFACT}} & \multicolumn{3}{c}{\textbf{ZSRE}} \\ \cmidrule(lr){3-7} \cmidrule(lr){8-10} & & \textbf{Eff. $\uparrow$} & \textbf{Para. $\uparrow$} & \textbf{Neigh. $\uparrow$} & \textbf{Flu. $\uparrow$} & \textbf{Consis. $\uparrow$} & \textbf{Eff. $\uparrow$} & \textbf{Para. $\uparrow$} & \textbf{Neigh. $\uparrow$} \\ \midrule \multirow{2}{*}{1} & MEND & 100.00 & 50.00 & 30.00 & 627.95 & 31.78 & 75.00 & 75.00 & 66.67 \\ & + REVIVE & 100.00 & 50.00 & 30.00 & 626.93 & 31.08 & 75.00 & 75.00 & 66.67 \\ \midrule \multirow{2}{*}{50} & MEND & 54.00 & 48.00 & 54.00 & 488.27 & 6.18 & 2.40 & 1.60 & 93.11 \\ & + REVIVE & 86.00 & 41.00 & 65.60 & 617.13 & 28.19 & 95.91 & 91.25 & 30.27 \\ \midrule \multirow{2}{*}{100} & MEND & 43.00 & 42.50 & 58.50 & 524.40 & 1.19 & 0.20 & 0.60 & 0.02 \\ & + REVIVE & 89.00 & 45.50 & 60.26 & 618.11 & 28.98 & 97.41 & 89.98 & 27.85 \\ \midrule \multirow{2}{*}{500} & MEND & 45.40 & 45.90 & 54.10 & 498.58 & 2.92 & 0.34 & 0.34 & 0.64 \\ & + REVIVE & 85.20 & 45.50 & 60.28 & 615.47 & 28.85 & 86.01 & 78.96 & 27.09 \\ \midrule \multirow{2}{*}{1000} & MEND & 46.10 & 46.40 & 53.60 & 488.05 & 2.41 & 0.00 & 0.00 & 0.08 \\ & + REVIVE & 86.60 & 50.80 & 56.77 & 619.08 & 29.38 & 81.20 & 74.38 & 25.97 \\ \bottomrule \end{tabular} 
\label{tab:mend_generalization} \end{table*}

\subsection{Computational Overhead Analysis}
\label{app:timeanalysis}
REVIVE introduces additional spectral projections based on singular value decomposition, which may incur extra computational cost. To quantify this overhead, we measure the runtime of a baseline editing method and its REVIVE-augmented variant when editing $100$ samples. Since the projection operation is independent of the underlying editing algorithm, we conduct this analysis using MEMIT as a representative baseline. As can be seen from Table \ref{tab:runtime}, since knowledge editing is applied only to the FFN weight matrices in a subset of model layers, the number of required SVD computations is limited, making the additional projection overhead negligible.

\begin{table}[H]
\centering
\caption{Runtime comparison under the setting of editing 100 samples.}
\label{tab:runtime}
\begin{tabular}{lccc}
\toprule
Method &  GPT-J & LLaMA3-8B \\
\midrule
MEMIT        & 319.21s & 813.86s \\
+REVIVE & 332.14s & 829.39s \\
\bottomrule
\end{tabular}
\end{table}

\end{document}